\RequirePackage{fix-cm}
\documentclass[twocolumn]{svjour3}          
\smartqed  
\usepackage{enumitem} 
\usepackage{graphicx} 
\usepackage{tikz, tkz-euclide} 
\usepackage{gensymb} 
\usepackage{algorithm}
\usepackage[noend]{algpseudocode}
\usepackage[miktex]{gnuplottex}  
\usepackage[caption=false]{subfig}
\usepackage{filecontents} 
\usepackage{datatool} 
\usepackage{pgfplotstable}
\usetikzlibrary{arrows}
\usetikzlibrary{positioning,calc}
\usetikzlibrary{decorations.pathreplacing}
\usetikzlibrary{decorations.markings}
\usetikzlibrary{fit}
\usetikzlibrary{shapes.callouts}
\usetikzlibrary{shapes.geometric}
\usetikzlibrary{shapes.arrows}
\usetikzlibrary{matrix}
\usetikzlibrary{intersections}
\usetikzlibrary{math}
\usepackage{pgfplots}
\pgfplotsset{compat=newest}
\usepackage{subfiles}
\usepackage[binary-units=true]{siunitx}
\usepackage{xcolor,colortbl}
\usepackage{tabularx}
\usepackage{array}
\usepackage{rotating}
\usepackage{tabularx}
\usepackage{graphbox} 
\usetikzlibrary{matrix}
\usetikzlibrary{intersections}
\usetikzlibrary{math}
\usepackage{amsmath,amssymb,amsfonts}
\usepackage[english]{babel}
\usepackage{blindtext}
\usepackage[skins]{tcolorbox} 
\usetikzlibrary{shapes,calc}
\usepackage{hyperref}
\usepackage[]{natbib}
\usepackage{etoolbox,lipsum} 
\usepackage{float} 
\usepackage{capt-of,etoolbox}
\usepackage{amsmath,amssymb,amsfonts}
\usepackage{filecontents} 
\usepackage{datatool} 
\usepackage{svg}
\usepackage{xcolor}
\usepackage{tikz,pgfplots}
\usepgfplotslibrary{fillbetween} 
\usetikzlibrary{fadings} 
\tikzfading[name=myfading, top color=transparent!100, bottom color=transparent!0]
\usepackage{pbox} 
\usepackage{tikz,pgfplots,pgfplotstable}
\usepackage[nodayofweek,level]{datetime}

\definecolor{mygreen}{rgb}{0.1,0.1,0.1}

\definecolor{mygreen2}{RGB}{102,252,102}

\newcommand{\norm}[1]{\left\lVert#1\right\rVert} 
\newcommand{\reffig}[1]{Fig.~\ref{#1}}
\newcommand{\reftab}[1]{Tab.~\ref{#1}}
\newcommand{\refsec}[1]{Sec.~\ref{#1}}
\def\CC{{C\nolinebreak[4]\hspace{-.05em}\raisebox{.4ex}{\tiny\bf ++}}} 

\DeclareMathOperator*{\argmax}{argmax} 

\begin{document}

\title{Semi-Supervised Semantic Mapping through Label Propagation with Semantic Texture Meshes
}

\author{Radu Alexandru Rosu \and
        Jan Quenzel \and
        Sven Behnke
}

\institute{All authors are with the Autonomous Intelligent Systems Group, University of Bonn, Bonn, Germany\\
\email{\{rosu, quenzel, behnke\}@ais.uni-bonn.de}
}
\date{Received: \formatdate{19}{7}{2018} / Accepted: \formatdate{30}{5}{2019}}

\maketitle

\thispagestyle{empty}
\pagestyle{empty}

\begin{tikzpicture}[remember picture,overlay]
\node[anchor=north west,align=left,font=\sffamily,yshift=-0.2cm] at (current 
page.north west) {%
  In: International Journal of Computer Vision (IJCV), 2019
};
\node[anchor=north east, align=right,font=\sffamily,yshift=-0.2cm] at (current 
page.north east) {%
  DOI: \href{https://doi.org/10.1007/s11263-019-01187-z}{10.1007/s11263-019-01187-z}
};
\end{tikzpicture}%

\begin{abstract}
Scene understanding is an important capability for robots acting in unstructured environments. 
While most SLAM approaches provide a geometrical representation of the scene, a semantic map is necessary for more complex interactions with the surroundings.
Current methods treat the semantic map as part of the geometry which limits scalability and accuracy.
We propose to represent the semantic map as a geometrical mesh and a semantic texture coupled at independent resolution.
The key idea is that in many environments the geometry can be greatly simplified without loosing fidelity, while semantic information can be stored at a higher resolution, independent of the mesh.
We construct a mesh from depth sensors to represent the scene geometry and fuse information into the semantic texture from segmentations of individual RGB views of the scene.
Making the semantics persistent in a global mesh enables us to enforce temporal and spatial consistency of the individual view predictions.
For this, we propose an efficient method of establishing consensus between individual segmentations by iteratively retraining semantic segmentation with the information stored within the map and using the retrained segmentation to re-fuse the semantics.
We demonstrate the accuracy and scalability of our approach by reconstructing semantic maps of scenes from NYUv2 and a scene spanning large buildings.

\end{abstract}

	\newcommand\coverLeftTrim{600bp}
	\newcommand\coverBottomTrim{290bp}
	\newcommand\coverRightTrim{450bp}
	\newcommand\coverTopTrim{260bp}

	\newcommand\coverLeftTrimFull{2300bp}
	\newcommand\coverBottomTrimFull{1500bp}
	\newcommand\coverRightTrimFull{2500bp}
	\newcommand\coverTopTrimFull{1500bp}
	
	\newcommand\newcoverLeftTrim{0bp}
	\newcommand\newcoverBottomTrim{0bp}
	\newcommand\newcoverRightTrim{200bp}
	\newcommand\newcoverTopTrim{150bp}

	\begin{figure}[!ht]
		\includegraphics[trim=\newcoverLeftTrim{} \newcoverBottomTrim{} \newcoverRightTrim{} \newcoverTopTrim{},clip, width=0.5\textwidth]{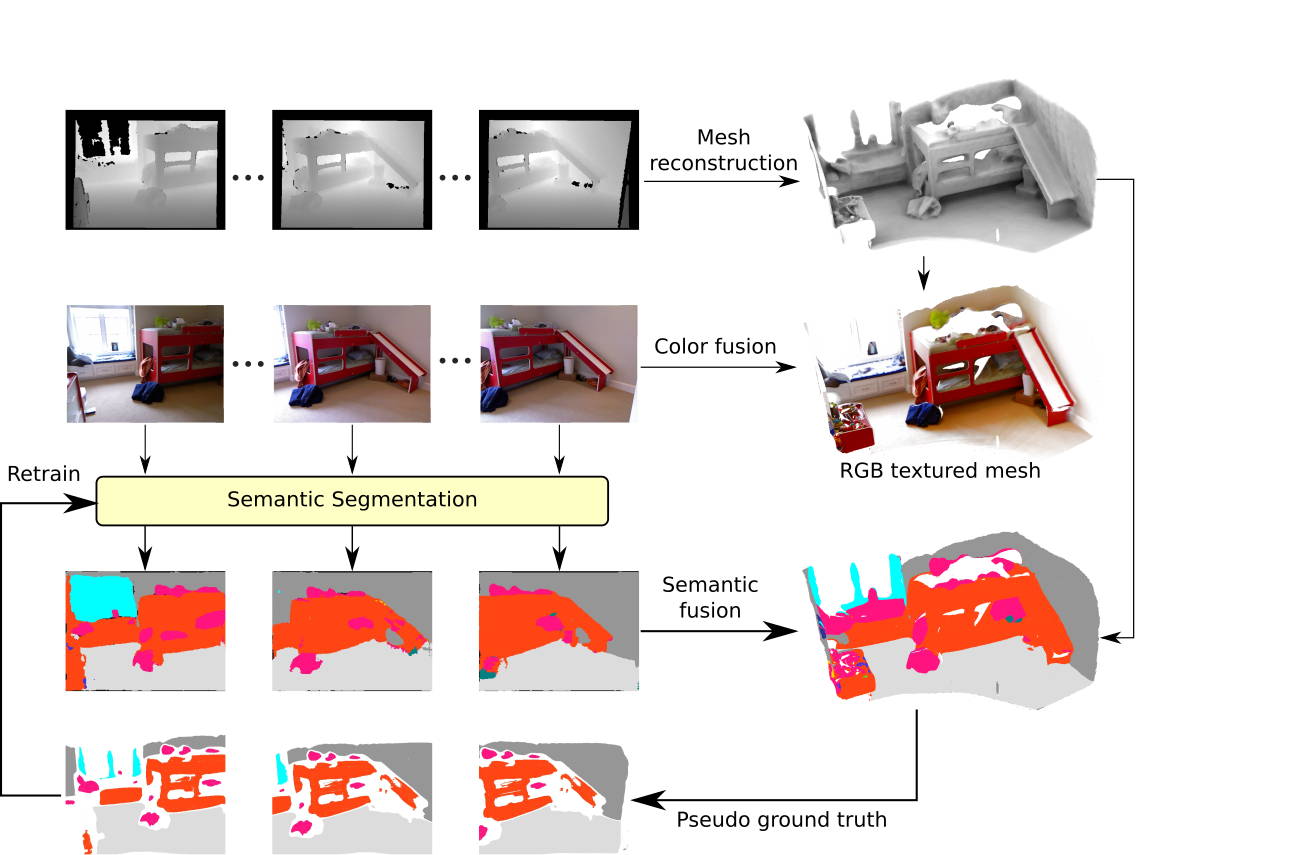}
	
		\caption{Semantic Reconstruction: We generate a mesh with RGB texture and semantic annotations. The mesh enables us to ensure temporal and spatial consistency between semantic predictions and allows us to perform label propagation for improved semantic segmentation. Color coding of semantic labels correspond to NYUv2 dataset~\citep{silberman2012nyu}.}
		\label{fig:teaser}
	\end{figure}
	
	\section{Introduction}
	Robots acting in real-world environments need the ability to understand their surroundings, and know their location within the environment. While the problem of geometrical mapping and localization can be solved through SLAM methods~\citep{zollhoefer2018survey}, many tasks require knowledge about the semantic meaning of objects or surfaces in the environment. The robot should, for instance, be able to recognize where the obstacles are in the scene, and also understand whether those obstacles are cars, pedestrians, walls, or otherwise.

	The problem of building maps has been extensively studied~\citep{kostavelis2015survey}. Most approaches can be grouped into the following three categories, based on map representation:
	\noindent\begin{itemize}
		\item Voxel-based: The scene is discretized into voxels, either using a regular grid, or an adaptive octree. Each voxel stores the binary occupancy value (occupied, empty, unknown) or the distance to the surface commonly referred to as Signed Distance Function (SDF).
		\item Surfel-based: The map is represented by small surface elements, which store the mean and covariance of a set of 3D-points. Surfels suffer from less discretization errors than voxels.
		
		\item Mesh-based: The map is represented as a set of vertices with faces between them. This naturally fills holes and allows for fast rendering using established graphics pipelines.
	\end{itemize}

	Current semantic mapping systems treat the semantic information as part of the geometry, and store label probabilities per map element (voxel, sufel or mesh vertex/face). This approach has the intrinsic disadvantage of coupling the resolution of the geometrical representation to the semantics, requiring a large number of elements to represent small semantic objects or surface parts. This is an undesirable effect as it leads to unnecessary memory usage especially in man-made environments, where the geometry is mostly planar, and high geometrical detail would be redundant. Often, it suffices to represent the semantics relative to a rough geometric shape.

	\begin{figure*}[ht!]
		\centering
		\includegraphics[width=\textwidth]{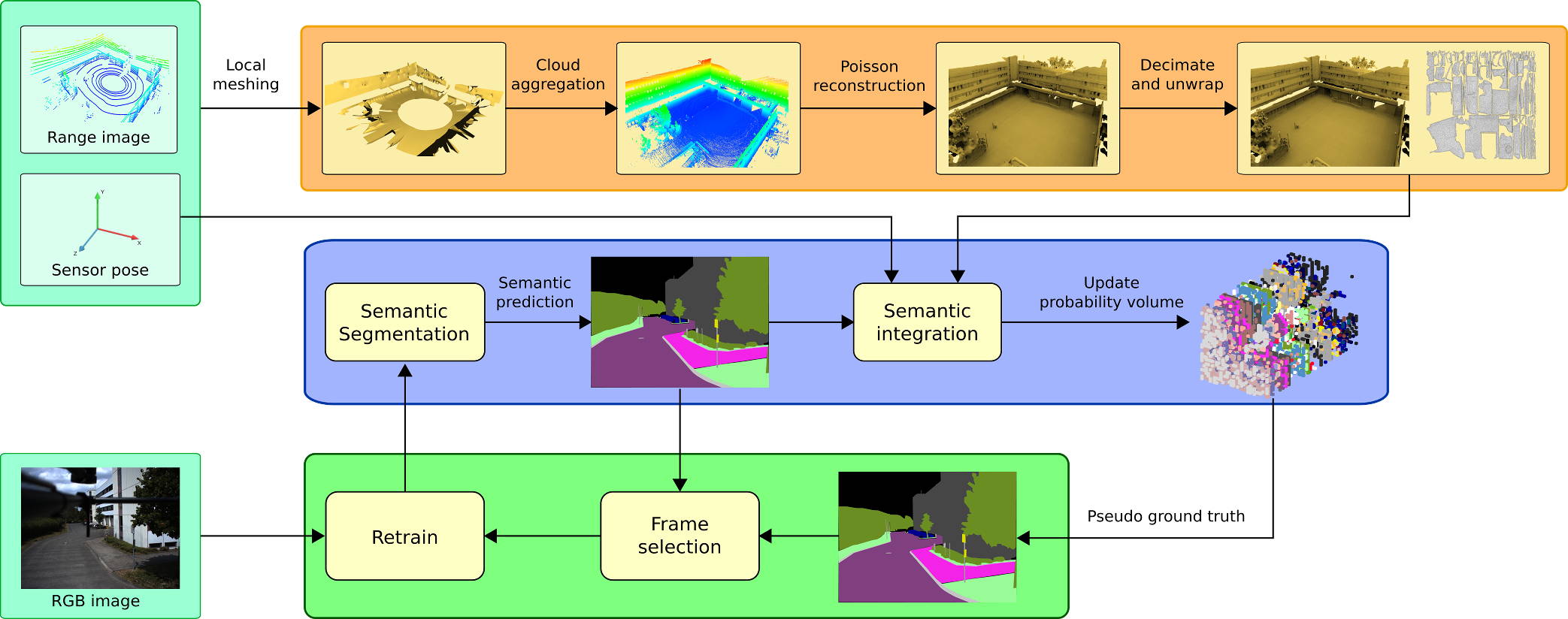}
		\caption{System overview: Individual range images are used to create local meshes for fast normal estimation and cloud simplification. The resulting points, equipped with normals, are aggregated into a global point cloud. Poisson reconstruction is employed to extract a mesh, which is simplified and unwrapped to obtain a lightweight scene representation.
			In the next step, we perform semantic integration. Individual RGB views are segmented using RefineNet~\citep{lin2017refinenet}. The semantic labels are then probabilistically fused into a semantic texture. Finally, label propagation is performed by inferring pseudo ground truth views, and using them to retrain the predictor. The retrained semantic segmentation is used to re-fuse the labels into the mesh yielding a more accurate semantic map.}
		\label{fig:overview}
	\end{figure*}
	
\begin{figure*}[]
		\captionsetup[subfloat]{labelformat=empty}
		\centering

		\subfloat[a) Fuse]{
			\fcolorbox{white}{white}{\includegraphics[width=0.3\textwidth] {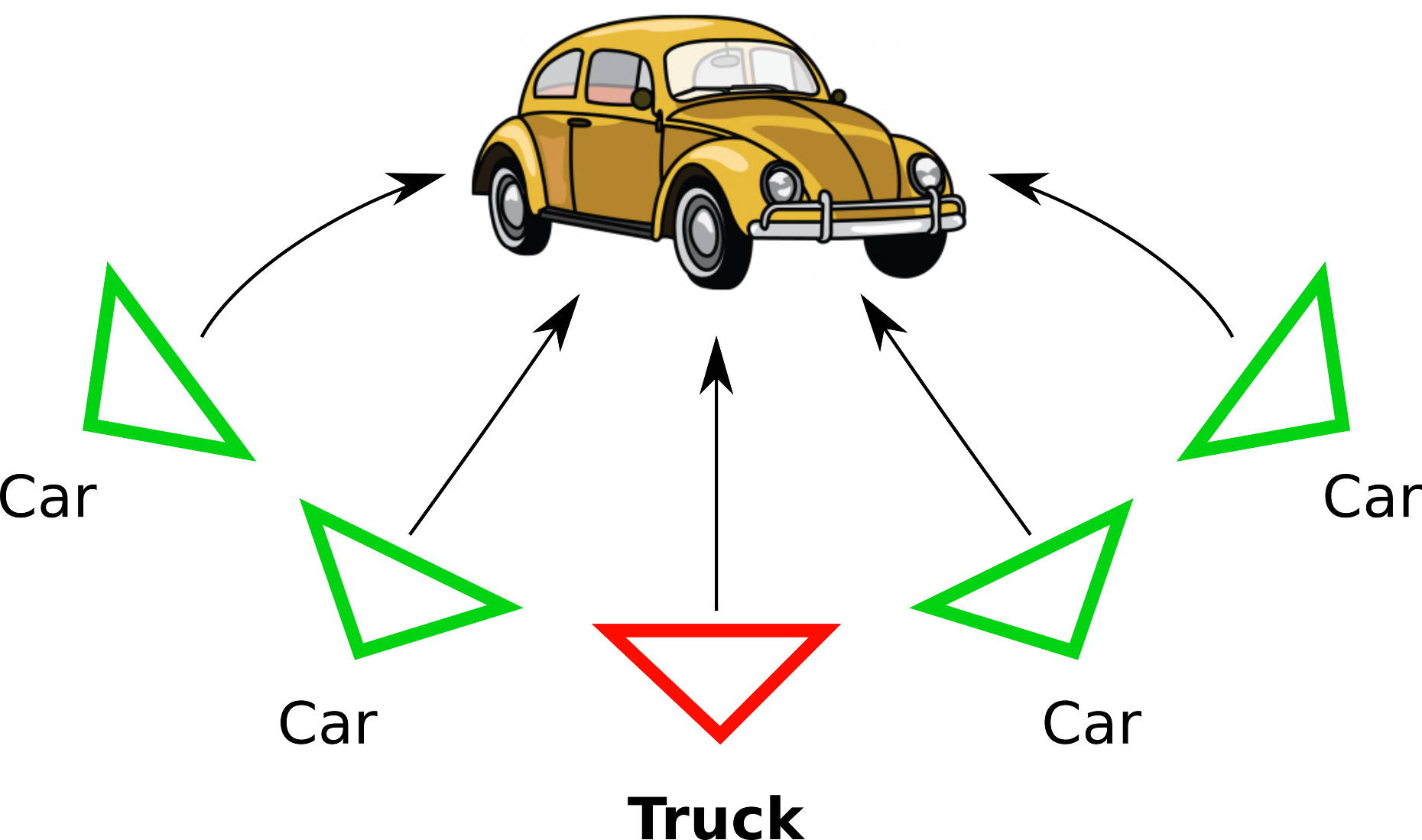}}
		}
		\subfloat[b) Retrain]{
			\fcolorbox{white}{white}{\includegraphics[width=0.3\textwidth] {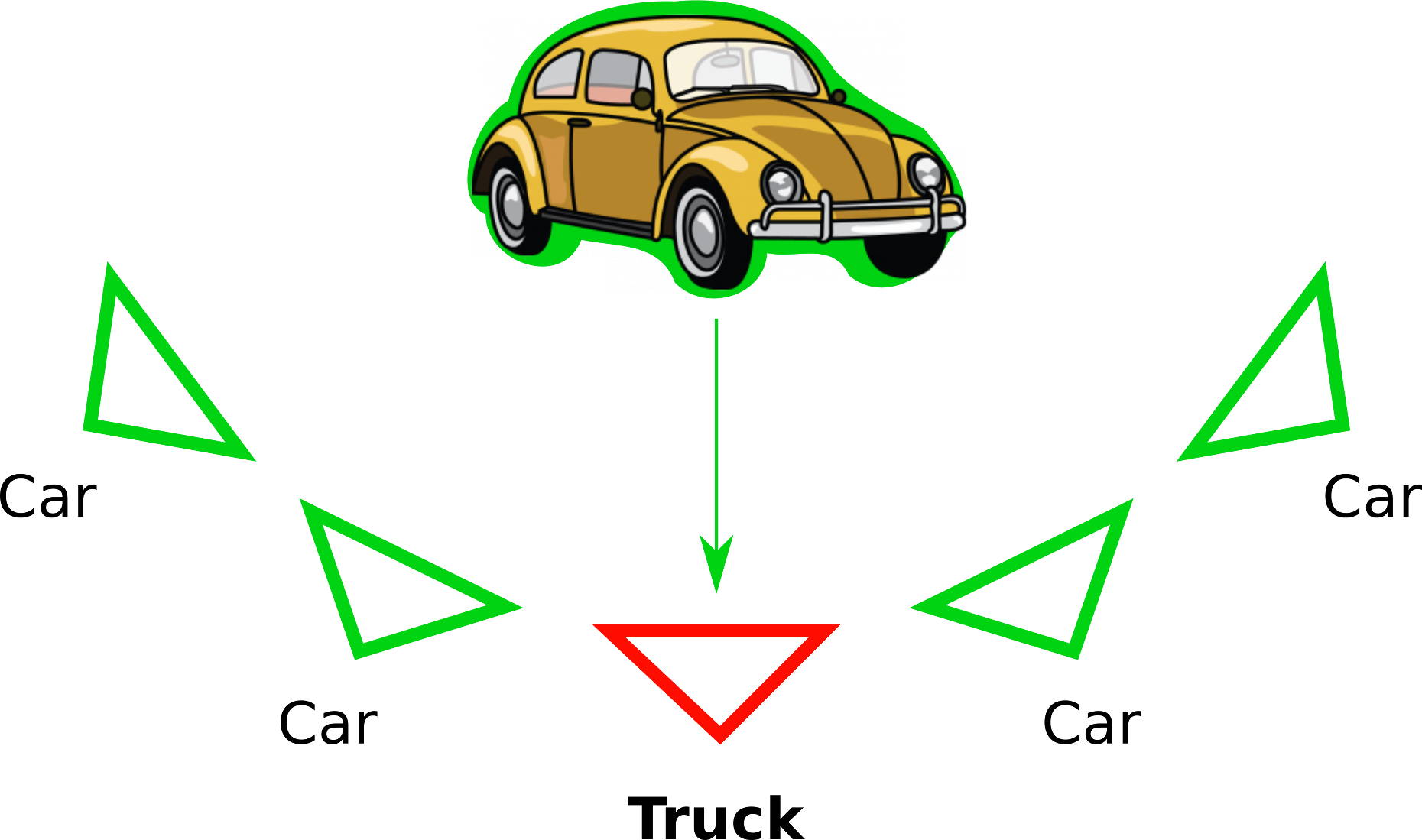}}
		}
		\subfloat[c) Re-fuse]{
			\fcolorbox{white}{white}{\includegraphics[width=0.3\textwidth] {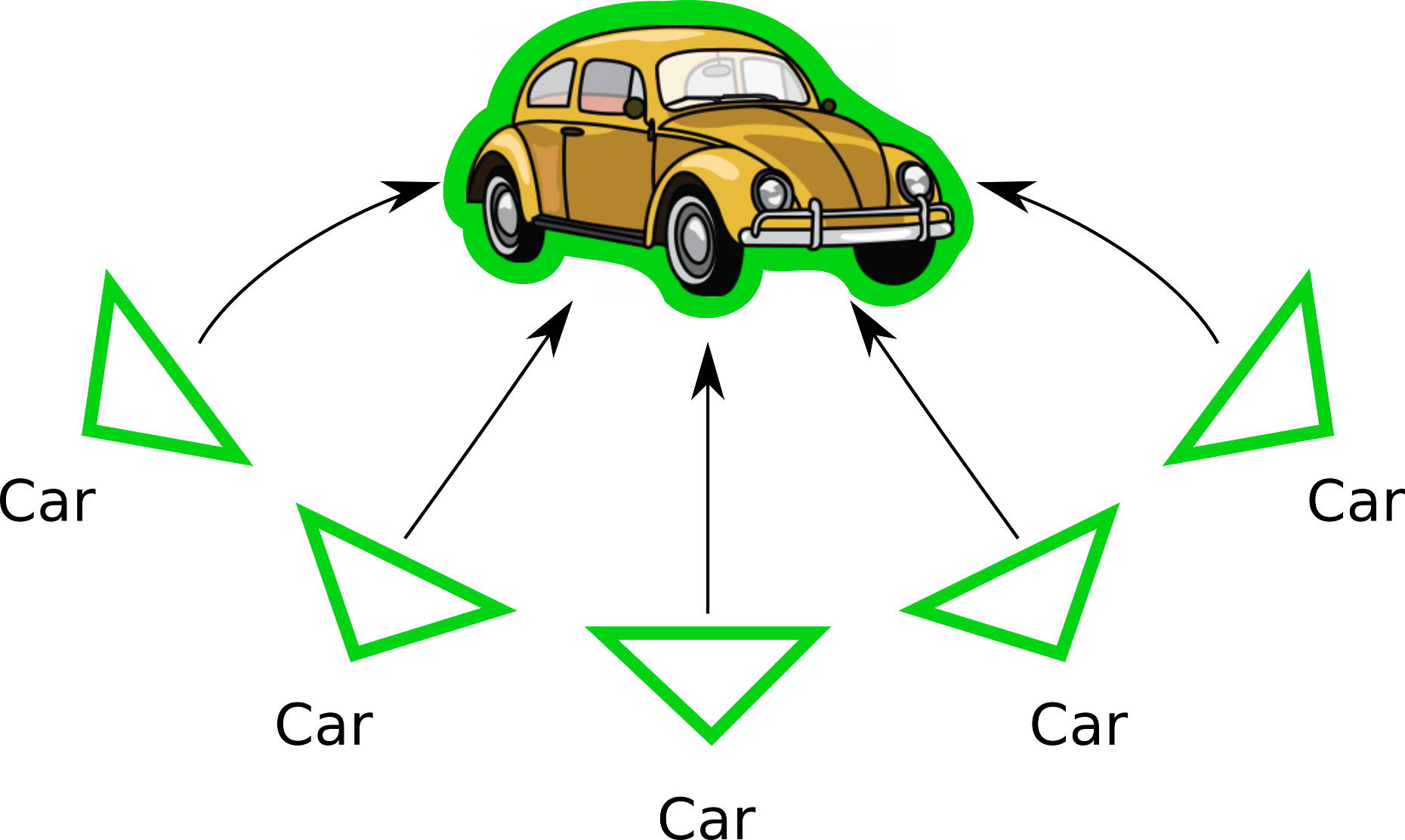}}
		}
		\caption{Label propagation: Semantic segmentations are probabilistically fused in the scene (left). Frames with largest deviation (middle) from the stable mesh are used for retraining. Inference is repeated for all images and re-fused into the scene mesh to achieve a more accurate semantic map. }
		\label{fig:LabelPropagation}
	\end{figure*}

	The key idea of our approach, visualized in \reffig{fig:teaser}, is to couple the scene geometry with the semantics at independent resolution by using a semantic texture mesh. In this, the scene geometry is represented by vertices and faces, whereas the semantic texture categorizes the surface with higher resolution. This allows us to represent semantics and geometry at different resolutions in order to build a large semantic map, while still maintaining a low memory usage.
	As our segmentation module we make use of RefineNet~\citep{lin2017refinenet} to predict a semantic segmentation for each individual RGB view of the scene. These predictions are probabilistically fused  onto the semantic texture that is supported by a coarse mesh representing the scene geometry.
	Having a globally persistent semantic map enables us to establish a temporal and spatial consistency that was previously unobtainable for individual-view predictor. To this end, we propose to propagate labels from the stable mesh by projection onto each camera frame, in order to retrain the semantic segmentation in a semi-supervised manner.
	Expectation Maximization (EM) is then carried out by alternating between fusing semantic predictions and propagating labels. This iterative refinement allows us to cope with view points which were not common in the training dataset. A predictor pretrained on street-level segmentation will not work well on images captured by a micro aerial vehicle (MAV) at higher altitudes or close to buildings.
	However, projecting confident semantic labels fused from street level onto less confident parts of views will enable to learn the semantic segmentation of new viewpoints (see \reffig{fig:LabelPropagation}).

	We compare our method with SemanticFusion~\citep{mccormac2017semanticfusion}, and evaluate the accuracy on the NYUv2 dataset~\citep{silberman2012nyu}.
	We show that the increased resolution of the semantic texture allows for more accurate semantic maps. Finally, propagation and retraining further improve the accuracy, surpassing SemanticFusion in every class.

	To showcase the benefits of textured meshes in terms of scalability and speed, we also recorded a dataset spanning multiple buildings, annotated with the \num{66} classes of the Mapillary dataset~\citep{neuhold2017mapillary}. We demonstrate that we are able to construct a large map using both RGB and semantic information in a time- and memory-efficient manner.

	\fboxsep=3mm
	\fboxrule=0pt

	\section{Related Work}
The annotation of large datasets is a costly and time-consuming matter. Hence, the automation of annotation as well as the transfer of knowledge across different domains and datasets are active research topics. 
Most networks for image segmentation or their respective backbone (e.g. RefineNet~\citep{lin2017refinenet} and Mask R-CNN~\citep{he2017mask} with ResNet-backbone~\citep{he2016resnet} ) are nowadays pretrained on large datasets like ImageNet~\citep{deng2009imagenet} and only finetuned for a specific dataset or purpose.

\Citet{vezhnevets2012active} classify super pixels in an automated manner using a pairwise conditional random field and request human intervention based on the \textit{expected change}. Likewise, \Citet{jain2016prop} use a Markov Random Field for joint segmentation across images given region proposals with similar saliency. The resulting proposals are later fused to obtain foreground masks while supervision is requested based on an images influence, diversity and the predicted annotation difficulty.
Instead, \Citet{yang2017suggestive} cluster unannotated data based on cosine similarity to other images and simply choose per cluster the one with most similar images for human labeling. 
\Citet{mackowiak2018cereals} take a more cost-centric approach and train one CNN for semantic segmentation and one for a cost model that estimates the necessary clicks for annotating a region.
The cost model predictions are then fused with the vote entropy of the segmenting networks activation and supervision is requested for a fixed number of regions.
\Citet{castrejon2017prnn} provide with Polygon-RNN a more interactive approach. Given a (drawn) bounding box around an object, the RNN with VGG-16 backbone~\Citep{VGG16} predicts an enclosing polygon around the object. The polygon can be corrected by a human annotator and fed back into the RNN to improve the overall annotation accuracy. \Citet{acuna2018prnnpp} improve upon Polygon-RNN through architecture modifications, training with reinforcement learning and increased polygonal output resolution.

Most semantic segmentation methods are not real-time capable. Hence, \Citet{sheikh2016prop} proposed to use quad-tree based super pixels where only the center is classified by a random forest and labels are propagated to a new image if the super pixels location and intensity do not change significantly. This inherently assumes small inter frame motion but does not take spatial correspondences into account, yet runs on a CPU with up to 30 fps.

While image segmentation is fairly advanced, labeling point clouds still has a large potential for improvement. Voxel-based approaches like OctNet~\citep{riegler2017octnet} precompute a voxel grid and apply 3D- convolutions. Most grid cells are empty for sparse LIDAR point clouds. Hence, recent research shifts towards using points directly~\citep{qi2017pointnet,qi2017pointnetpp}, forming cluster of points~\citep{landrieu2017spg}, applying convolutions on local surfaces~\citep{tatarchenko2018tangentconv} or lifting points to a high-dimensional sparse lattice~\citep{su2018splatnet}. 
These methods do not enforce consistent labels for sequential data and would need to be recomputed once new data is aggregated while being strongly memory constrained.
Nevertheless, \Citet{zaganidis2018segicp} showed that semantic predictions can improve point cloud registration with GICP and NDT.

	Semantic reconstruction and mapping received much attention in recent years.
	\Citet{civera2011towards}, for example, paved the way towards a semantic SLAM system by presenting an object reasoning system, able to learn object models using feature descriptors in an offline step and then recognizing and registering them to the map at run time. However, their system was limited to a small number of objects and apart from the recognized objects, the map was represented only as a sparse point cloud.
	\Citet{bao2011semantic} exploit semantics for Structure-from-Motion (SfM) to reduce the initial number of possible camera configurations and add a semantic term during Maximum-Likelihood estimation of camera poses and scene structure.
	Subsequently, \citet{bao2013dense} use the estimated scene structure to generate dense reconstructions from learned class-specific mean shapes with anchor points. The mean shape is warped with a 3D thin plate spline and local displacements are obtained from actual details of the instance.
	
	Instead, \Citet{hane2013joint} fuse single frame depth maps from plane sweep stereo to reconstruct a uniform voxel grid and jointly label these voxels by rephrasing the fusion as a multi-label assignment problem. A primal-dual algorithm solves the assignment while penalizing the transition between two classes based on class-specific geometry priors for surface orientation. Their method is also able to reconstruct and label underlying voxels and not only visible ones. 
	
	In subsequent work, more elaborate geometry priors have been learned, e.g. using Wolff shapes from surface normal distributions~\Citep{hane2014wolff} and recently end-to-end-learned with a 3D-CNN~ \Citep{cherabier2018learn}. The data term in the optimization has been improved~\citep{savinov16visconst}, memory consumption and runtime reduced~\Citep{cherabier2016multi,blaha2016large}, and an alignment to shape priors integrated~\citep{maninchedda2016head}.

\Citet{schonberger2018semloc} utilize the approach of \citet{hane2013joint} for visual localization with semantic assistance to learn descriptors. An encoder-decoder CNN is trained on the auxiliary task of Scene Completion given incomplete subvolumes. The encoder is then used for descriptor estimation. Given a bag of words with a corresponding vocabulary one can thus query matching images for a given input frame.

For incremental reconstruction, \Citet{stuckler2014semantic} presented a densely- represented approach using a voxel grid map. Semantic labels were generated for individual RGB-D views of the modeled scene by a random forest. Labels were then assigned projectively to each occupied voxel and fused using a Bayesian update. The update effectively improved the accuracy of backprojected labels compared to instantaneous segmentation of individual RGB-D views.

	Similarly, \Citet{hermans2014dense} fused semantic information obtained from segmenting RGB-D frames using random forests but represented the map as a point cloud. Their main contribution was an efficient spatial regularizing Conditional Random Field (CRF), which smoothes semantic labels throughout the point cloud. \Citet{li2016semi} extended this approach to monocular video while using the semi-dense map of LSD-SLAM~\citep{engel2014lsd}. Here, the DeepLab-CNN~\citep{chen2018deeplab} was used instead of a random forest for segmentation.
	
	\Citet{vineet2015incremental} achieve a virtually unbounded scene reconstruction through the use of an efficient voxel hashed data structure for the map. This further allows them to incrementally reconstruct the scene. Instead of RGB-D cameras, stereo cameras were employed and depth was estimated by stereo disparity. Semantic segmentation was performed through random forest.
	The requirement for dense depth estimates is lifted in the approach of \Citet{kundu2014joint}. They use only sparse triangulated points obtained through monocular Visual SLAM and recover a dense volumetric map through a CRF that jointly infers semantic category and occupancy for each voxel.

	A different approach is used in the keyframe-based monocular SLAM system by \Citet{tateno2017cnn} where a CNN predicts per keyframe the pixel-wise monocular depth and semantic labels.

\Citet{lianos2018vso} reduce drift in visual odometry via establishing of semantic correspondences over longer periods than possible with pure visual correspondences. The intuition is that the semantic class of a car will stay a car even under diverse illumination and view point changes while visual correspondences may be lost. However, semantic correspondences are not discriminative in the short term.

	\Citet{tulsiani2017multi} perform single view reconstruction of a dense voxel grid with a CNN. During training multiple views of the same scene guide the learning by enforcing the consistency of viewing rays incorporating information from multiple sources like foreground masks, depth, color or semantics. Whereas, \Citet{ma2017multiview} examined the use of warping RGB-D image sequences into a reference frame for semantic segmentation to obtain more consistent predictions.
	\citet{sun2018roctomap} extend OctoMap~\citep{hornung2013octomap} with a LSTM per cell to be able to account for long term changes like dynamic obstacles.
	
\Citet{nakajima2018geosem} segment surfels from the depth image and semantic prediction using connected component analysis and further refined incrementally over time. Geometric segments along with their semantic label are stored in the 3D map. The probabilistic fusion combines the rendered current view with the current frame and its low resolutional semantics.

	Surfels are also used in the work of \Citet{mccormac2017semanticfusion}. The authors integrated semantics into ElasticFusion~\citep{whelan2015elasticfusion} which represents the environment as a dense surfel map. ElasticFusion is able to reconstruct the environment in real-time on a GPU given RGB-D images and can handle local as well as global loop closure. Semantic information is stored on a per-surfel basis. Inference is done by an RGB-D-CNN before fusing estimates probabilistically. 
SemanticFusion fuses for each visible surfel and all possible classes which is very time- and memory-consuming since the class probabilities need to be stored per surfel and class on the GPU.
Objects normally consist of a large number of surfels and share in reality a single class label even though semantic information within a surfel would only be required at the border of the object where the class is likely to change.
Hence, many surfels store the same redundant information and since GPU memory is notoriously limited, memory usage becomes a problem for larger surfel maps. Furthermore, SemanticFusion tends to create many unnecessary surfels with differing scales and labels for the same surface when sensed from different distances.

	Closely related to our approach is the work of \citet{valentin2013mesh}. Their map is represented as a triangular mesh. They aggregate depth images in a Truncated Signed Distance Function (TSDF) and obtain the explicit mesh representation via the marching cubes algorithm. Afterwards, semantic inference is performed for each triangle independently using a learned classifier on an aggregation of photometric (color dependent) and handcrafted local geometric (mesh related) features. Spatial regularization is ensured through a CRF over the mesh faces.
	Their classifier infers the label with all visible pixels per face at once and is not designed to incrementally fuse new information. Furthermore, the pairwise potential of the CRF does not take the likelihood for other classes in to account. Especially around object borders this may lead to suboptimal results. The semantic resolution is tied to the geometry of the mesh, hence to have fine details the mesh resolution needs to be fine grained. Geometrically a wall can be described with a small number of vertices and faces, but to semantically distinguish between an attached poster and the wall itself the mesh would need a high resolution.
	
	In comparison, we only store the likelihood for a small number of most probable classes and the meshing creates a single simplified surface while the texture resolution can be chosen independent of the geometry yet appropriate to the scene. During fusion we further include weighting to account for the sensor distance and in the case of color integration include vignetting and viewing angle.
	
	\section{Overview}
	In this paper, we present a novel approach to building semantic maps by decoupling the geometry of the environment from its semantics by using semantic textured meshes. This decoupling allows us to store the geometry of the scene as a lightweight mesh which efficiently represents even city-sized environments. 
	
	Our method (see \reffig{fig:overview}) operates in three steps: mesh generation, semantic texturing and label propagation.
	
	In the \emph{mesh generation} step, we create a mesh of the environment by aggregating the individual point clouds recorded by a laser scanner or an RGB-D camera. We assume that the scans are preregistered into a common reference frame using any off-the-shelf SLAM system. We calculate the normals for the points in each scan by estimating an edge-maintaining local mesh for the scan. Once the full point cloud equipped with normals is aggregated, we extract a mesh using Poisson reconstruction~\citep{kazhdan2013screened} and further simplify it using QSlim~\citep{garland1998qslim}.
	Our main contribution for 3D reconstruction is the proposal of system capable of fast normal estimation by using a local mesh and also local line simplification which heavily reduces the number of points, therefore reducing the time and memory used by Poisson reconstruction.

	In the \emph{semantic texturing} step, we first prepare the mesh for texturing by parameterizing it into a 2D plane. Seams and cuts are added to the mesh in order to deform it into a planar domain.
	A semantic texture is created in which the number of channels corresponds to the number of semantic classes. The semantic segmentation of each individual RGB frame is inferred by RefineNet and fused probabilistically into the semantic texture. We ensure bounded memory usage on the GPU by dynamically allocating and deallocating parts of the semantic texture as needed. Additionally, the RGB information is fused in an RGB texture.

	In the \emph{Label Propagation} step, we project the stable semantics, stored in the textured mesh, back into the camera frames and retrain the predictor in a semi-supervised manner using high confidence fused labels as ground truth, allowing the segmentation to learn from novel view points.

	Hence, the contribution presented in this article is fourfold:
	\begin{itemize}
		\item a scalable system for building accurate meshes from range measurements with coupled geometry and semantics at independent resolution,
		\item an edge-maintaining local mesh generation from lidar scans,
		\item a label propagation that ensures temporal and spatial consistency of the semantic predictions, which helps the semantic segmentation to learn and perform segmentation from novel view points,
		\item fast integration of probability maps by leveraging the GPU with bounded memory usage.
	\end{itemize}

	\section{Notation}
	In the following, we will denote matrices by bold uppercase letters and (column-)vectors with bold lowercase letters.
	The rigid transformation $\mathbf{T}_{F_2F_1}$ is represented as $4\times4$ matrix and maps points from coordinate frame $F_1$ to coordinate frame $F_2$ by operating on homogeneous coordinates. When necessary, the frame in which a point is expressed is added as a subscript: e.g. $\mathbf{p}_w$ for points in world coordinates.
	A point $\mathbf{p}_w$ is projected into frame $F$ with the pose $\mathbf{T}_F$ and the camera matrix $\mathbf{K}_F\in\mathbb{R}^{3\times 3}$. For the camera matrix, we assume a standard pinhole model with focal length $f_x, f_y$, and principal point $c_x, c_y$. The projection  of $\mathbf{p}_w$ into image coordinates $\mathbf{u}=(u_x,u_y)^\intercal_F \in \mathbf{\Omega} \subset \mathbb{R}^2$ is given by the following mapping:
	\begin{align}
	g_F(\mathbf{p}_w) &: \mathbf{p}_w \rightarrow \mathbf{p}_F, \\
	(\mathbf{p}_F, 1)^\intercal &= \mathbf{T}_{F_w} \cdot (\mathbf{p}_w, 1)^\intercal, \\
	\pi_F(\mathbf{p}_F) &: \mathbf{p}_F \rightarrow \mathbf{u}_F, \\
	(x,y,z)^\intercal_F &= \mathbf{K}_F \cdot \mathbf{p}_F, \\
	\mathbf{u}_F &= (x/z, y/z)^\intercal.
	\end{align}
	An image or a texture is denoted by $I(\mathbf{u}): \mathbf{\Omega} \to \mathbb{R}^n$, where $\mathbf{\Omega} \subset \mathbb{R}^2$ maps from pixel coordinates $\mathbf{u}=(u_x,u_y)^\intercal$ to $n$-channel values.
	\section{Method}
	The input to our system is a sequence of organized point clouds $\{ \mathcal{P}^t \}$\footnote{An organized point cloud exhibits an image resembling structure, e.g. from commodity RGB-D sensors.}, and RGB images $\{ I^t \}$ ($t$ indicates the time step).
	We assume that the point clouds are already registered into a common reference frame, and the extrinsic calibration $\mathbf{T}_{cd}$ from depth sensor to camera, as well as camera matrices, are given. The depth sensor can be an RGB-D camera or a laser scanner.
	The output of our system is threefold:
	\begin{itemize}
		\item a triangular mesh of the scene geometry, defined as a tuple $\mathcal{M} = (\mathcal{V}, \mathcal{F})$ of vertices $\mathcal{V}$ and faces $\mathcal{F}$. Each vertex $\in (\mathbb{R}^3 \times \mathbb{R}^2)$ contains a 3D point and a UV texture coordinate, while the mesh face is represented by the indices $\in\mathbb{N}^3$ of the three spanning vertices within $\mathcal{V}$. 
		\item a semantic texture $S$ indicating the texels class probabilities,
		\item an RGB texture $C$ representing the surface appearance.
	\end{itemize}
	After describing the necessary depth preprocessing in \refsec{sec:DepthPro}, we will explain in detail the mesh generation and parametrization~(\refsec{sec:Mesh}), before elaborating on the semantic~(\refsec{sec:Semantic}) and color integration~(\refsec{sec:Color}), sparse representation~(\refsec{sec:SparseSemantic}) and label propagation (\refsec{sec:LabelPropagation}).

	\subsection{Depth Preprocessing}\label{sec:DepthPro}
	As previously mentioned, our system constructs a global mesh from the aggregation of a series of point clouds $\{ \mathcal{P}^t \}$ recorded from a depth sensor. Many surface reconstruction algorithms require accurate per-point normal. One way to obtain these normals is by aggregating the full global point cloud, and using the k-nearest-neighbors to estimate the normals for each point. However, this would be prohibitively slow as it requires a spatial subdivision structure, like a Kd-tree, which can easily grow to a considerable size for large point clouds, limiting the scalability of the system.
	For fast normal estimation we take advantage of the structure of the recorded point cloud. Since depth from an RGB-D sensor is typically structured as an image, we can easily query adjacent neighboring points. Similarly, rotating lidar sensors can produce organized scans. A complete revolution of e.g. a Velodyne VLP-16 produces a 2D array of size $\num{16}\times N$ containing the measured range of each recorded point, where $N$ is determined by the speed of revolution of the laser scanner.	
	
	\newcommand\localmeshLeftTrim{400bp}
	\newcommand\localmeshrBottomTrim{270bp}
	\newcommand\localmeshRightTrim{440bp}
	\newcommand\localmeshTopTrim{300bp}
	\begin{figure}[]
		\captionsetup[subfloat]{labelformat=empty}
		\centering
		\subfloat[a) Point cloud from lidar scanner]{
			\includegraphics[trim=\localmeshLeftTrim{} \localmeshrBottomTrim{} \localmeshRightTrim{} \localmeshTopTrim{},clip,width=0.25\textwidth] {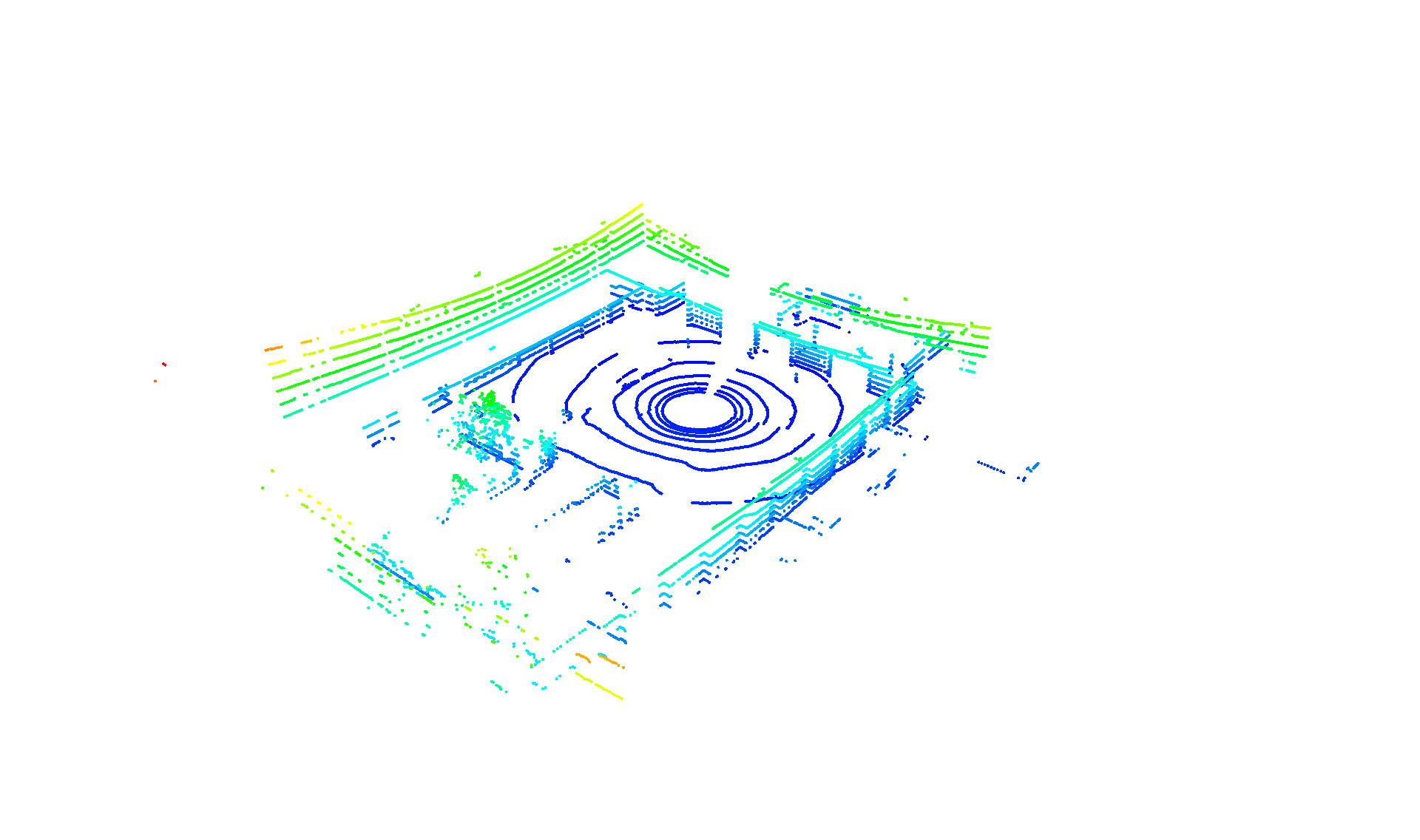}
		}
		\subfloat[b) Reconstructed local mesh]{
			\includegraphics[trim=\localmeshLeftTrim{} \localmeshrBottomTrim{} \localmeshRightTrim{} \localmeshTopTrim{},clip,width=0.25\textwidth] {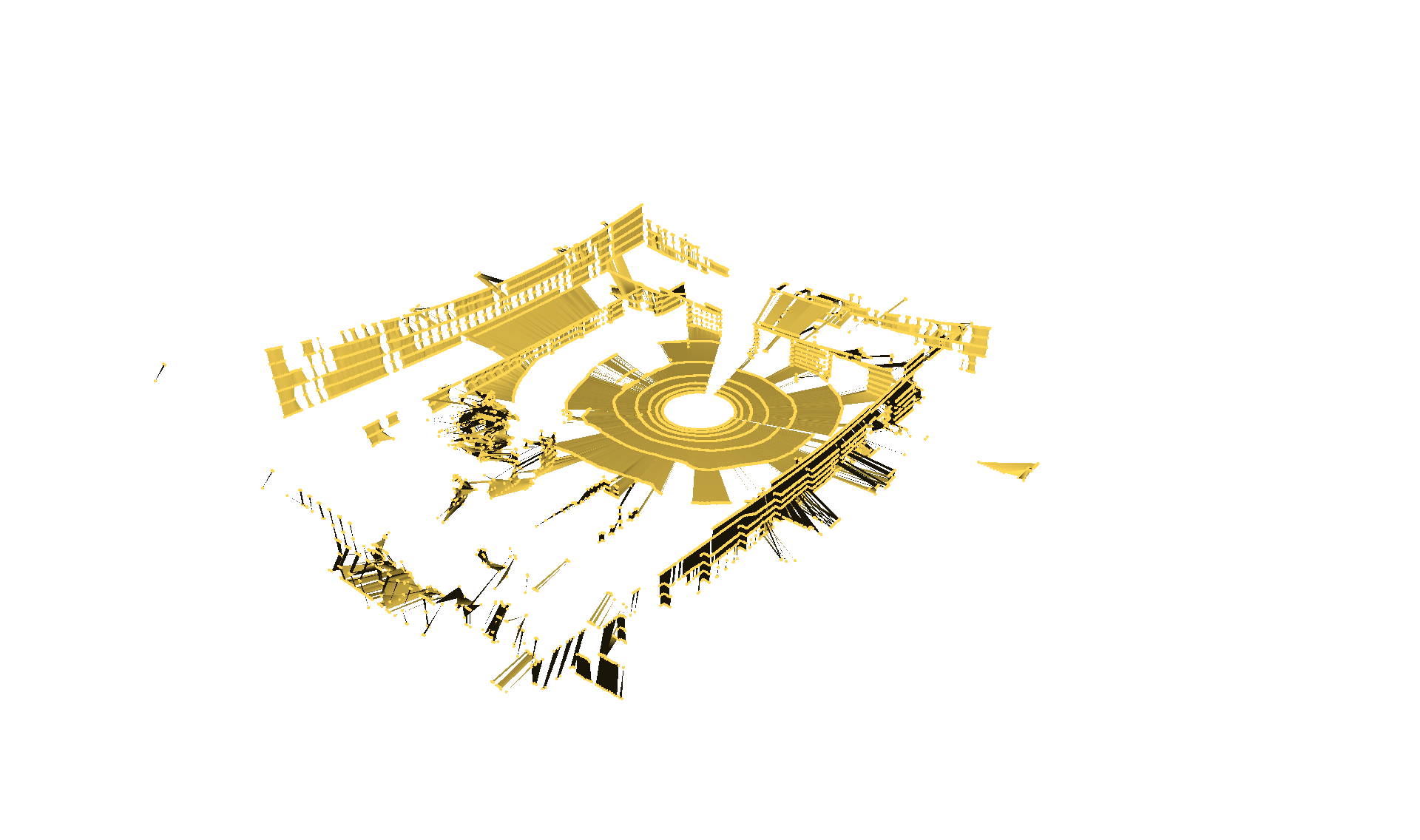}
		}
		\caption{Local mesh: We reconstruct an approximate local mesh from the given range measurements in order to estimate point normals needed for Poisson reconstruction.}
		\label{fig:LocalMesh}
	\end{figure}
	
	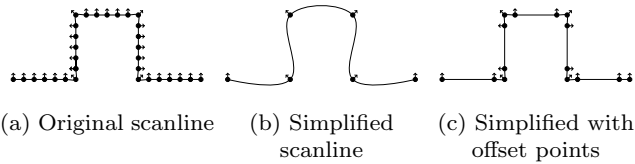
\begin{figure}[b]
		\captionsetup[subfigure]{justification=centering}

		\subfloat[Original scanline]{
			\resizebox {0.15\textwidth} {!} {
				\begin{tikzpicture}[baseline]
				\begin{axis}[xmin=-0.5,xmax=9.5,ymin=-1.0,ymax=7,axis line style={draw=none}, axis lines=none]
				\addplot[smooth, tension=0.5,mark=*] table [x=x,y=y,meta=label] {./data_dense.txt};

				\DTLsetseparator{ }%
				\begin{filecontents*}{./data_dense.txt}
   nodes     x         y       label nx ny
   1.0000 0.0   0.0  a 0.0 1.0
   1.0000 0.5   0.0  a 0.0 1.0
   1.0000 1.0   0.0  a 0.0 1.0
   1.0000 1.5   0.0  a 0.0 1.0
   1.0000 2.0   0.0  a 0.0 1.0
   1.0000 2.5   0.0  a 0.0 1.0
   2.0000 3.0   0.0 a -0.7 0.7
   2.0000 3.0   0.5 a -1.0 0.0
   2.0000 3.0   1.0 a -1.0 0.0
   2.0000 3.0   0.5 a -1.0 0.0
   2.0000 3.0   1.5 a -1.0 0.0
   2.0000 3.0   2.0 a -1.0 0.0
   2.0000 3.0   2.5 a -1.0 0.0
   3.0000 3.0   3.0 a -0.7 0.7
   3.0000 3.5   3.0 a 0.0 1.0
   3.0000 4.0   3.0 a 0.0 1.0
   3.0000 4.5   3.0 a 0.0 1.0
   3.0000 5.0   3.0 a 0.0 1.0
   3.0000 5.5   3.0 a 0.0 1.0
   4.0000 6.0   3.0 a 0.7 0.7
   4.0000 6.0   2.5 a 1.0 0.0
   4.0000 6.0   2.0 a 1.0 0.0
   4.0000 6.0   1.5 a 1.0 0.0
   4.0000 6.0   1.0 a 1.0 0.0
   4.0000 6.0   0.5 a 1.0 0.0
   5.0000 6.0   0.0 b 0.7 0.7
   5.0000 6.5   0.0 b 0.0 1.0
   5.0000 7.0   0.0 b 0.0 1.0
   5.0000 7.5   0.0 b 0.0 1.0
   5.0000 8.0   0.0 b 0.0 1.0
   5.0000 8.5   0.0 b 0.0 1.0
   6.0000 9.0    0.0 b 0.0 1.0
  
\end{filecontents*}
\DTLloaddb[noheader=false]{data_dense_file}{./data_dense.txt}
				\DTLforeach*{data_dense_file}{\xP=x, \yP=y, \xN=nx, \yN=ny}{%
					\edef\temp{\noexpand\draw [->] (\xP,\yP) -- (\xP+\xN*0.3,\yP+\yN*0.3);}\temp
				}

				\end{axis}
				\end{tikzpicture}
			}
		}
		\subfloat[Simplified scanline]{

			\resizebox {0.15\textwidth} {!} {
				\begin{tikzpicture}[baseline]
				\begin{axis}[xmin=-0.5,xmax=9.5,ymin=-1.0,ymax=7,axis line style={draw=none}, axis lines=none]
				\addplot[smooth, tension=0.7,mark=*] table [x=x,y=y,meta=label] {./data_sparse.txt};

				\DTLsetseparator{ }%
				\begin{filecontents*}{./data_sparse.txt}
   nodes  x         y       label nx ny
   1.0000 0.0 0.0 a 0.0 1.0
   2.0000 3.0 0.0 a -0.7 0.7
   3.0000 3.0 3.0 a -0.7 0.7
   4.0000 6.0 3.0 a 0.7 0.7
   5.0000 6.0 0.0 b 0.7 0.7
   6.0000 9.0 0.0 b 0.0 1.0
  
\end{filecontents*}
\DTLloaddb[noheader=false]{data_sparse_file}{./data_sparse.txt}
				\DTLforeach*{data_sparse_file}{\xP=x, \yP=y, \xN=nx, \yN=ny}{%
					\edef\temp{\noexpand\draw [->] (\xP,\yP) -- (\xP+\xN*0.3,\yP+\yN*0.3);}\temp
				}

				\end{axis}
				\end{tikzpicture}
			}
		}
		\subfloat[Simplified with offset points]{
			\resizebox {0.15\textwidth} {!} {
				\begin{tikzpicture}[baseline]
				\begin{axis}[xmin=-0.5,xmax=9.5,ymin=-1.0,ymax=7,axis line style={draw=none}, axis lines=none]
				\addplot[smooth, tension=0.3,mark=*] table [x=x,y=y,meta=label] {./data_protected.txt};

				\DTLsetseparator{ }%
				\begin{filecontents*}{data_protected.txt}
   nodes     x         y       label nx ny
   1.0000    0.0   0.0  a 0.0 1.0
   1.0000    2.5   0.0  a 0.0 1.0
   2.0000     3.0   0.0 a -0.7 0.7
   2.0000     3.0   0.5 a -1.0 0.0
   2.0000     3.0   2.5 a -1.0 0.0
   3.0000     3.0   3.0 a -0.7 0.7
   3.0000     3.5   3.0 a 0.0 1.0
   3.0000     5.5   3.0 a 0.0 1.0
   4.0000     6.0   3.0 a 0.7 0.7
   4.0000     6.0   2.5 a 1.0 0.0
   4.0000     6.0   0.5 a 1.0 0.0
   5.0000     6.0   0.0 b 0.7 0.7
   5.0000     6.5   0.0 b 0.0 1.0
   5.0000     8.5   0.0 b 0.0 1.0
   6.0000    9.0    0.0 b 0.0 1.0
  
\end{filecontents*}
\DTLloaddb[noheader=false]{data_protected_file}{data_protected.txt}
				\DTLforeach*{data_protected_file}{\xP=x, \yP=y, \xN=nx, \yN=ny}{%
					\edef\temp{\noexpand\draw [->] (\xP,\yP) -- (\xP+\xN*0.3,\yP+\yN*0.3);}\temp
				}
				\end{axis}
				\end{tikzpicture}
			}
		}
		\caption{Line simplification:
The original scan line (left) is excessively dense in planar areas.
The original simplification greatly reduces the number of points but creating a global surface using a method like Poisson reconstruction overly smoothes the edges (middle). Our extension simplifies the line and preserves hard edges by adding further constraints which allow Poisson reconstruction to maintain sharp features (right).}
		\label{fig:LineSimplification}
	\end{figure}
	
	\newcommand{\offsetElemWidth}{4.2cm}
	\renewcommand{\tabcolsep}{1pt}
	\newcommand\columnLeftTrim{0bp}
	\newcommand\columnBottomTrim{0bp}
	\newcommand\columnRightTrim{0bp}
	\newcommand\columnTopTrim{0bp}
	\fboxsep=0mm
	\fboxrule=1pt
	\definecolor{blue_square}{RGB}{100 150 250}
	\begin{figure*}
		\centering
		\begin{tabular}{>{\centering\arraybackslash}m{2.0cm}ccc}
			& Point cloud&Poisson mesh&Column from above \\
			
			No offset points &
			\includegraphics[align=c,
			width=\offsetElemWidth{}]{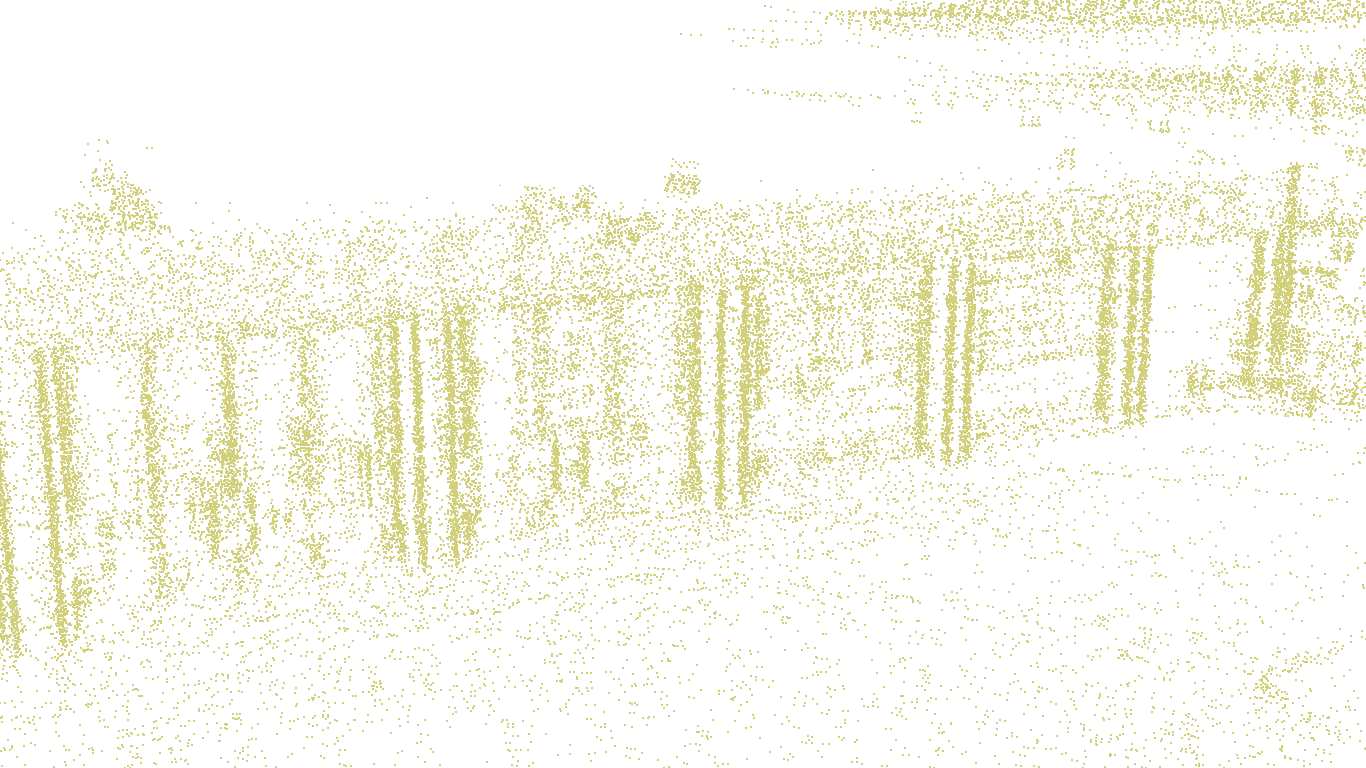}& 
			\includegraphics[align=c,
			width=\offsetElemWidth{}]{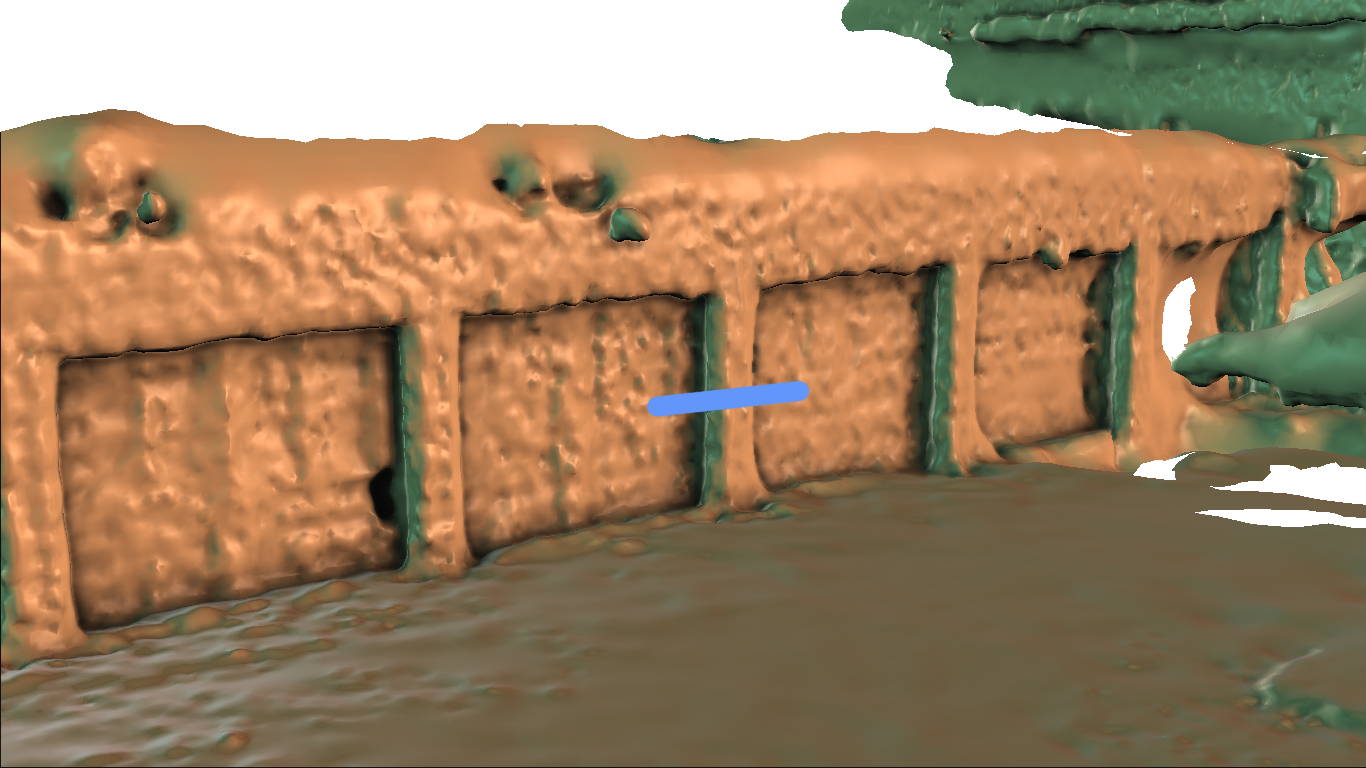}&
			\fcolorbox{blue_square}{white}{				
				\includegraphics[align=c,
				width=\offsetElemWidth{}, trim=\columnLeftTrim{} \columnBottomTrim{} \columnRightTrim{} \columnTopTrim{},clip]{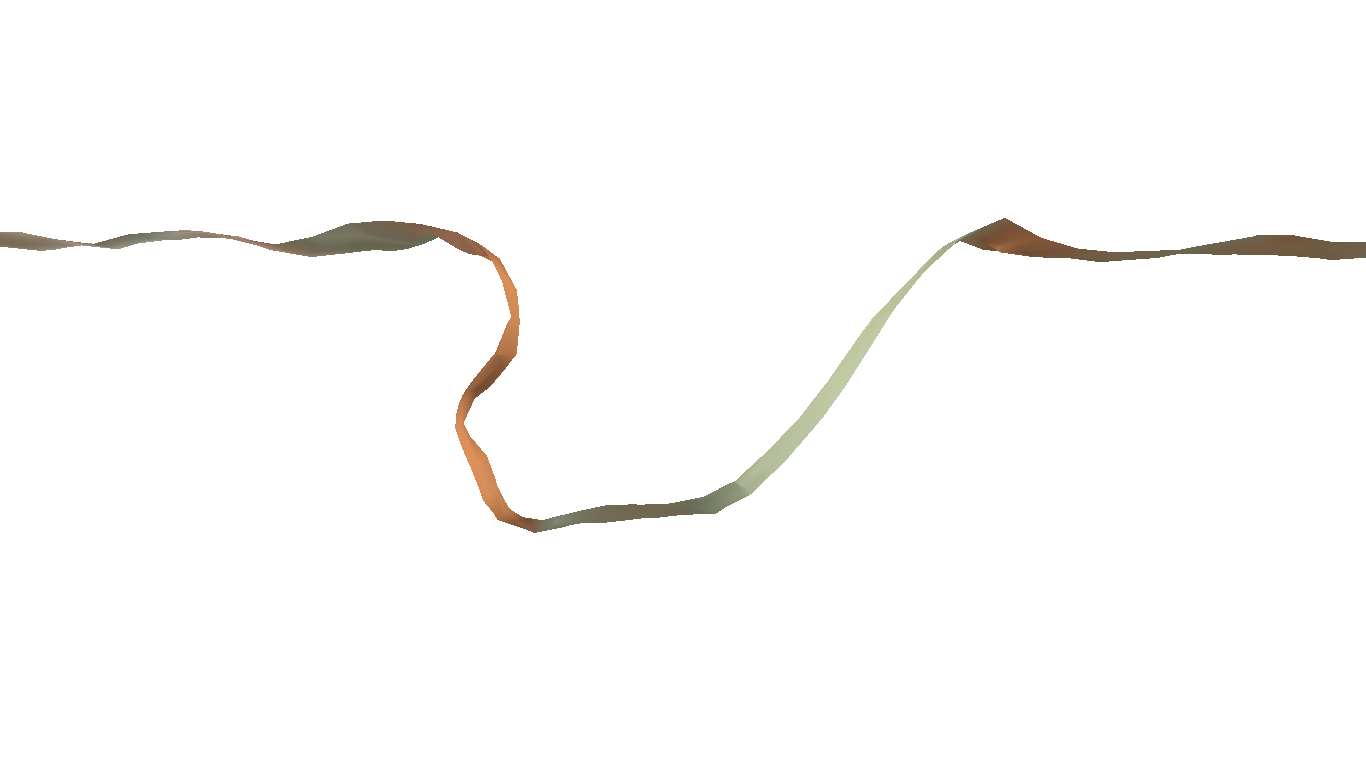}
			}
			\\ 
			
			With offset points &
			\includegraphics[align=c,
			width=\offsetElemWidth{}]{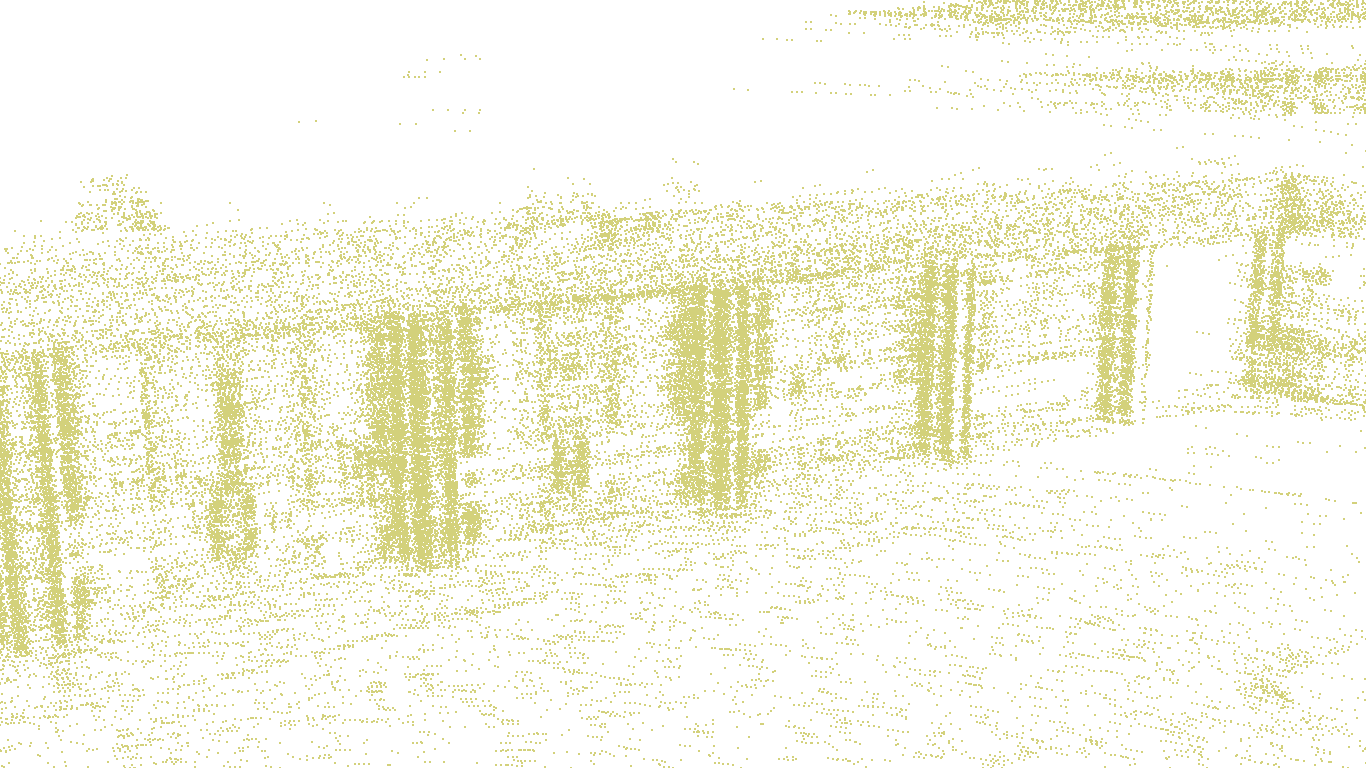}& 
			\includegraphics[align=c,
			width=\offsetElemWidth{}]{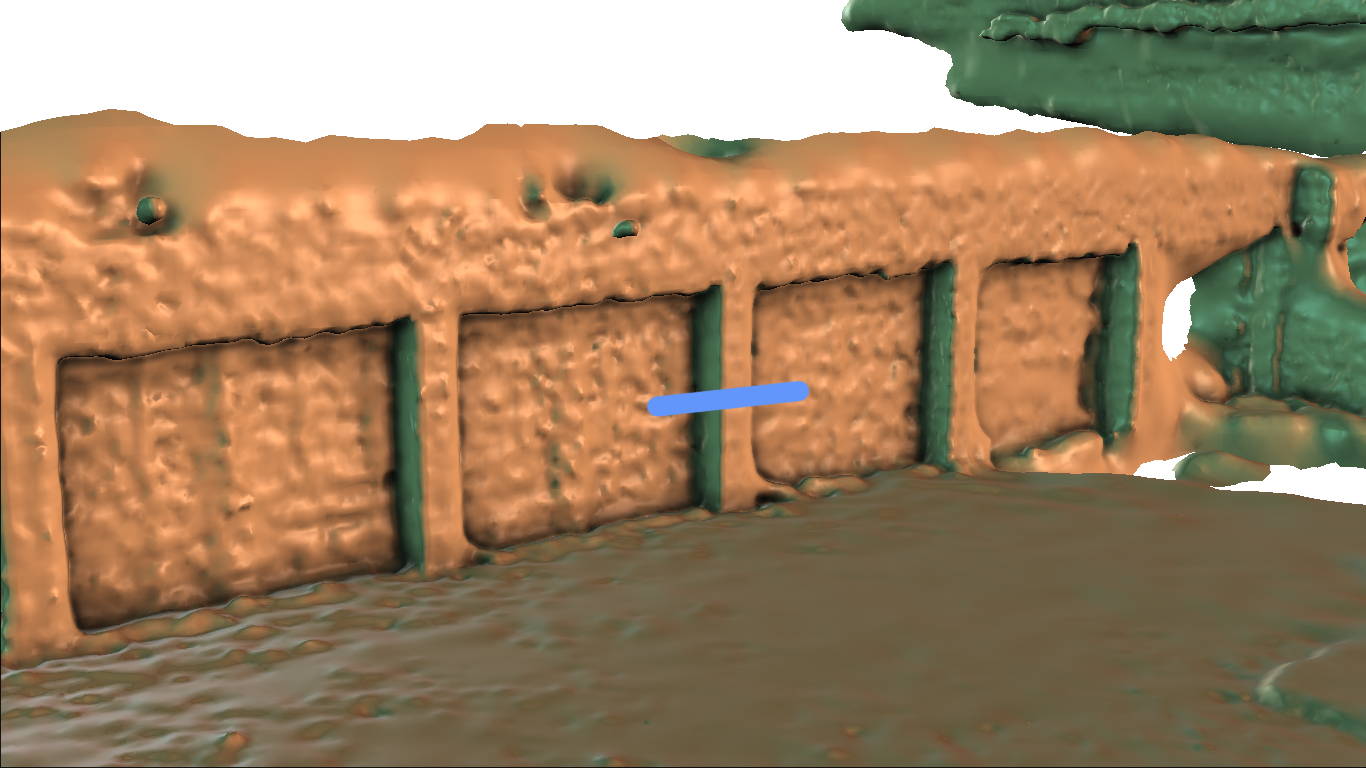}&
			\fcolorbox{blue_square}{white}{
				\includegraphics[align=c,
				width=\offsetElemWidth{}, trim=\columnLeftTrim{} \columnBottomTrim{} \columnRightTrim{} \columnTopTrim{},clip]{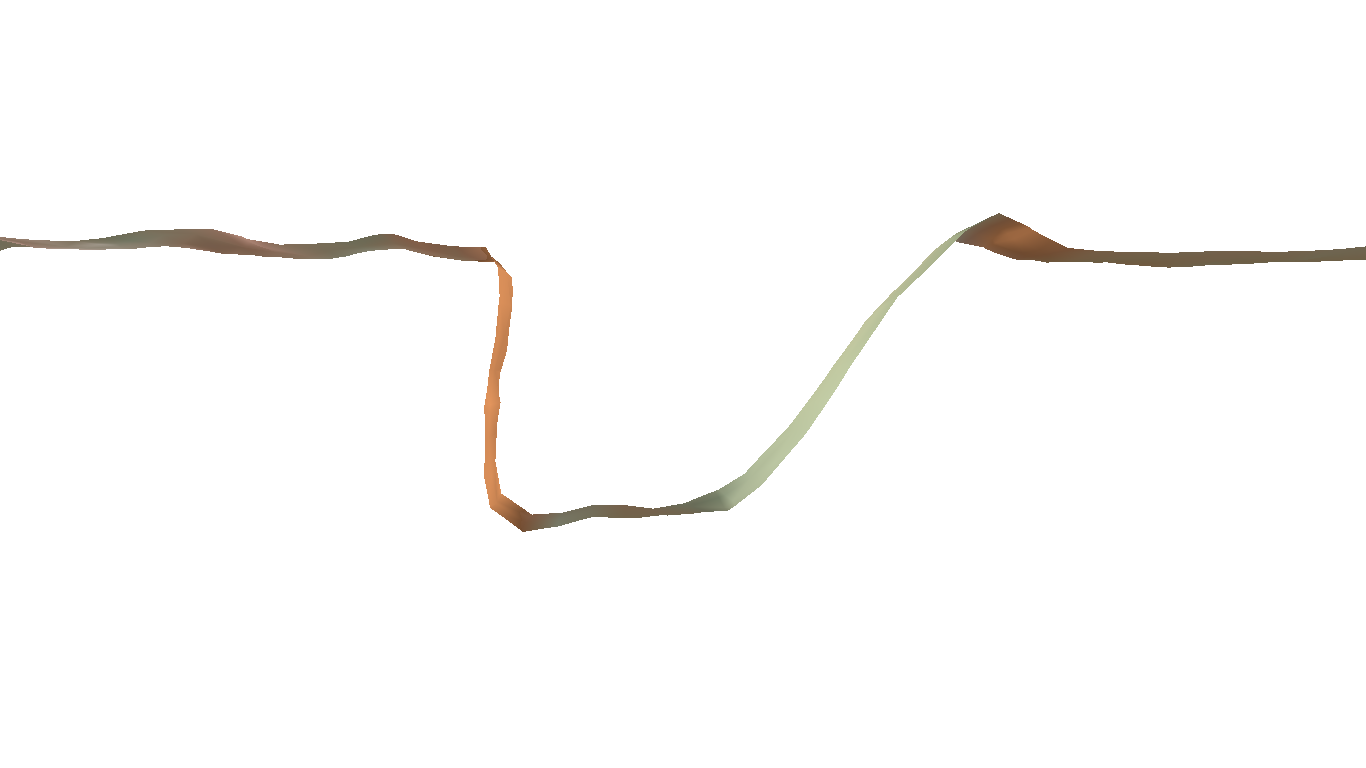}
			}
			\\ 
	
		\end{tabular}
		
		\caption{Offset points: The impact of adding offset points during line simplification is evaluated on the courtyard dataset. The absence of offset points leads to a Poisson reconstruction that is overly smooth (first row). Adding offset points increases the density of the point cloud around the edges of the columns and results in a sharper reconstruction (second row). The smooth right hand side of the columns is due to undersampling from occlusion.}
		\label{fig:MeshOffsetPoints}
	\end{figure*}
	
	\newcommand\spikeslocalmeshLeftTrim{400bp}
	\newcommand\spikeslocalmeshrBottomTrim{210bp}
	\newcommand\spikeslocalmeshRightTrim{440bp}
	\newcommand\spikeslocalmeshTopTrim{350bp}
	\begin{figure*}[]
		\captionsetup[subfigure]{justification=centering}
		\captionsetup[subfloat]{labelformat=empty}
		\centering
		\subfloat[a) Point cloud from lidar scanner]{
			\includegraphics[trim=\spikeslocalmeshLeftTrim{} \spikeslocalmeshrBottomTrim{} \spikeslocalmeshRightTrim{} \spikeslocalmeshTopTrim{},clip,width=0.18\textwidth] {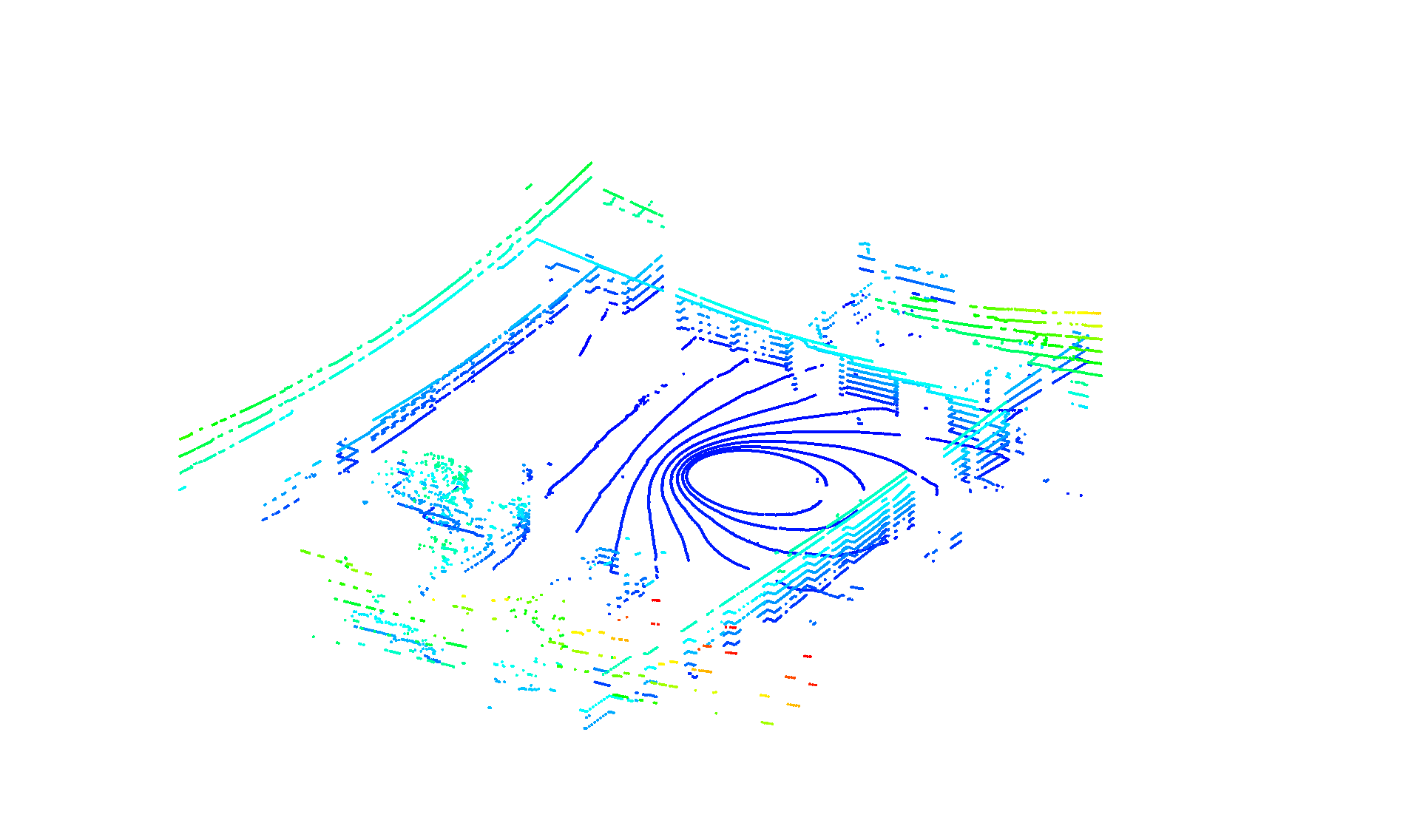}
		}
		\subfloat[b) Local mesh]{
			\includegraphics[trim=\spikeslocalmeshLeftTrim{} \spikeslocalmeshrBottomTrim{} \spikeslocalmeshRightTrim{} \spikeslocalmeshTopTrim{},clip,width=0.18\textwidth] {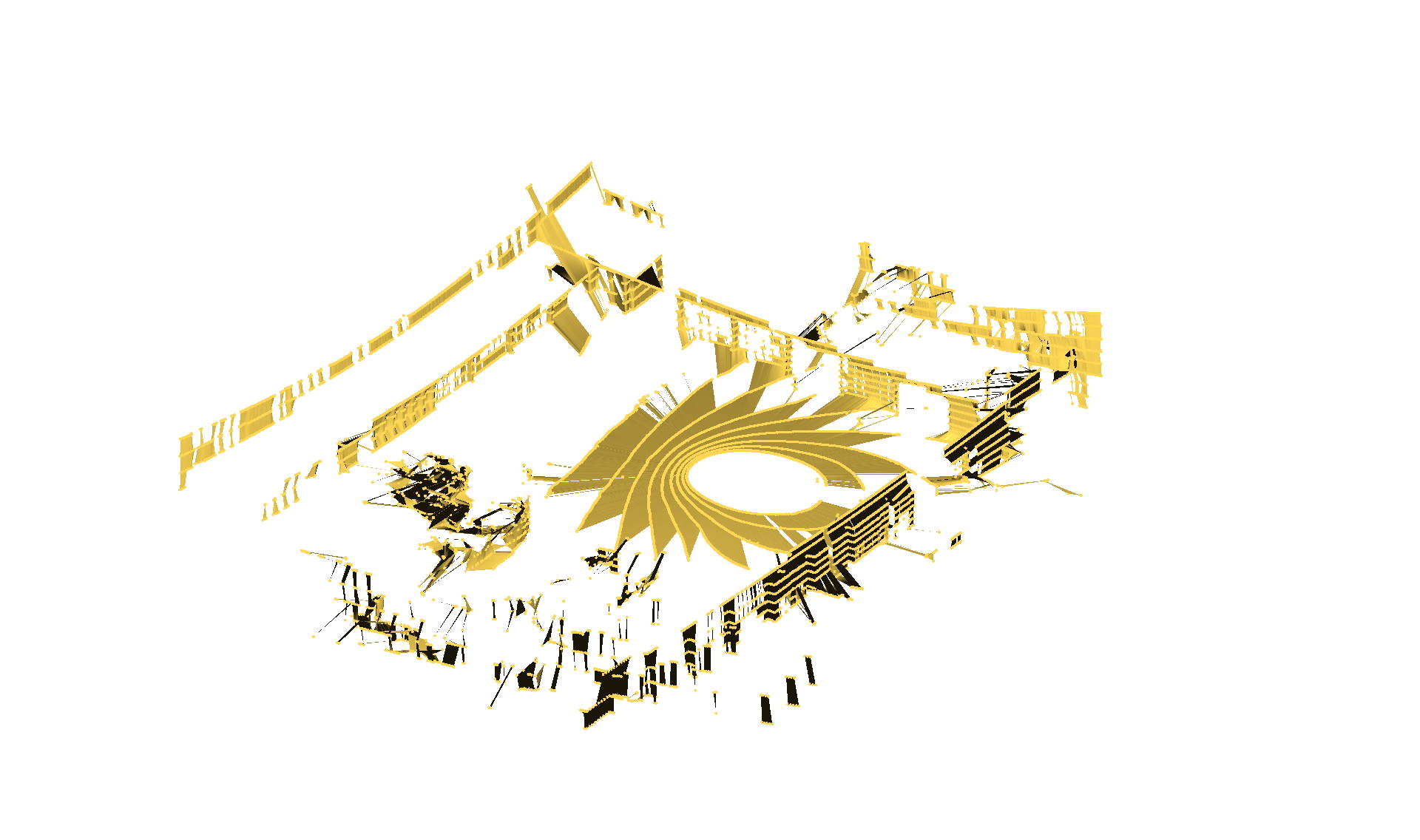}
		}
		\subfloat[c) Top view of the\newline ground plane]{
			\includegraphics[width=0.18\textwidth] {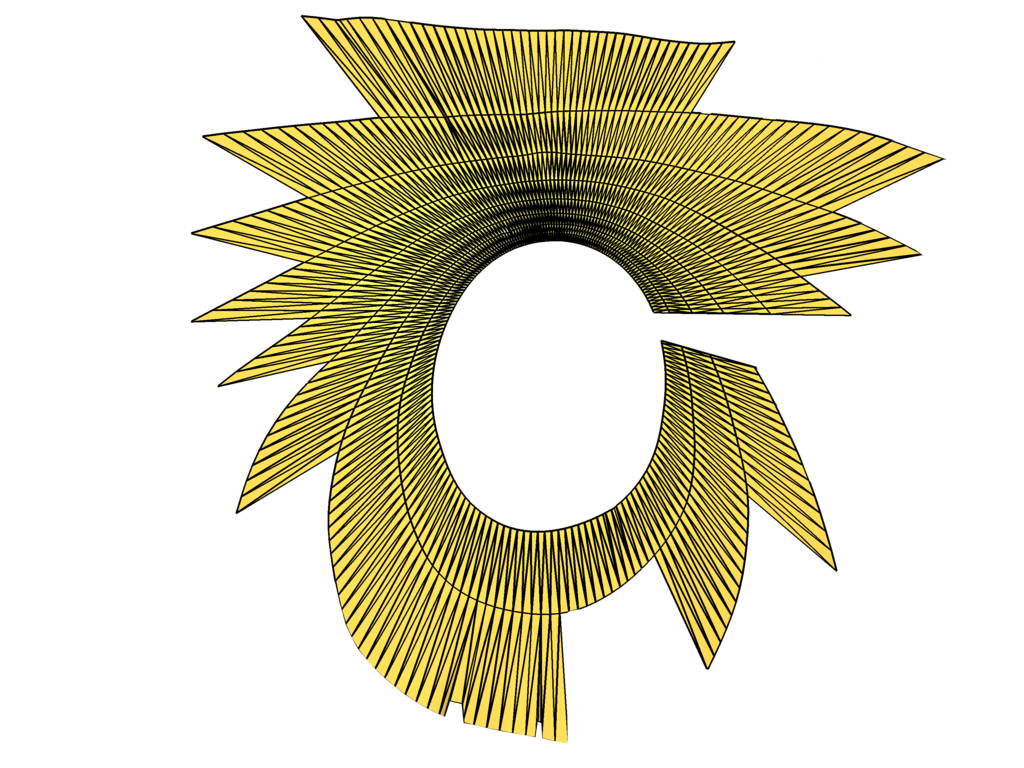}
		}
		\subfloat[d) After edge-flipping]{
			\includegraphics[width=0.18\textwidth] {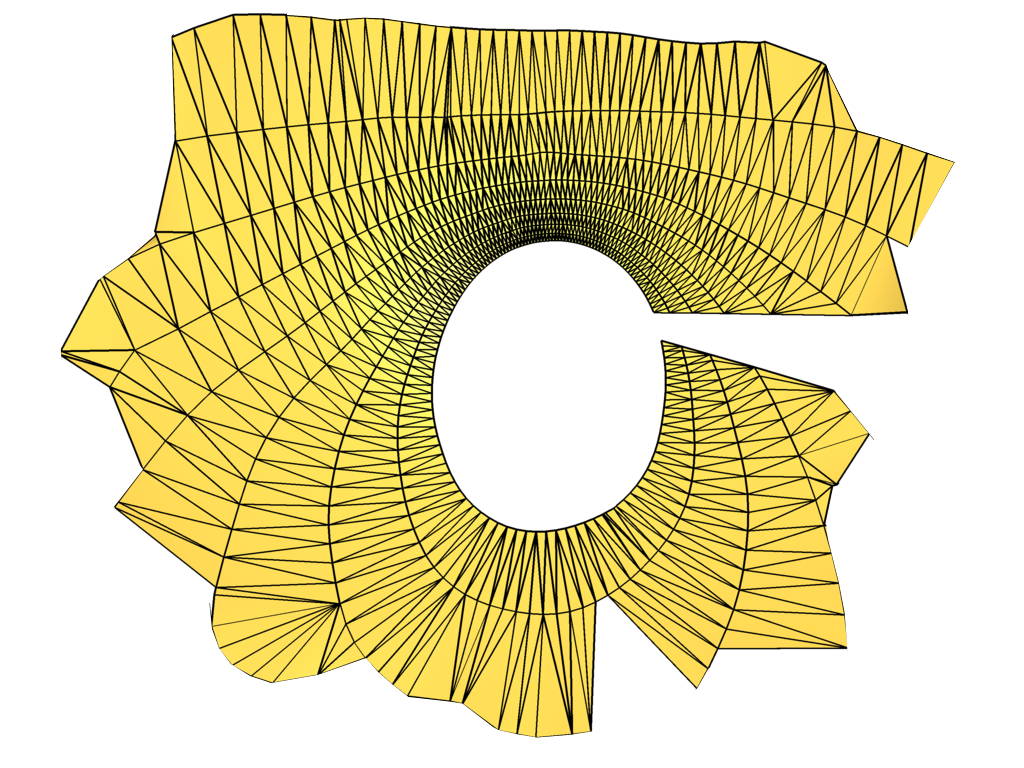}
		}
		\subfloat[e) After edge-flipping\newline and simplification]{
			\includegraphics[width=0.18\textwidth] {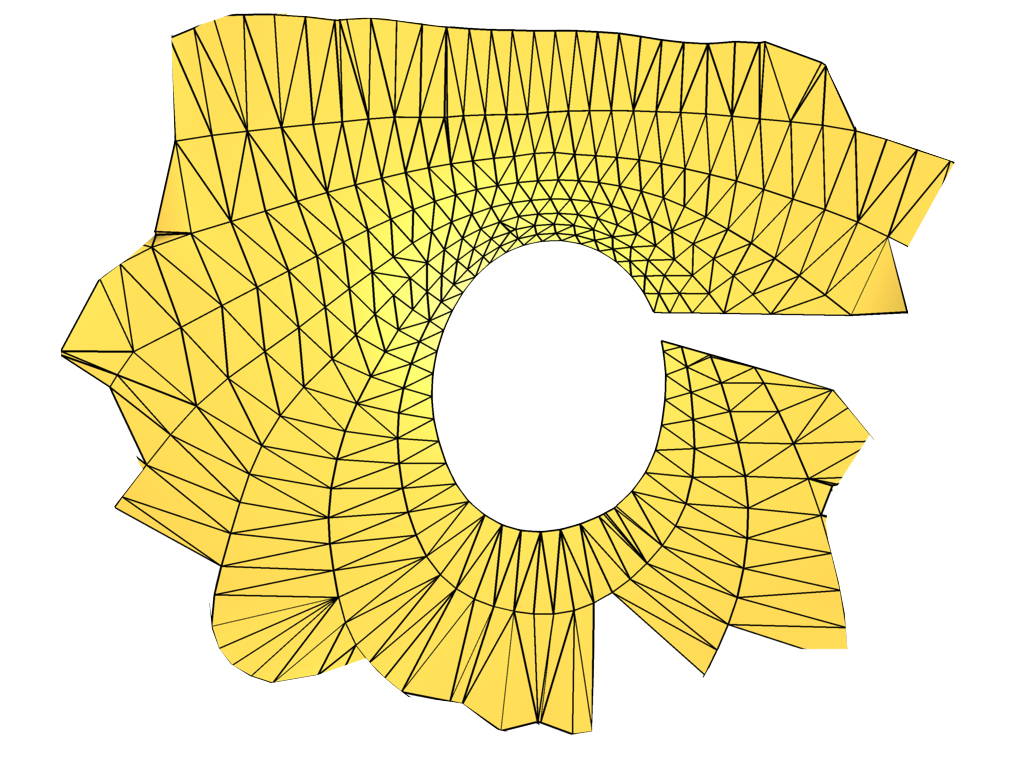}
		}
		\caption{Local mesh connections problem: During sudden movements of the laser scanner, the scan rings are compressed behind and expanded in front of the sensor. This creates many small and steep triangles which degrades normal estimation. We perform iterative edge-flipping in order to connect each vertex with their closest neighboring vertex, hence, improving the likelihood for estimating correct normals. Furthermore, we apply line simplification to each scan ring independently for data reduction without sacrificing mesh fidelity.}
		\label{fig:LocalMeshSpikes}
	\end{figure*}
	
	Given the organized structure, we could create a triangular local mesh with approximate normals as introduced by \cite{holz2014registration}. However, this would imply using all aggregated points despite the fact that a significant number of them are redundant, since they lay on a common plane.
	Hence, we propose a method for fast normal estimation and point cloud reduction which helps to reconstruct the mesh in a fast and memory efficient manner (\reffig{fig:LocalMesh}). We first simplify each individual scan to obtain a reduced point cloud without sacrificing geometrical fidelity. For that, we start with the line simplification algorithm of Ramer-Douglas-Peucker~\citep{douglas1973RamerDouglasPeucker} which is applied on each scan ring. To maintain hard edges of the point cloud, we extend Ramer-Douglas-Peucker to create offset points around the simplified edges to add a further constraint on the normals of the points, and allow the subsequent mesh generator to recover sharp features as visualized in \reffig{fig:LineSimplification} and \reffig{fig:MeshOffsetPoints}.
	The local mesh is created by unwrapping the scan in 2D using polar coordinates, and performing a constrained 2D Delaunay triangulation. Constrained edges are set to the ones obtained from the line simplification. This ensures that points that lie on the same scan ring will be connected together by triangles.
	After recovering a local mesh from the point cloud, normals are first estimated per face by calculating the cross product between two of the edges, and finally per vertex using a \emph{Mean Weighted by Angle} (MWA) scheme \citep{thurrner1998computing} which weighs each triangle's contribution by the angle under which it is incident to the vertex:
	\begin{gather}
	\begin{aligned}
	\vec{n}_f &= \frac{ \mathbf{e}_{1,2} \times \mathbf{e}_{3,2} }{  \norm { \mathbf{e}_{1,2} \times \mathbf{e}_{3,2} } }, \\
	\vec{n}_v &= \sum_{f \in AdyF(v)}^n \alpha_f \cdot \vec{n}_f,\\
	\vec{n}_v &= \vec{n}_v / \norm{\vec{n}_v}.
	\end{aligned}
	\end{gather}
	where $\vec{n}_f$ and $\vec{n}_v$ denote the face and vertex normals, respectively, and $\alpha_f$ is the angle between the two edge vectors $\mathbf{e}_1$ and $\mathbf{e}_2$ that share the vertex. The angle between the two edge vectors can be computed using the dot product between them: $ \alpha_f= \arccos \left(\frac{\mathbf{e}_1 \cdot \mathbf{e}_2 }{ \norm{\mathbf{e}_1} \norm{\mathbf{e}_2}  } \right)$.

	Until now, we have obtained fast normals using only the points contained in one revolution of the Velodyne. However, due to the anisotropy in the sampling of a laser scanner, the connections between points created by the Delaunay triangulation in 2D may not be optimal when lifted to 3D, as seen in \reffig{fig:LocalMeshSpikes}. Since the connections between points are crucial for an accurate normal estimation, an iterative local mesh refinement is performed. The refinement ensures that each point will be connected to the neighbors that lie spatially close in 3D. Again, in order to avoid cumbersome and slow spatial subdivision structures we introduce an edge flipping algorithm which iteratively \emph{flips} the edge shared between two triangles to increase the following quality measure:
	\begin{equation}
	q=\frac{4a\sqrt{3}}{h_1^2 + h_2^2 + h_3^2},
	\end{equation}
	where $a$ denotes the area of the triangle and $h$ is the length of the edge. We chose it such that it is monotonically increasing, and promotes more equilateral triangles.
	We perform edge flipping in a greedy fashion by choosing first the triangle which will experience the most quality increase. After performing the flip for a triangle, the quality of adjacent triangles may change and is updated. We continue flipping edges until the quality measure can no longer be increased.

	\subsection{Mesh Generation}\label{sec:Mesh}
	After aggregation of the points with corresponding normals from all simplified scans, we perform Poisson reconstruction to recover a high quality mesh, despite having potentially noisy data and missing measurements. The resulting mesh can still be overly dense in areas which are geometrically simple, like the ground. Hence, we apply a second global simplification step following the QSlim method~\citep{garland1998qslim}. This approach iteratively collapses edges until a certain error threshold, or a predetermined number of faces, is reached.

	The last step in the creation of the global mesh is to prepare it for texturing by parameterizing the mesh in the 2D domain in order to obtain UV coordinates for the vertices. For that, we make use of the \emph{UV smart project} function provided within Blender\footnote{\url{https://www.blender.org}}.
	\subsection{Semantic Integration}\label{sec:Semantic}
	In this section we detail our approach on how to update the global semantic texture $S$ using individual color images $I^t$.
	For semantic segmentation, we retrain RefineNet on the Mapillary dataset~\citep{neuhold2017mapillary} for street level segmentation. The dataset contains \num{25000} images densely labeled with \num{66} classes.
	Given the input image $I_k$, the output of the predictor can be interpreted as a per-pixel probability over all the class labels $P(O_\mathbf{u}=l_i| I_k)$, with $\mathbf{u}$ denoting pixel coordinates. 
	One common approach to integrate semantic information is to perform a Bayesian update over the classes probability, fusing new observations into the global belief for the semantic labels.
	However, this scheme of updating becomes slow for a large number of classes since the belief for all labels needs to updated.
	
	In our approach we choose to approximate the probability over the classes with only the \emph{argmax} probability. Hence, a new observation will consist of only the \emph{argmax} label and its probability instead of the full distribution. This enables us to use a fast integration scheme whose runtime is independent of the number of classes. We observe that in practice this approximation works well.
	
	We define the best class $L$ and its corresponding probability $P^*$ using:
	\begin{gather}
	\begin{aligned}
	L&=\argmax_c P(O_u=l_i| I_k), \\
	P^*&=\max_c P(O_u=l_i| I_k).
	\end{aligned}
	\end{gather}
	We perform a visibility check prior to updating the global semantic texture using individual segmentation results.
	Inspired by \emph{shadow mapping} techniques in computer graphics, we first render a depth map $D$ from the current camera view. In order to ensure that every texel is checked for visibility, we obtain for each texel $x$ with UV coordinates $\mathbf{u}_x$ the 3D point $\mathbf{p}_x$ from the vertices of the corresponding mesh face by barycentric interpolation.
	The point $\mathbf{p}_x$ is then projected into the current view, and discarded if the depth $d_x$ is larger than the stored value within the depth map $D(\pi(g_F(\mathbf{p}_x)))$, as it lies behind the visible part of the mesh. To indicate visibility, we use a per texel indicator variable $r_x\in\left\lbrace 0,1\right\rbrace$:
	\begin{align}
	r_{x_i} &=  
	\begin{cases}
	1,				& \text{if } d_x \geq D(\pi(g_F(x_w)))\\
	0,              & \text{otherwise}
	\end{cases}.
	\end{align}
	All remaining texels $(r_x > 0)$ are fused with the current segmentation result by increasing the probability of the obtained classes:
	\begin{align}
	S(\mathbf{u}_x,l)^{t} &= S(\mathbf{u}_x)^{t-1} + r_{x_i}\cdot w_{x_i}\cdot p_{x_i}, \\
	W(\mathbf{u}_x)^{t} &= W(\mathbf{u}_x)^{t-1} + r_{x_i}\cdot w_{x_i}, \\
	l_{x_i} &=  L (\pi(g_F(\mathbf{p}_x))),  \\
	p_{x_i} &= P^* (\pi(g_F(\mathbf{p}_x))).
	\end{align}
	Additionally, we weigh the fused probability by the faces distance from the camera under the assumption that pixels are more difficult to recognize from farther away, due to the low resolution of semantic segmentation.
	\begin{align}
	w_{x_i} &=
	\begin{cases}
	1,				& \text{if } d_x \leq d_{min}\\
	1 - \frac{d_x - d_{min}} {d_{max} - d_{min}} ,              & \text{otherwise}
	\end{cases},
	\end{align}
	where $d_x$ denotes the depth of the current texel, and $d_{min}$ and $d_{max}$ are thresholds for the distance which define a linear fall-off for the weight. In our experiments we set them to \SI{30}{\metre} and \SI{100}{\metre}, respectively.

	\newcommand\sparseLeftTrim{800bp}
	\newcommand\sparseBottomTrim{400bp}
	\newcommand\sparseRightTrim{800bp}
	\newcommand\sparseTopTrim{400bp}
	\begin{figure*}[]
		\captionsetup[subfloat]{labelformat=empty}
		\centering

		\subfloat[a) Dense volume]{
			\includegraphics[trim=\sparseLeftTrim{} \sparseBottomTrim{} \sparseRightTrim{} \sparseTopTrim{},clip, width=0.33\textwidth] {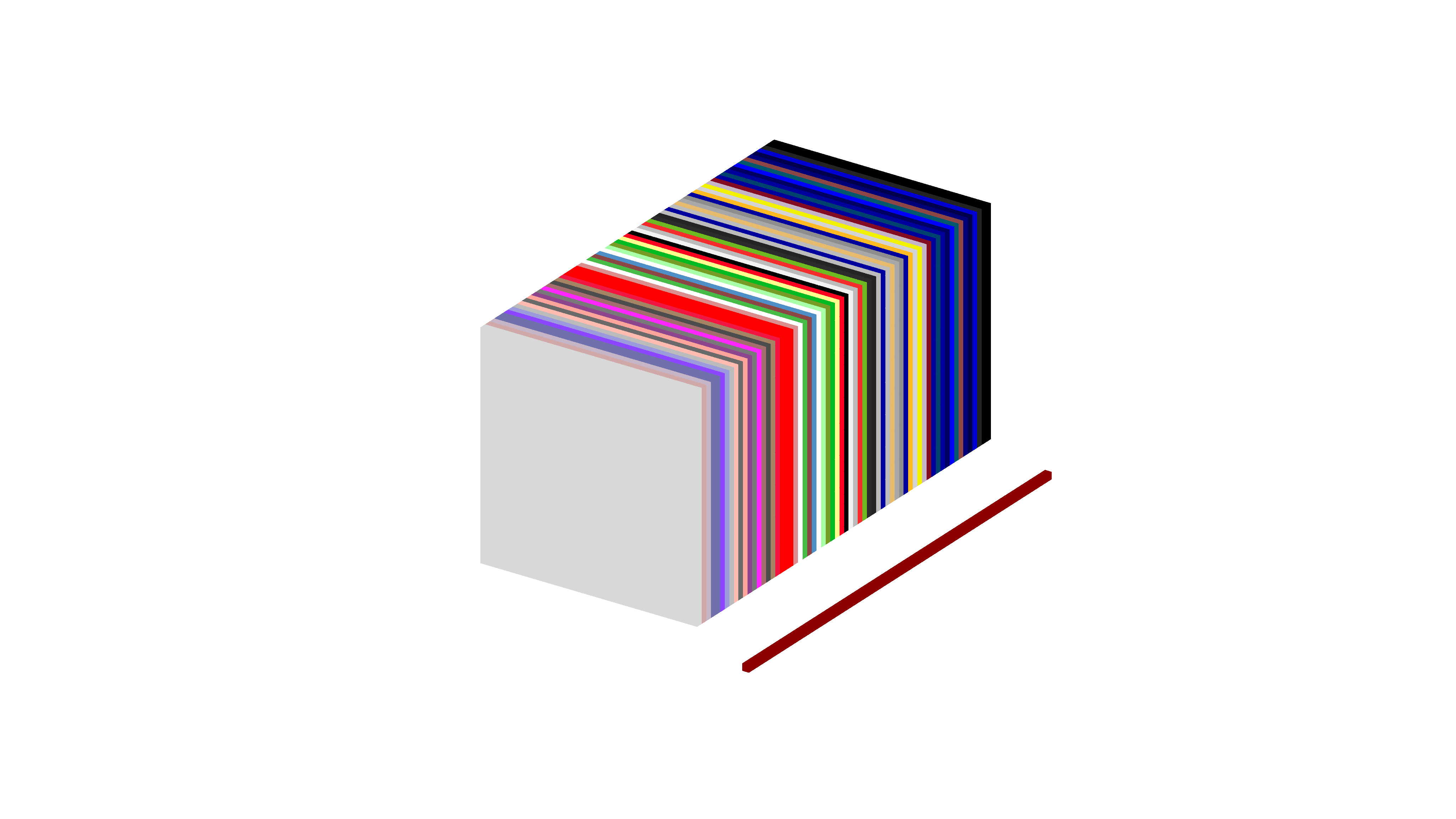}
		}
		\subfloat[b) Sparse, only committing]{
			\includegraphics[trim=\sparseLeftTrim{} \sparseBottomTrim{} \sparseRightTrim{} \sparseTopTrim{},clip, width=0.33\textwidth] {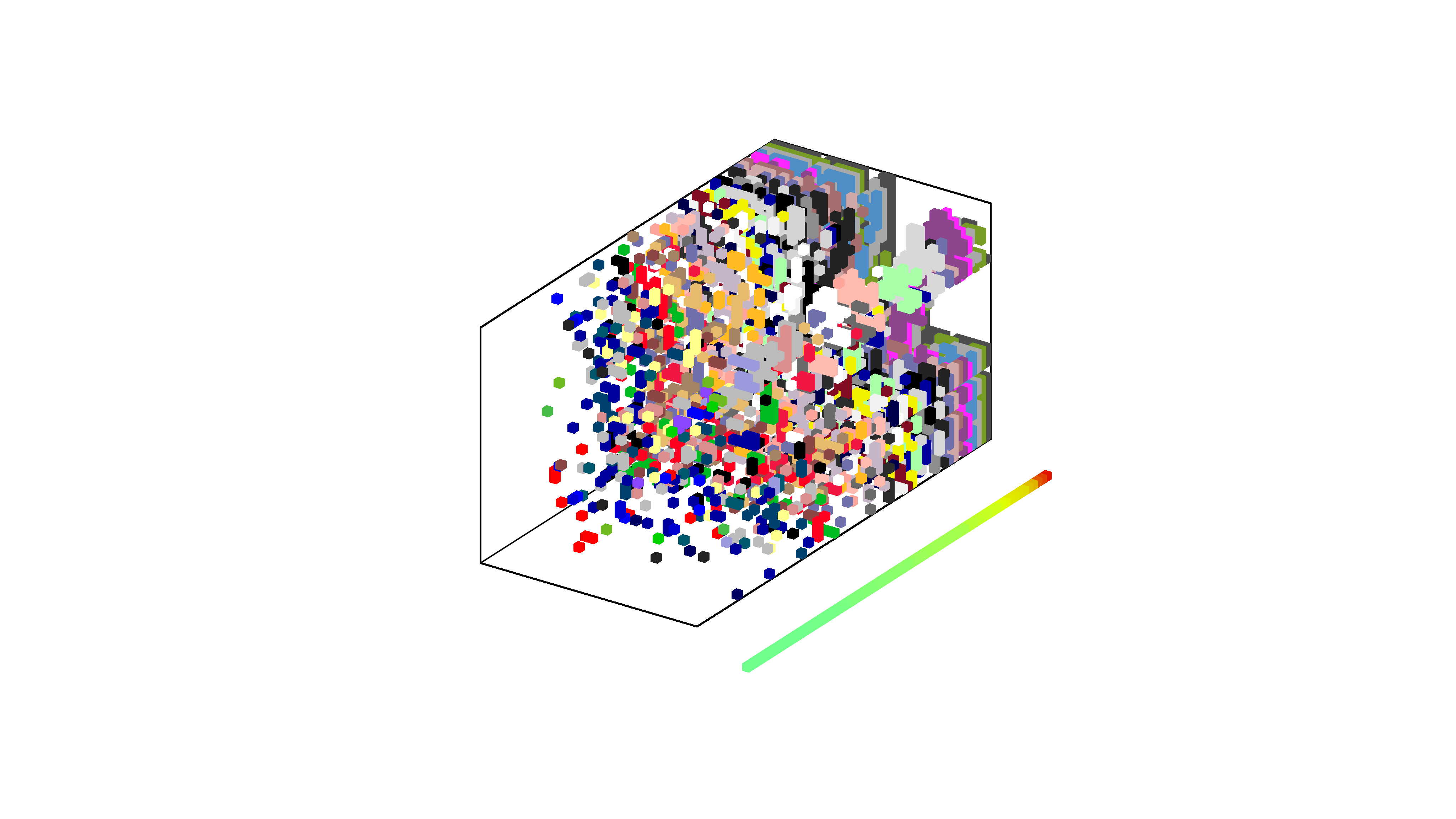}
		}
		\subfloat[c) Sparse]{
			\includegraphics[trim=\sparseLeftTrim{} \sparseBottomTrim{} \sparseRightTrim{} \sparseTopTrim{},clip, width=0.33\textwidth] {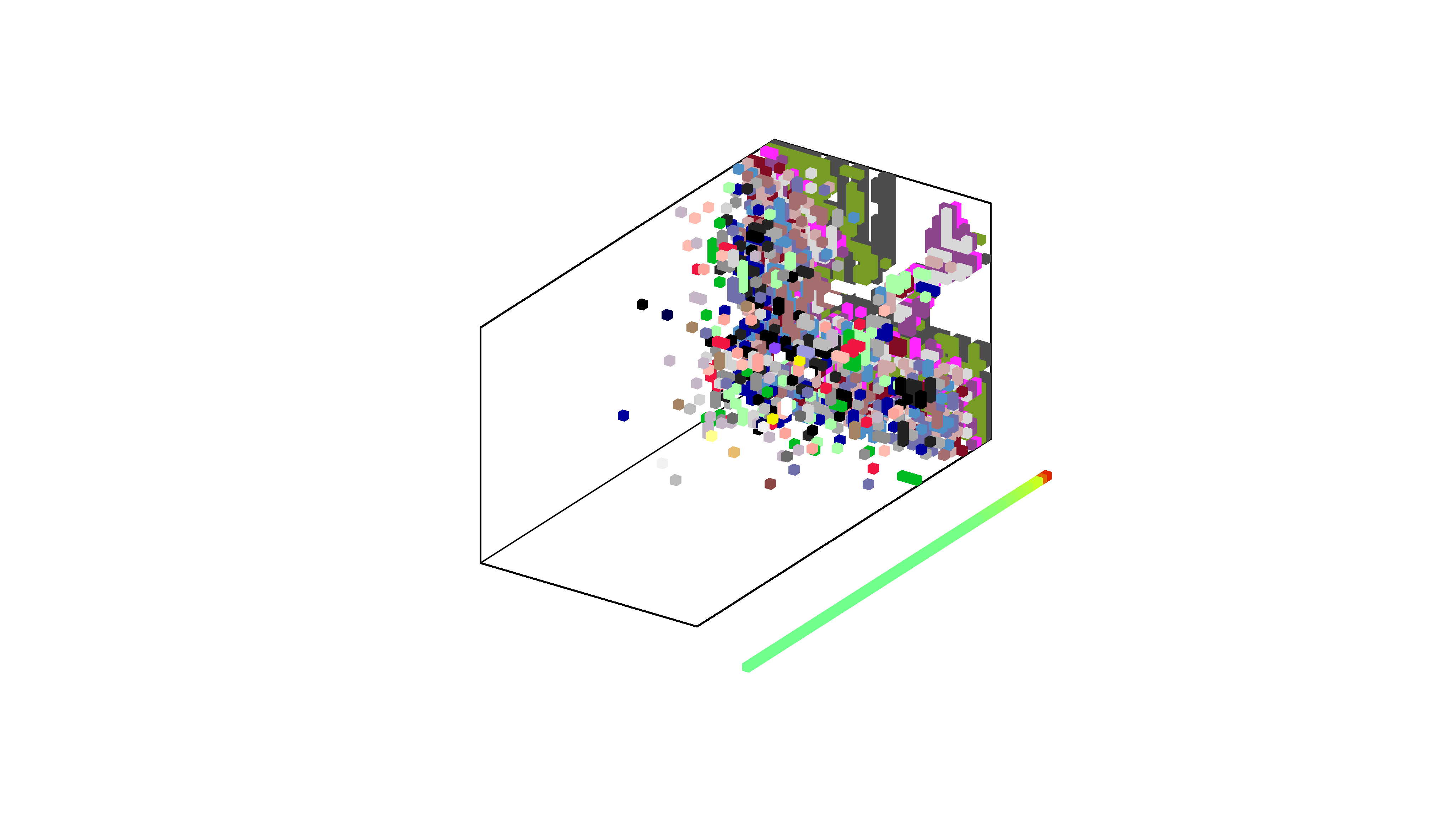}
		}
		\caption{Semantic probability volume for the courtyard dataset: The memory consumption of the dense 3D semantic texture (left) is prohibitive for most modern GPUs.
		Committing memory pages with non-negligible probabilities results in a sparse volume (middle) with only \SI{12.29}{\percent} allocated.
		Periodically removing low probability pages (right) further reduces the necessary memory (\SI{4.25}{\percent}) and ensures bounded memory usage that easily fits into GPU memory.}
		\label{fig:SparseVolume}
	\end{figure*}	
	
	\subsection{Color Integration}\label{sec:Color}
	In addition to the semantics, we also fuse the raw images into a global color texture. The fusion is carried out by a weighted running average:
	\begin{gather}
	\begin{aligned}
	C(\mathbf{u}_x)^{t} &= \frac{ W(\mathbf{u}_x)^{t-1} C(\mathbf{u}_x)^{t-1} + w_x I(\pi(g_F(\mathbf{p}_x)))  }{ W(\mathbf{u}_x)^{t-1} + w_x},\\
	W(\mathbf{u}_x)^{t} &= W(\mathbf{u}_x)^{t-1} + w_x.
	\end{aligned}
	\end{gather}
	The weight $w_x$ takes the distance, the radial intensity fall-off within an image, and the viewing angle into account:
	\begin{gather}
	\begin{aligned}
	w_{x} 	 &= w_{dist} \cdot w_{vign} \cdot w_{view},  \\
	w_{dist} &= (\norm{g_F(\mathbf{p}_{w,x})}^2_2)^{-1}, \\
	w_{vign} &= \cos(\theta_x)^4, \\
	w_{view} &= (\mathbf{o}_w-\mathbf{p}_{w,x}) \cdot \mathbf{n}_x.
	\end{aligned}
	\end{gather}
	Here, $w_{dist}$ is the inverse distance from the texel to the camera, which promotes frames that are spatially closer to the mesh, improving the resolution of the fused colors. The viewing angle $\theta_x$ between reprojection of the texel and the principal axis of the camera is used to account for the radial decrease in intensity by following the $\cos^4$ law~\citep{Goldman2005cos4}. 
	The third term $w_{view}$ increases the weight for texels that are imaged by the camera originating at $\mathbf{o}_w$ from a frontal perspective, further improving the quality of the fused texture.
	
	We chose different update schemes for color and semantics as their behavior is radically different. Firstly, the semantic segmentation is trained to be robust to illumination changes, hence the vignetting term is unnecessary. Secondly, it is not clear that the angle to the surface is a good indicator for confidence in semantic segmentation. In our experiments, we observed that the predictor learns to some extent the relative angle of surfaces with respect to camera view, thus weighting based on relative angle may be adversarial.
	For example during sudden camera movements tilting toward the ground, the semantic segmentation decreases in accuracy as the view is unfamiliar to the predictor. Thirdly, the distance weight assumes that the accuracy of the semantics is more confident for closer surfaces. However, this is not accurate for large semantic entities like buildings for which the accuracy decreases as we go closer, due to large untextured areas, and increases as we take a step back and observe the bigger picture.

	\subsection{Sparse Semantic Volume}\label{sec:SparseSemantic}
	The semantic 3D texture $S$ contains for each texel the probability distribution over all the classes. However, for reasonable sized resolutions and number of classes, this volume can occupy more memory than is typically available in modern GPUs, rendering this process infeasible. For a texture with $\num{8.192}\times\num{8.192}$~pixels and the \num{66}~classes of the Mapillary dataset, we would need to allocate a volume of \SI{16.5}{\giga\byte} (assuming we store each element as a floating point number of \num{4} bytes). This problem will only become worse as we add more class labels or increase texture size.
	In order to overcome this issue, we propose to store the semantics into a sparse 3D texture in which we allocate and deallocate dynamically the memory, ensuring bounded memory usage.

	In the first step, we divide our global semantic volume into pages of size $\num{128}\times \num{128}\times \num{1}$. Each page\footnote{Page size was chosen based on common supported values for multiple computers used during development.} stores the probability for only one class and can be either allocated in GPU memory or not. The volume starts initially with all pages in a deallocated state. Hence, it occupies no space on the GPU.

	When fusing the semantic probability from the current frame into $S$, the corresponding pages that will be affected are computed and targeted for committing on the CPU. This ensures that we only add the parts that are actually relevant. After each frame, we also check for pages that have low probability and deallocate them from memory.
	The probability for a texel is computed as:
	\begin{gather}
	\begin{aligned}
	p_{x_l} &= S(\mathbf{u}_x,l) / W(u_x).\\
	\end{aligned}
	\end{gather}
	If any texel inside a page is above a certain threshold, we will keep the corresponding page in memory, otherwise we target it for decommitting.
	This scheme of committing and decommitting portions of the memory can be seen as intrinsically \emph{tracking} the modes of the distribution over the classes, ignoring parts with negligible probability.
	
	Other schemes for memory reduction can be employed like quantization of the semantic 3D texture in which the stored probability is represented with a lower bit depth (reducing it to 2 bytes or even a mere 1 byte). However this approach is not scalable to bigger datasets with more classes and is prone to rounding errors in the case of very low bit depth. For this reason we deem that taking advantage of the sparsity in the semantic volume is a more appropriate method to deal with the memory issue.

	\subsection{Label Propagation}\label{sec:LabelPropagation}
	The fusion of semantic information from various view points into a global representation opens up possibilities to enforce temporal and spatial consistency of the semantic predictions by propagating the labels. The key insight is that if the majority of observations of the texture element predicts the correct class, the fused information $S^*$ will be confident enough ($p_{x_l} \geq p_{min}$) in order to be used as ground truth. Hence, we can reproject the mesh into any camera frame, propagate the label $\widetilde{L}(\mathbf{u}_F)$ and retrain the semantic segmentation in a semi-supervised fashion to minimize discrepancy between image predictions $L$ and mesh label $S^*$.
	Reprojecting the semantics into a camera frame is performed as follows:
	\begin{gather}
	\begin{aligned}
	\mathbf{u}_F &=\pi(g_F(\mathbf{p}_x)), \\ 
	S^* &= \argmax_c{S}, \\
	S^*_{F}(\mathbf{u}_F) &=  S^*(\mathbf{u}_x), \\ 
	\widetilde{L}(\mathbf{u}_F) &=
	\begin{cases}
	S^*_{F}(\mathbf{u}_F),		& p_{x_l} \geq p_{min}\\
	Unlabeled ,     & \text{otherwise}
	\end{cases}.
	\end{aligned}
	\end{gather}
	The result consists of the image $\widetilde{L}(\mathbf{u}_F)$ which we denote as \emph{pseudo ground truth}. This semantic labeling, together with the corresponding RGB view $I$ will be used to retrain the predictor.

	The retraining ensures that the semantic segmentation will segment objects and surfaces more consistent as belonging to a certain class, regardless of different or extreme view points or even illumination changes. Furthermore, the label propagation and retraining stages can be applied iteratively in an Expectation Maximization scheme, e.g. for a certain number of iterations, or until all camera frames reach a consensus. This process is illustrated in \reffig{fig:LabelPropagation}. The obvious caveat are wrong predictions used for retraining since bootstrapping this has a self-reinforcing character. We have seen this behavior only in some rare cases where the initial single-frame predictions were already incorrect, e.g. the table in the bottom row of \reffig{fig:NYUcompare}.

	Due to the significant number of camera frames contained in a large-scale dataset, we perform a conservative frame selection in order to choose only a subset of frames on which to reproject the semantics for retraining.
	The frame selection is based on an inconsistency coefficient, which rates frames higher the more instantaneous semantic segmentations deviates from mesh semantics.

	The inconsistency coefficient $\gamma$ is calculated as:
	\begin{gather}
	\begin{aligned}
	\gamma &=\sum_i B(L_i,  S^*_{F}(\mathbf{u}_F))  P^*,\\
	B(L_i,  S^*_{F}(\mathbf{u}_F)) &=
	\begin{cases}
	1,				&  L_i \neq S^*_{F}(\mathbf{u}_F)\\
	0,              & L_i = S^*_{F}(\mathbf{u}_F)
	\end{cases} \\
	\end{aligned}
	\end{gather}

	Given this coefficient for each frame, we select a restricted percentage of frames with the highest coefficient (in our experiments we choose \SI{5}{\percent} of the frames for NYU and \SI{15}{\percent} for the courtyard dataset) to be used for retraining. The intuition is that there is more information to gain by retraining on inconsistent frames as opposed to the ones which are already correct.
	Moreover, we also restrict the selected views to be at least 10 frames apart from one another to avoid adding redundant views that are too similar.
	In order to prevent forgetting during retraining, we add the original training set (in our case all the images from the NYU or Mapillary dataset, respectively) to the new views.
	
\begin{figure*}[]
		\captionsetup[subfloat]{labelformat=empty}
		\centering
		\subfloat[a) Fuse]{
			\fcolorbox{white}{white}{\includegraphics[width=0.3\textwidth] {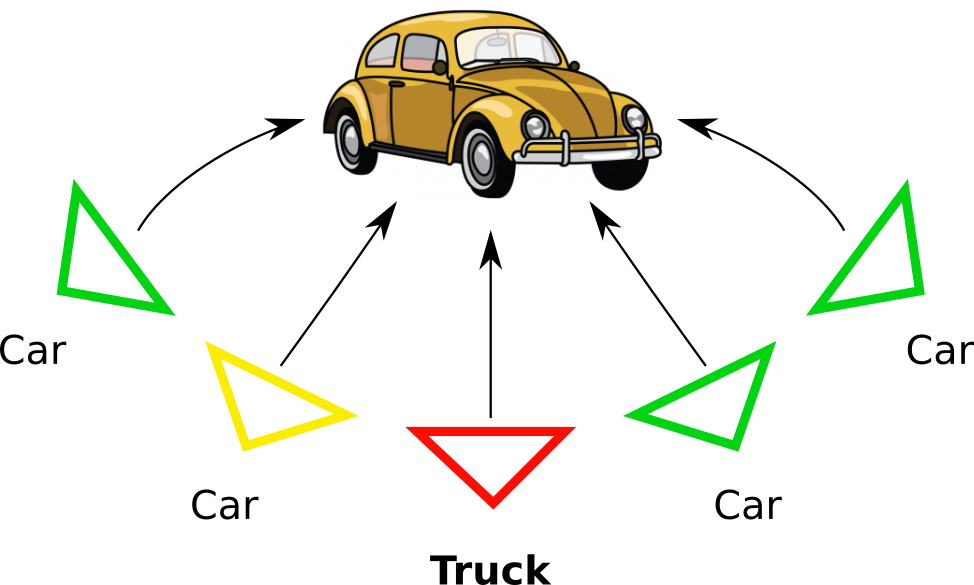}}	
		}
		\subfloat[b) Retrain]{
			\fcolorbox{white}{white}{\includegraphics[width=0.3\textwidth] {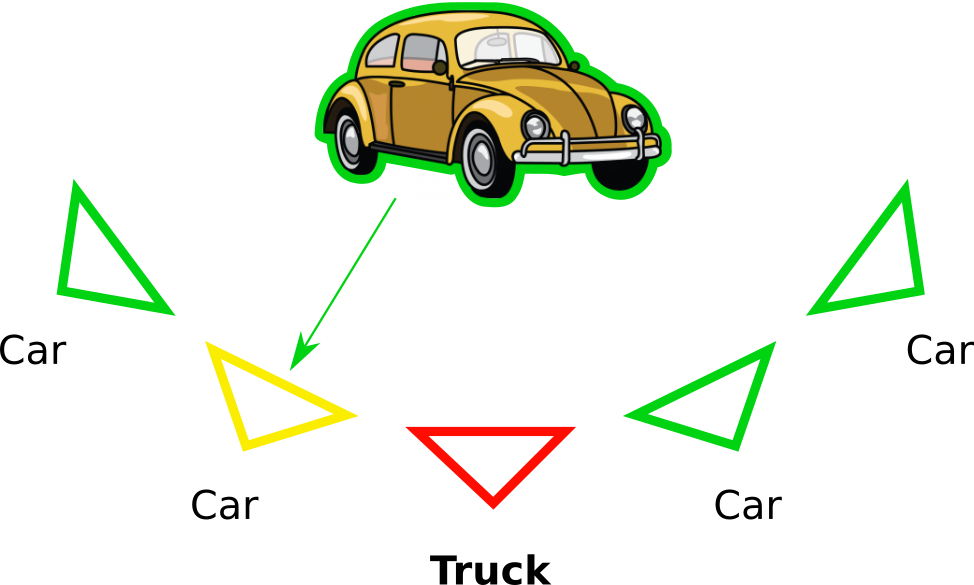}}	
		}
		\subfloat[c) Re-fuse]{
			\fcolorbox{white}{white}{\includegraphics[width=0.3\textwidth] {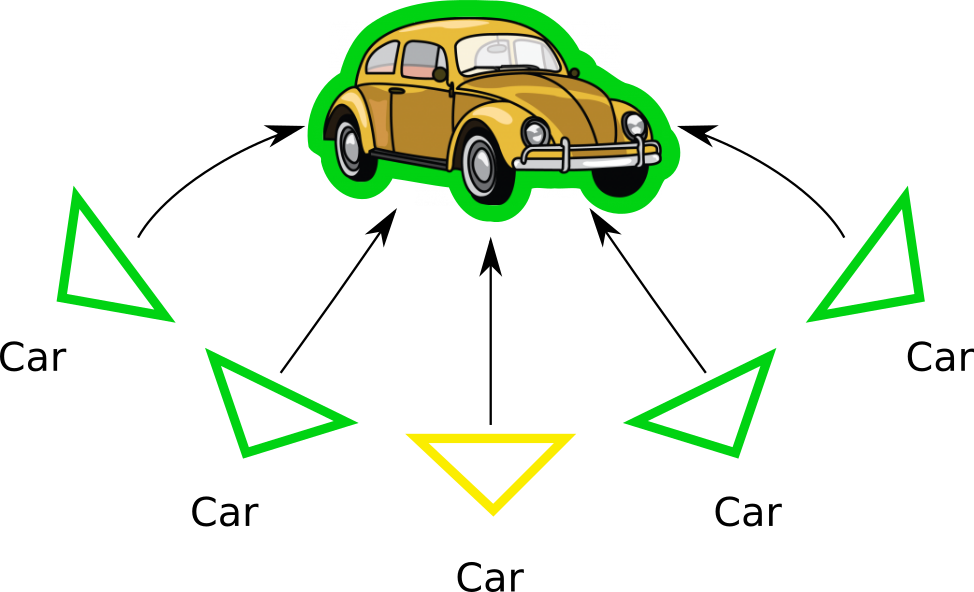}}	
		}
		\caption[Label propagation best frames]{Label propagation with best frames: Semantic segmentations are probabilistically fused in the scene (left). Frames with \emph{low} (but not too low) deviation from the stable mesh are used for retraining (middle). Close frames are naturally improved thereby. Inference is repeated for all images and the semantic labels are re-fused into the scene mesh to achieve a more accurate semantic map (right).}
		\label{fig:LabelPropagationBest}
	\end{figure*}

	Additionally, we experiment with another type of view selection in which we select for retraining the ones with the lowest rather than highest inconsistency coefficient. The intuition behind this alternative scheme is that by reinforcing good frames (or the ones that are close to being good), the other frames that are "close" to them will also be improved. Therefore, performing the label propagation iteratively will eventually "lift" the less consistent frames to become better by reinforcing the ones that are already good or close to being good. An illustration of this process is depicted in~\reffig{fig:LabelPropagationBest}. We ignore for this selection type the frames which have a too high consistency level as we saw that in the general case they provide too little new information. We rather focus on frames that are quite consistent but not fully. Hence, frames with more than \SI{98}{\percent} of consistent pixels are ignored. From the remaining set we select the same percentages as mentioned earlier.
	
	\section{Implementation}
		Our pipeline consists of three modules, the mesh generator, the semantic texture integrator and the segmentation retraining.
		The mesh generator module is fully implemented in \CC{} and integrated into ROS~\citep{quigley2009ros} for ease of interaction with other ROS packages.
		The texturer was also developed in \CC{} with the addition of OpenGL for rendering and semantic integration using GLSL compute shaders.
		
		We will describe in detail the optimization choices made for semantic integration as they are the main focus of this work.
		The segmentation map is initially precalculated from the predictor and stored to disk. An asynchronous module reads the RGB images and the corresponding segmentation maps and stores them in ringbuffers ready to be processed by the texturing module.
		The texturing module receives the images and transfers them to the GPU using double buffered Pixel Buffer Objects (PBO) in a method commonly known as \emph{ping ponging}.
		This ensures that the transfer can be done on the GPU side using Direct Memory Access (DMA), freeing the CPU to do other tasks in the meantime.
		The semantic integration is performed fully on the GPU through efficient compute shaders.
		
		In order to deal with the sparsity of our semantic 3D texture we make use of the \textit{GL\_ARB\_sparse\_texture} extension from OpenGL which provides functionality for committing and decommitting from sparse (partially resident) textures.
		However, the committing and decommitting of pages can only be performed from the CPU side, which requires synchronization and communication between CPU and GPU. We use two buffers for this communication, one for signaling to the CPU which pages require committing and one for confirmation to the GPU that they were committed. These buffers are updated asynchronously and are also double buffered to prevent stalling the pipeline.
		
		While double buffering allows maximum usage of the available resources, it also implies a delay of one frame between the CPU-GPU communication which in our case does not pose a problem due to the high frame rate at which the camera images arrive.
		
	\definecolor{my_darkgray}{RGB}{70, 70, 70}
	\newcommand\RotText[1]{\rotatebox{90}{\parbox{2cm}{#1}}}
	\newcolumntype{Y}{>{\centering\arraybackslash}X}
	
\definecolor{Background}{rgb}{0.07843137 0.07843137 0.07843137}
	\definecolor{Bed}{rgb}{0.         0.50196078 0.50196078}
	\definecolor{Bookshelf}{rgb}{0.98039216 0.19607843 0.19607843}
	\definecolor{Ceiling}{rgb}{0.4        0.         0.8       }
	\definecolor{Chair}{rgb}{0.19607843 0.19607843 0.98039216}
	\definecolor{Floor}{rgb}{0.8627451  0.8627451  0.8627451 }
	\definecolor{Furniture}{rgb}{1.         0.27058824 0.07843137}
	\definecolor{Objects}{rgb}{1.         0.07843137 0.49803922}
	\definecolor{Picture}{rgb}{0.19607843 0.19607843 0.58823529}
	\definecolor{Sofa}{rgb}{0.87058824 0.70588235 0.54901961}
	\definecolor{Table}{rgb}{0.19607843 0.98039216 0.19607843}
	\definecolor{Tv}{rgb}{1.         0.84313725 0.        }
	\definecolor{Wall}{rgb}{0.58823529 0.58823529 0.58823529}
	\definecolor{Window}{rgb}{0.         1.         1.        }	
	\bgroup
	\arrayrulecolor{gray} 
	\renewcommand{\arraystretch}{1.5}  
	\begin{table*}[ht]
		\centering
		\begin{tabularx}{\textwidth}{@{\extracolsep{\fill}}|c|Y|Y|Y|Y|Y|Y|Y|Y|Y|Y|Y|Y|Y|c|}

			\hline
			Method &
			\RotText{\cellcolor{Bed}\textcolor{white}{Bed}} &
			\RotText{\cellcolor{Bookshelf}Bookshelf} &
			\RotText{\cellcolor{Ceiling}\textcolor{white}{Ceiling}} &
			\RotText{\cellcolor{Chair}\textcolor{white}{Chair}}&
			\RotText{\cellcolor{Floor}Floor} &
			\RotText{\cellcolor{Furniture}Furniture} &
			\RotText{\cellcolor{Objects}Objects} &
			\RotText{\cellcolor{Picture}\textcolor{white}{Picture}} &
			\RotText{\cellcolor{Sofa}Sofa} &
			\RotText{\cellcolor{Table}Table} &
			\RotText{\cellcolor{Tv}Tv} &
			\RotText{\cellcolor{Wall}Wall} &
			\RotText{\cellcolor{Window}Window} &
			Mean IoU\\
			\noalign{
				\color{my_darkgray}
				\hrule height 1.2pt
			}%

			\rule{0pt}{0.7cm} 
			Single frame &
			0.46 &
			0.17 &
			0.13 &
			0.25 &
			0.68 &
			0.34 &
			0.28 &
			0.33 &
			0.22 &
			0.15 &
			0.12 &
			0.51 &
			0.29 &
			0.302 \\ 
			\hline

			Single frame with LP &
			0.52 &
			0.20 &
			0.14 &
			0.27 &
			0.68 &
			0.35 &
			0.30 &
			0.35 &
			0.26 &
			0.14 &
			0.14 &
			0.53 &
			0.31 &
			0.322 \\ 
			\noalign{
				\color{my_darkgray}
				\hrule height 1.2pt
			}%

			SemanticFusion &
			0.47 &
			0.15 &
			0.18 &
			0.30 &
			0.65 &
			0.36 &
			0.30 &
			0.35 &
			0.24 &
			0.15 &
			0.20 &
			0.53 &
			0.33 &
			0.324 \\ 
			\hline

			SF with LP &
			0.52 &
			0.18 &
			0.21 &
			0.31 &
			0.65 &
			0.38 &
			0.31 &
			0.38 &
			0.28 &
			0.16 &
			0.20 &
			0.54 &
			0.36 &
			0.343 \\ 
			\noalign{
				\color{my_darkgray}
				\hrule height 1.2pt
			}%

			\rule{0pt}{0.5cm}
			Ours  &
			0.54 &
			0.17 &
			0.23 &
			0.35 &
			\textbf{0.71} &
			0.40 &
			0.33 &
			0.39 &
			0.28 &
			\textbf{0.18} &
			0.21 &
			0.56 &
			0.37 &
			0.363 \\ 
			\hline

			Ours with LP &
			\textbf{0.56} &
			\textbf{0.20} &
			\textbf{0.25} &
			\textbf{0.35} &
			0.68 &
			\textbf{0.40} &
			\textbf{0.34} &
			\textbf{0.41} &
			\textbf{0.31} &
			0.17 &
			\textbf{0.23} &
			\textbf{0.57} &
			\textbf{0.38} &
			\textbf{0.372} \\ 
			\hline
		\end{tabularx}
		\caption{NYUv2 results: We compare our method against single-frame predictions and SemanticFusion~\citep{mccormac2017semanticfusion}. All cases are evaluated with and without Label Propagation (LP). For the case of single-frame, we exclude pixel without a valid depth measurement. All evaluations were performed at $320\times240$ resolution.}
		\label{tab:NYUiou}
	\end{table*}
	\egroup		
			
	\section{Experiments}
	All tests were performed on a Intel Core i7-940 $\SI{2.93}{\giga\hertz}$ CPU with an NVIDIA Titan GPU.
	\subsection{NYUv2 Dataset}
		For comparison against SemanticFusion~\citep{mccormac2017semanticfusion} we utilize the NYUv2 dataset~\citep{silberman2012nyu} and use \num{108} out of all sequences where ElasticFusion did not exhibit significant drift. For fairness comparability, we use the final surfel map for meshing and the segmentation module of \cite{eigen2015cnn} used within SemanticFusion that is trained on the \num{13} NYU classes. 
		Furthermore, we store the semantics for Intersection-over-Union (IoU) calculation after the scene is completed and fused semantics are static.
		
		The implementation of SemanticFusion provides two CNNs pretrained on the NYUv2 dataset, one that receives RGB only and another that receives RGB-D data from the Kinect. While the original work of SemanticFusion evaluates their approach using the RGB-D-CNN ( including scene depth as an additional feature map ) we use for our evaluation their RGB-CNN since we want our method to work well even in outdoor scenarios where dense depth may not be obtainable.
	
		The network is retrained using both methods for view selection (worse or best frames) with a learning rate of \num{e-6}. The model is saved after each epoch and training is stopped when the model begins to overfit or the IoU for the epoch starts to decrease.

		For retraining with the worse frames we add from each NYU scene \SI{5}{\percent} of the frames with the highest inconsistency coefficient. 
		This percentage is chosen such that the amount of pseudo ground truth frames is comparable to the \num{795} originally used for training. The retraining then uses both the original training set and our pseudo ground truth.
		In the case of retraining with the best frames we choose the \SI{5}{\percent} frames with the \emph{lowest} inconsistency coefficient, ignoring however those that have too few inconsistent pixels (in our case we choose those that have at least \SI{2}{\percent} of the pixels labeled as inconsistent).

	\subsection{Courtyard Dataset}
		The courtyard dataset~\citep{droeschel2018laser} was captured using a DJI Matrice 600, with a horizontally attached Velodyne VLP-16 laser scanner. The lidar has \num{16} horizontal scan lines and a vertical field-of-view of \SI{30}{\degree} with a maximum range of \SI{100}{\metre}. The color images were captured at \SI{10}{\hertz} using two synchronized global shutter Point Grey Blackfly-S U3-51S5C-C color cameras, with a resolution of $\num{2448}\times\num{2048}$ pixels. The MAV poses for this dataset were provided by \cite{droeschel2018laser}. The camera poses were calculated from the provided extrinsics and the continuous-time trajectory under consideration of an additional \SI{40}{\milli\second} time offset. The two cameras are mounted outward pointing on the left and right side of the copter to improve visual coverage. In total \num{13458} frames were captured during the experiment.
		We densely annotated \num{48} images spread throughout the area to conduct accuracy experiments.

		RefineNet with ResNeXt-101~\citep{xie2017resnext} was trained on the Mapillary dataset for \num{10} epochs with a learning rate of \num{e-5} and a batch size of \num{1}. The final IoU achieved is \num{0.4546} placing the result on 3rd position on the Mapillary leaderbord\footnote{\url{https://eval-vistas.mapillary.com/featured-challenges/1/leaderboard/1}}. The retraining with pseudo ground truth has to be done with a larger batch size of \num{16} to account for the decreased signal-to-noise-ratio introduced by the pseudo ground truth. We performed five epochs of further retraining.
		The retrained RefineNet achieves \num{0.4487} IoU on the Mapillary dataset. The slight reduction in accuracy is explained by the fact that the neural network is forced to learn from a new dataset which is different from the Mapillary one in which it is evaluated.
	
\subsection{Accuracy Evaluation}

		We execute the two variants of Label Propagation for three iterations on the NYUv2 dataset.  We observe that retraining on the worse frames yields a higher IoU hence we prefer this option for all further experiments. Furthermore, the highest increase in IoU is experienced after the first iteration, while subsequent ones yield a noticeably less improvement. We hypothesize that a random selection strategy would be placed somewhere in between both variants since some with high and some with low inconsistency coefficients would be chosen.
		
		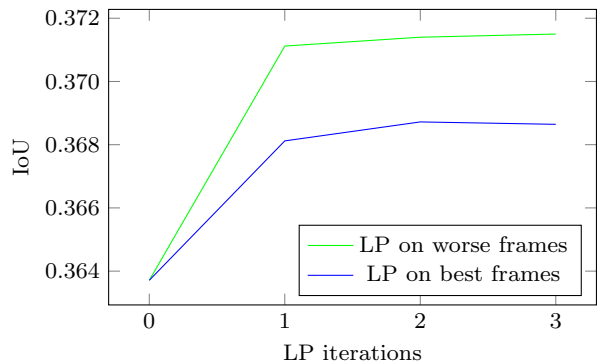
\begin{figure}
			\centering
			\begin{tikzpicture}
			\begin{axis}[
			width=8.0cm,
			height=5.5cm,
			legend pos=south east,
			xlabel={LP iterations},
			ylabel={IoU },
			y tick label style={/pgf/number format/.cd, fixed, fixed zerofill,precision=3},
			xtick={0,1,2,3}
			]
			
			\addplot [color=green]  table [x=iter, y=worse_increasing_data_iou, col sep=comma, mark=none,
			] {ilp_nyu.txt};
			\addplot [color=blue]  table [x=iter, y=best_increasing_data_iou, col sep=comma, mark=none,
			] {ilp_nyu.txt};
			
			\addlegendentry{LP on worse frames}
			\addlegendentry{LP on best frames}
			
			\end{axis}
			
			\end{tikzpicture}
			\caption[Incremental LP NYU]{LP variants: We evaluate the IoU increase by performing LP on the worse or the best frames, respectively. Propagating the labels towards the worse frames yields a higher IoU. Both LP variants converge quickly after the first iteration.}
			\label{plt:ilpNYU}
		\end{figure}
	
		We computed the IoU for different configurations on the NYUv2 dataset, including single-frame predictions and Semantic Fusion. Label propagation was performed using our approach to retrain the predictor. We denote in \reftab{tab:NYUiou} the use of the retrained semantic segmentation as \emph{with LP}.
		\reftab{tab:NYUiou} shows that our method outperforms single-frame as well as SemanticFusion. Using label propagation further improves the IoU. A visual comparison is provided in \reffig{fig:NYUcompare} for four different scenes. Already SemanticFusion improves single-frame predictions e.g. on the TV (yellow, first row), the window (blue, third row) and wall (gray, last row), but the result is noisy and partially inconsistent. We attribute this mostly to surfels on different scales that are not correctly fused. In comparison, our mesh is more consistent, for example on the bath tub (second row), but smoother around the edges. Yet, the bed (third row) is still mostly classified as a sofa. Through our label propagation and subsequent retraining, we were able to correct the classification. In the last row, we show a failure case in which the Label Propagation decreases the accuracy as the table (green) gets segmented as furniture. This decrease in accuracy is due to the fact that most single-frame predictions are wrongly labeling the object and establishing consistency through LP reinforces this wrong labeling.
		Further complete reconstructed scenes from NYUv2 are visualized in \reffig{fig:NYUreconstruction}.

	\newcommand{\qualitativeElemWidth}{2.2cm}
	\renewcommand{\tabcolsep}{1pt}
	\begin{figure*}
		\centering
		\begin{tabular}{cccccc}
			RGB image&Ground Truth&Single frame& SemanticFusion&Ours&Ours with LP \\
			\includegraphics[width=\qualitativeElemWidth{}]{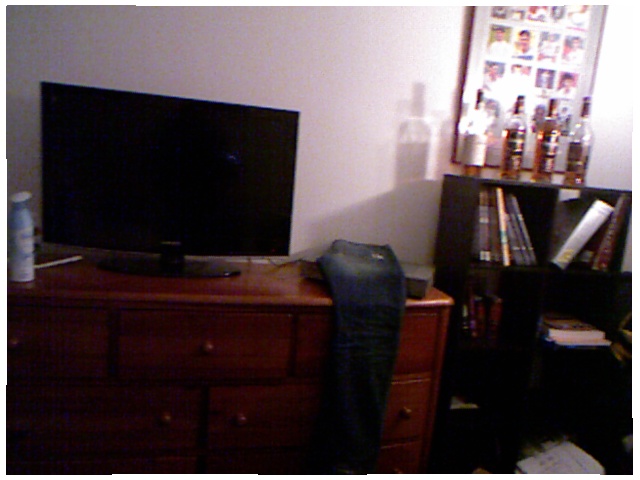}& 
			\includegraphics[width=\qualitativeElemWidth{}]{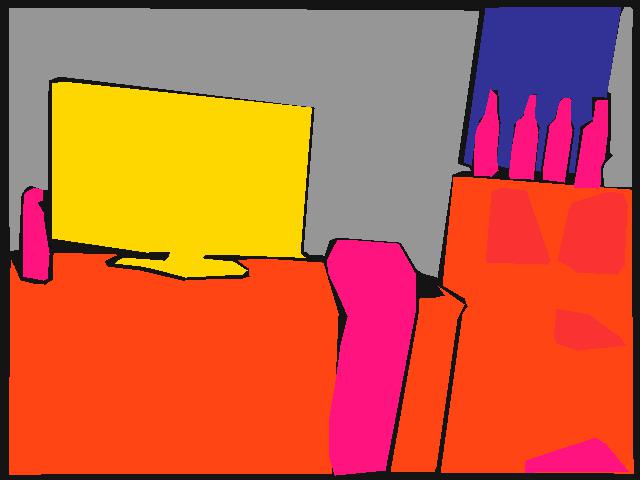}&           
			\includegraphics[width=\qualitativeElemWidth{}]{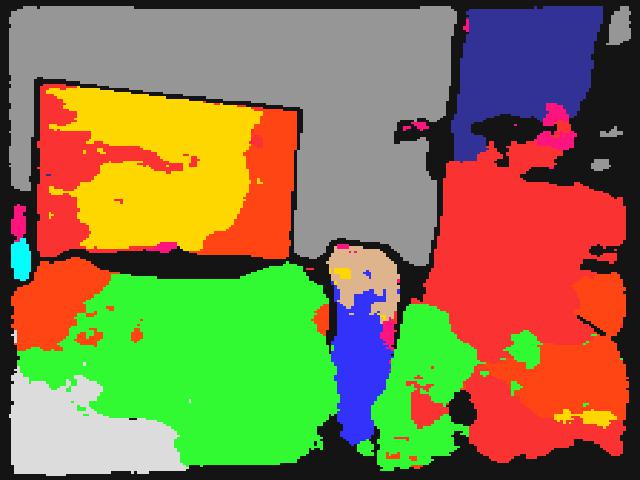}&  
			\includegraphics[width=\qualitativeElemWidth{}]{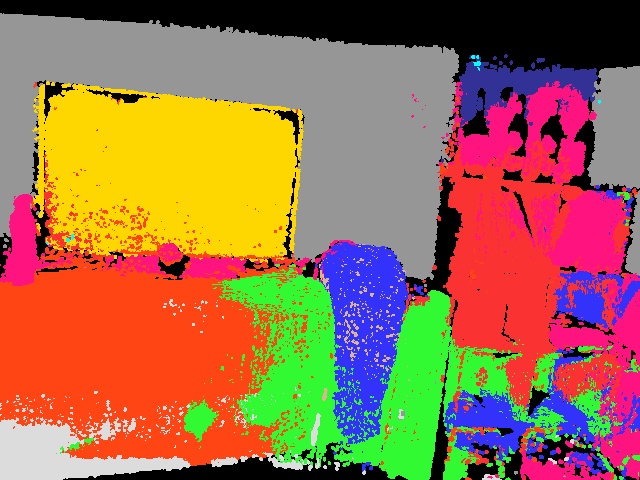}& 
			\includegraphics[width=\qualitativeElemWidth{}]{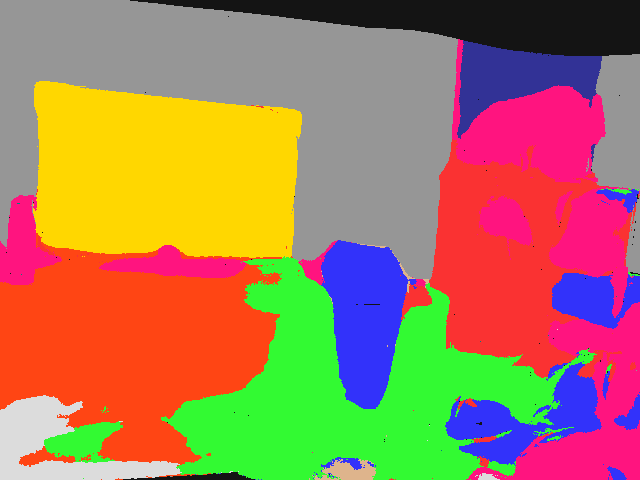}&       
			\includegraphics[width=\qualitativeElemWidth{}]{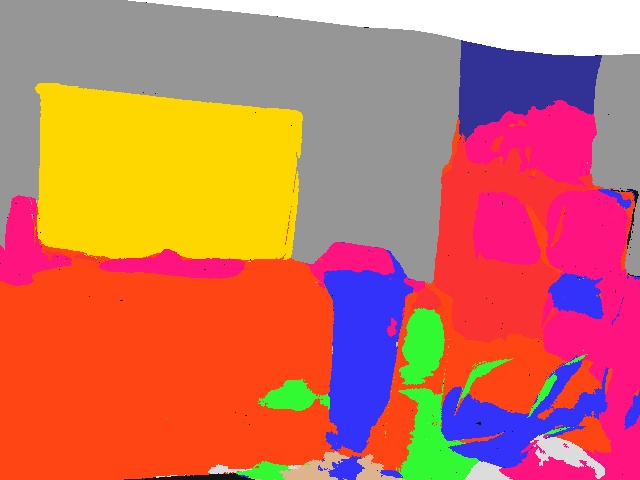}\\   

			\includegraphics[width=\qualitativeElemWidth{}]{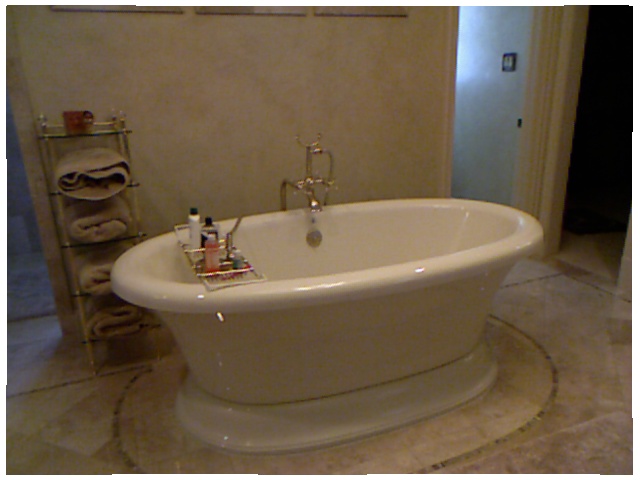}& 
			\includegraphics[width=\qualitativeElemWidth{}]{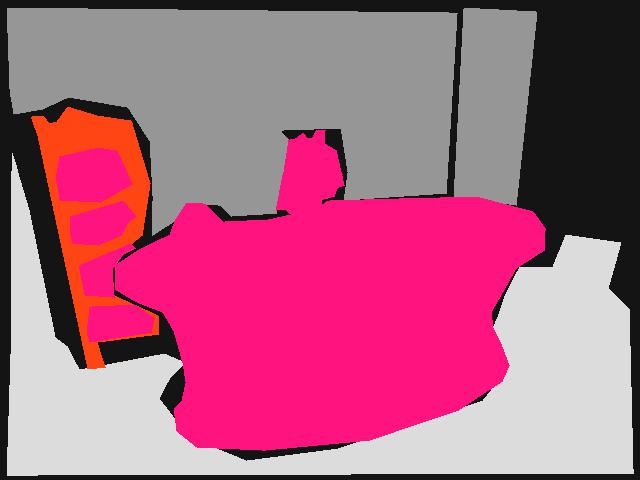}&           
			\includegraphics[width=\qualitativeElemWidth{}]{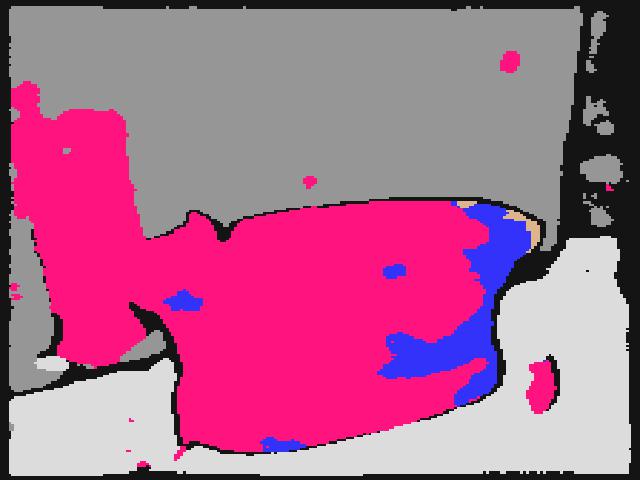}&  
			\includegraphics[width=\qualitativeElemWidth{}]{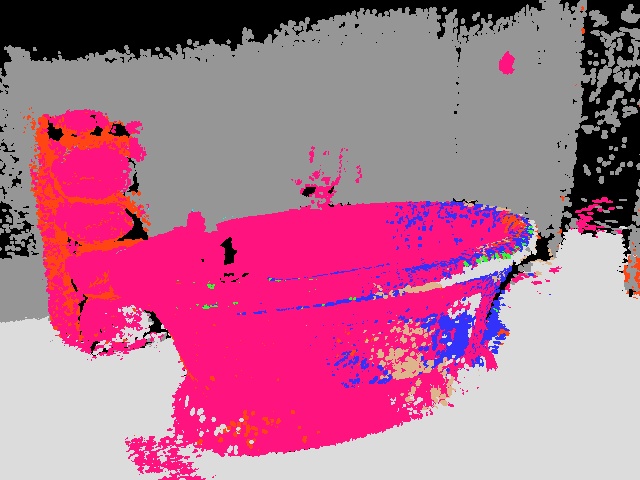}& 
			\includegraphics[width=\qualitativeElemWidth{}]{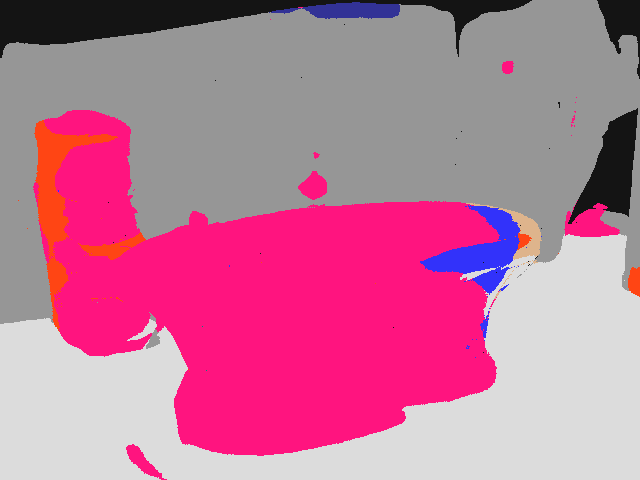}&       
			\includegraphics[width=\qualitativeElemWidth{}]{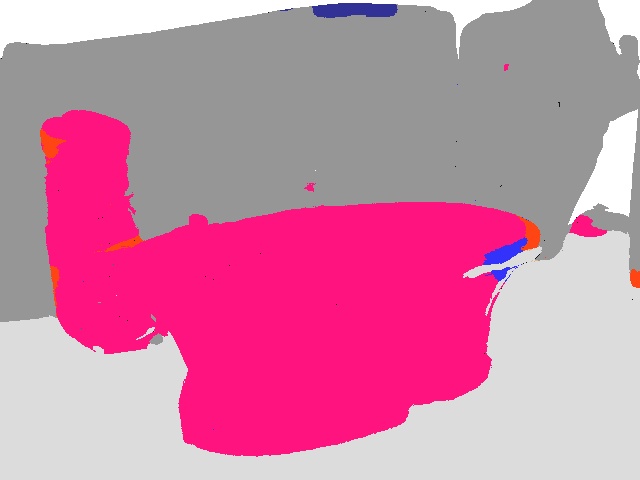}\\   

			\includegraphics[width=\qualitativeElemWidth{}]{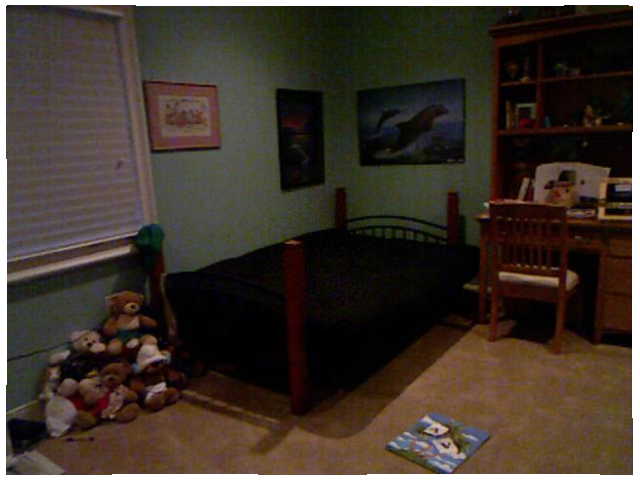}& 
			\includegraphics[width=\qualitativeElemWidth{}]{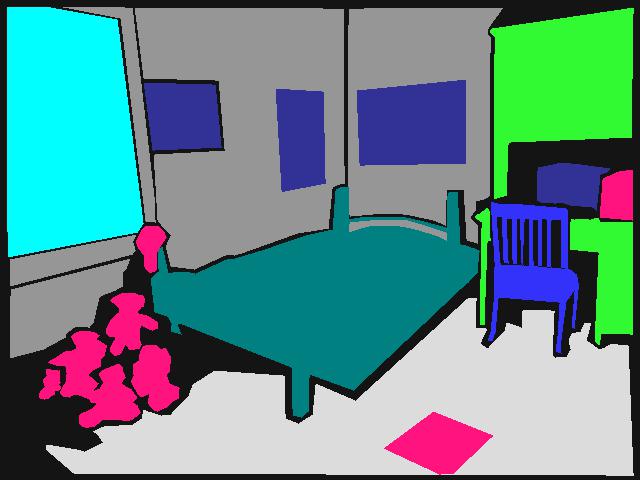}&           
			\includegraphics[width=\qualitativeElemWidth{}]{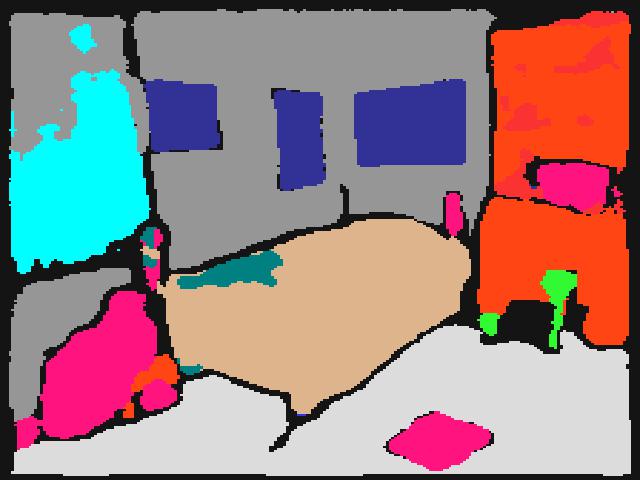}&  
			\includegraphics[width=\qualitativeElemWidth{}]{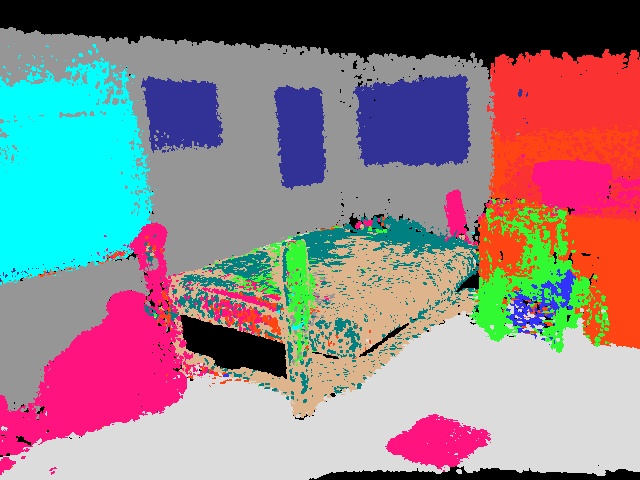}& 
			\includegraphics[width=\qualitativeElemWidth{}]{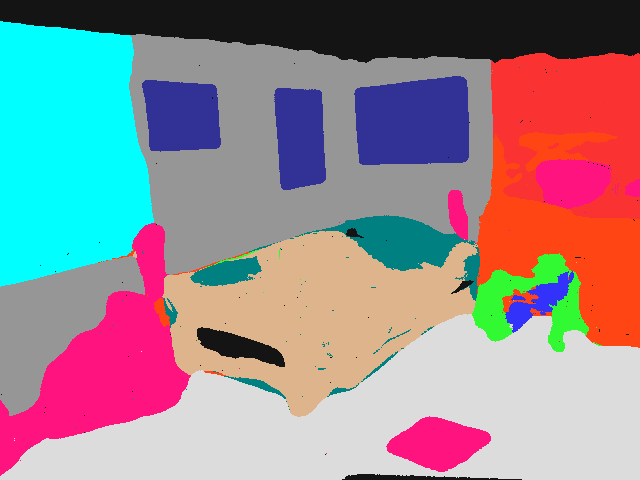}&       
			\includegraphics[width=\qualitativeElemWidth{}]{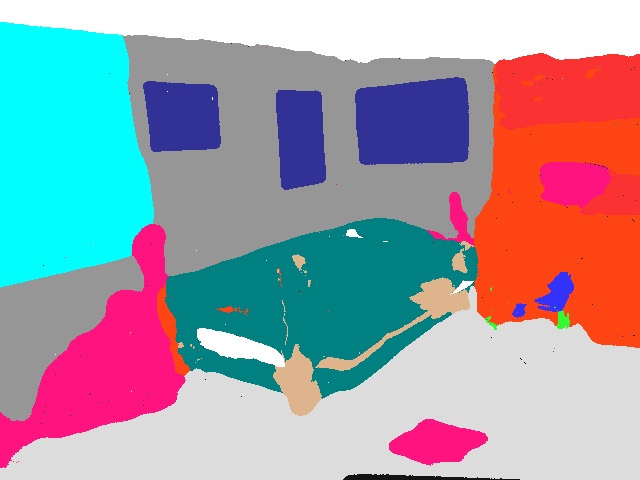}\\   

			\includegraphics[width=\qualitativeElemWidth{}]{home_office_10_rgb.jpg}& 
			\includegraphics[width=\qualitativeElemWidth{}]{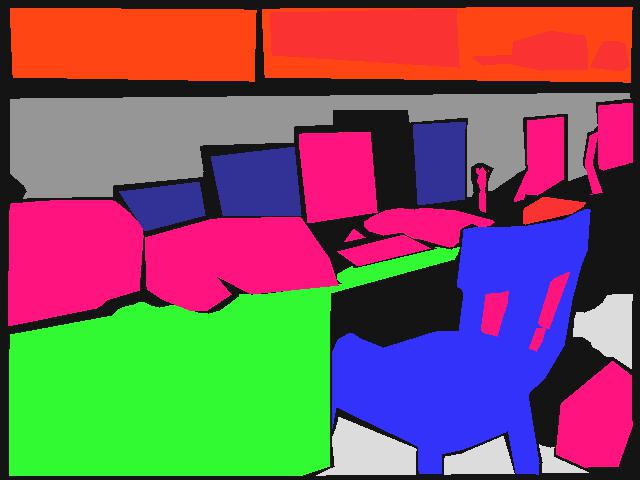}&           
			\includegraphics[width=\qualitativeElemWidth{}]{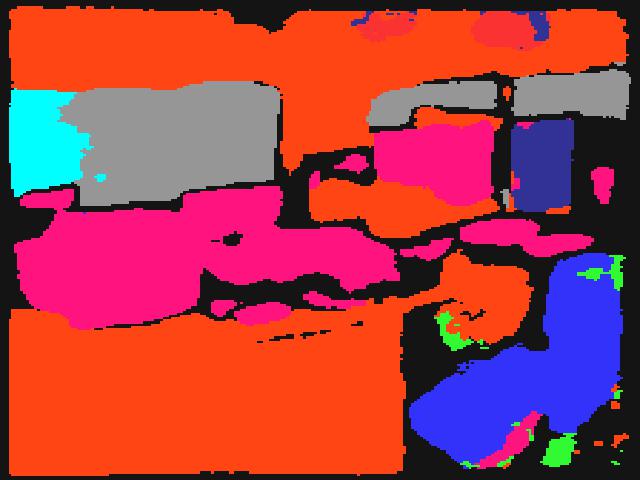}&  
			\includegraphics[width=\qualitativeElemWidth{}]{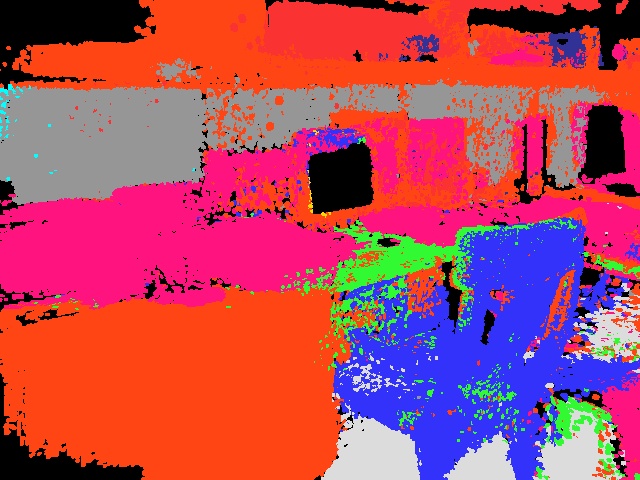}& 
			\includegraphics[width=\qualitativeElemWidth{}]{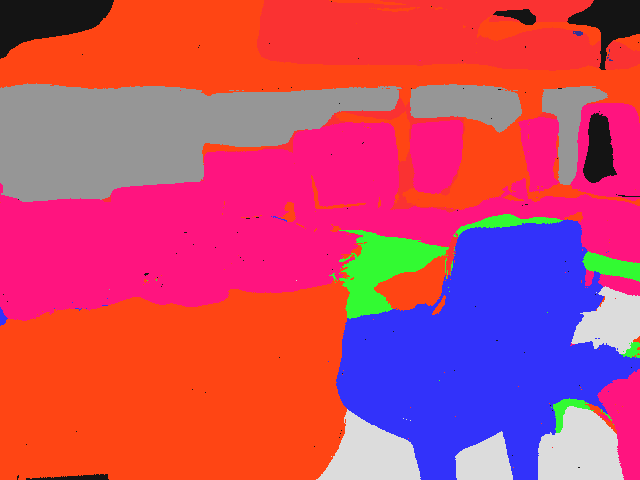}&       
			\includegraphics[width=\qualitativeElemWidth{}]{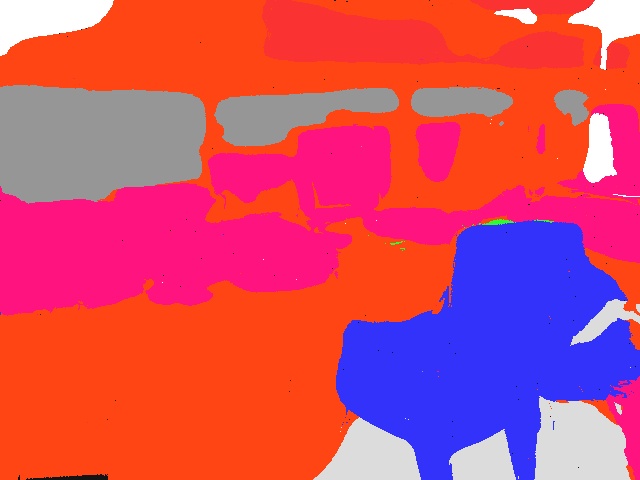}\\   
		\end{tabular}
		\caption{NYUv2 qualitative results: We compare our method, with and without Label Propagation, against single-frame predictions and SemanticFusion. The first three rows show a clear improvement achieved through Label Propagation, as the predictor learns to segment the table, bathtub and bed more accurately. The last row shows a failure case in which the Label Propagation decreases the accuracy as the table represented in green gets segmented as furniture. This decrease in accuracy is due to the fact that most single-frame predictions are wrongly labeling the object and establishing consistency through LP reinforces this wrong labeling.}
		\label{fig:NYUcompare}
	\end{figure*}

	\newlength{\nIW}
	\newlength{\nIH}
	\newcommand{\scenesTableElemWidth}{5.5cm}
	\begin{figure*}
		\centering
		\begin{tabular}{cccc}
			Mesh& RGB reconstruction& Semantic map \\
			\settowidth{\nIW}{\includegraphics{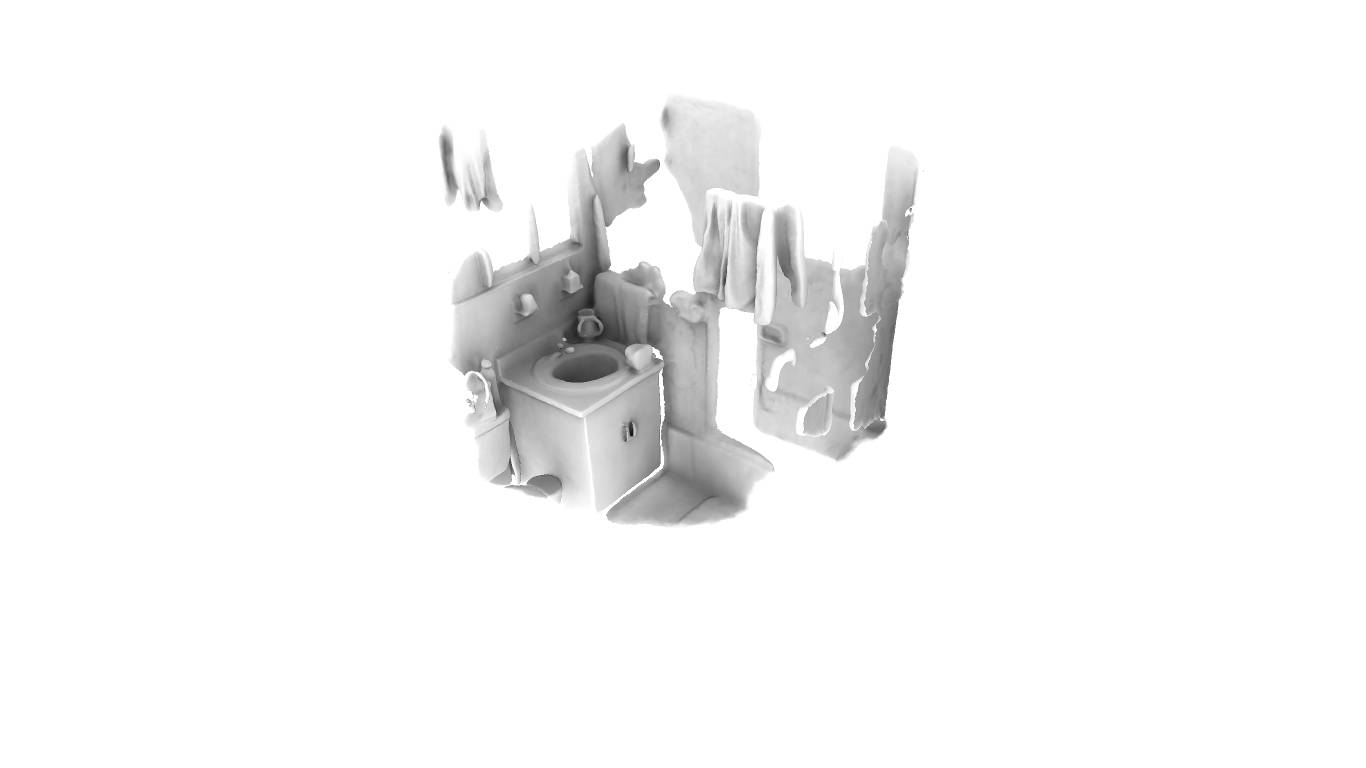}}
			\settoheight{\nIH}{\includegraphics{bathroom_0003_mesh.jpg}}
			\includegraphics[trim=.256\nIW{} .3255\nIH{} .2745\nIW{} .078\nIH{},clip, width=\scenesTableElemWidth{}]{bathroom_0003_mesh.jpg}& 
			\settowidth{\nIW}{\includegraphics{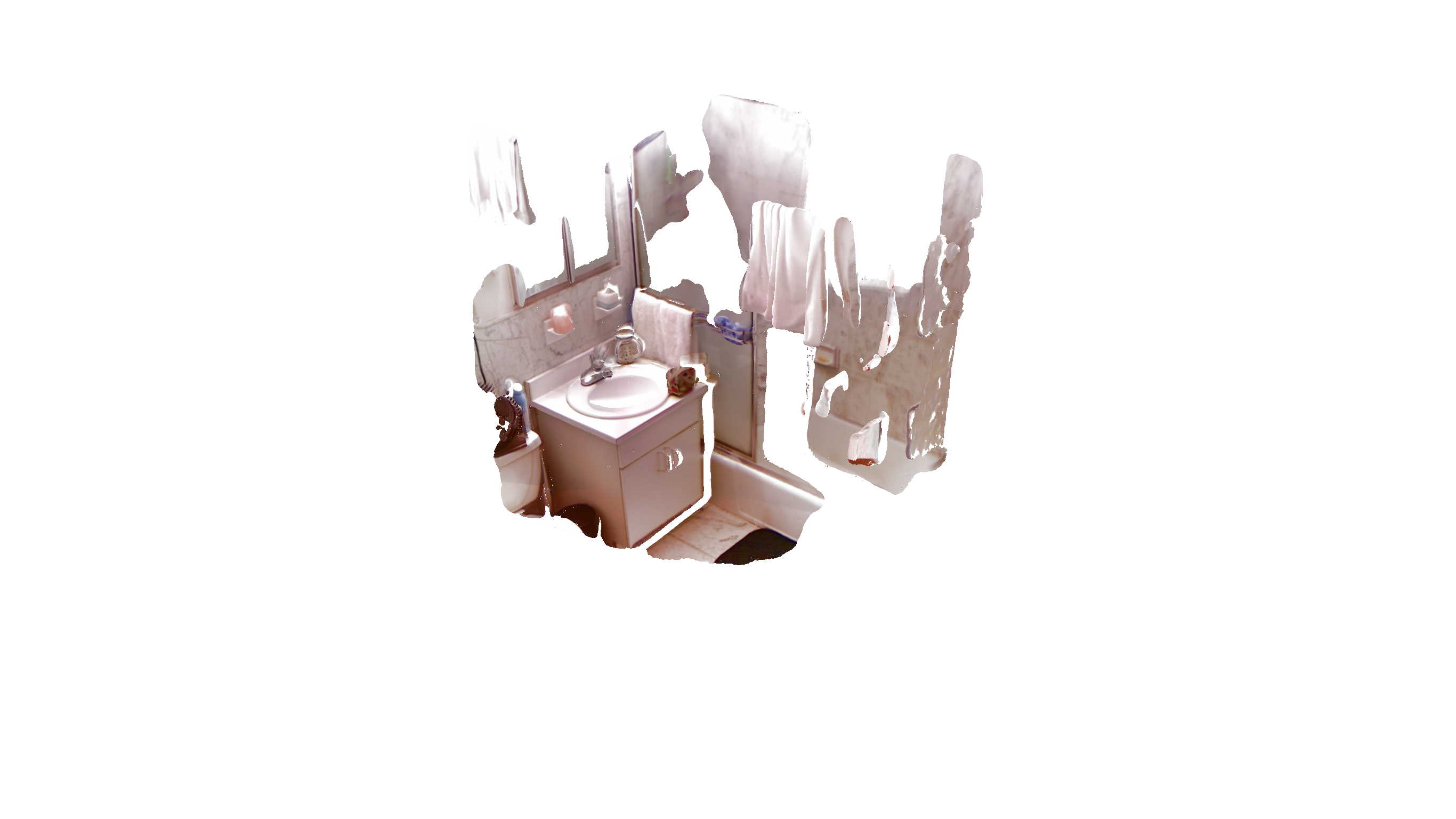}}
			\settoheight{\nIH}{\includegraphics{bathroom_0003.jpg}}
			\includegraphics[trim=.256\nIW{} .3255\nIH{} .2745\nIW{} .078\nIH{},clip, width=\scenesTableElemWidth{}]{bathroom_0003.jpg}& 
			\settowidth{\nIW}{\includegraphics{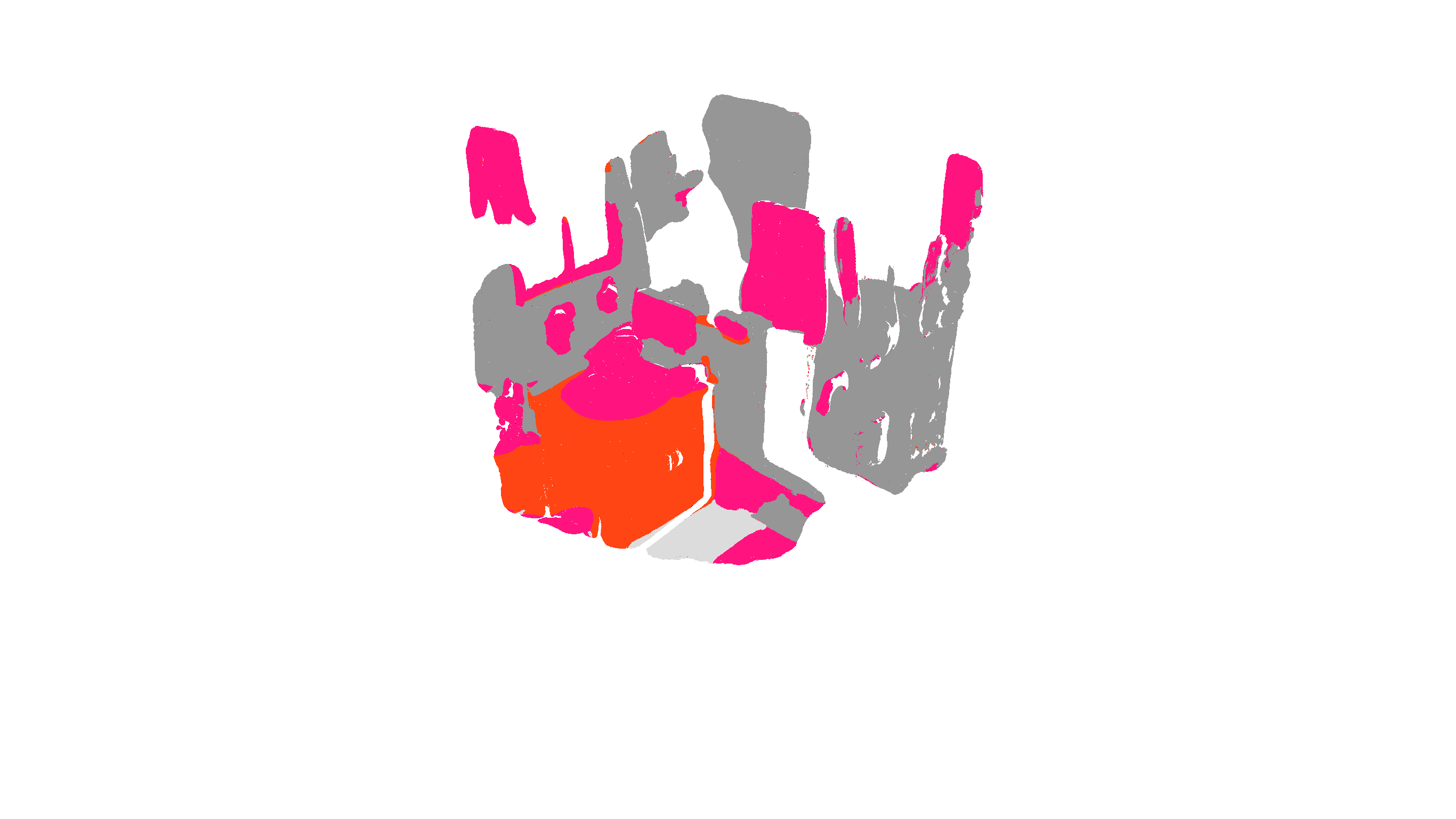}}
			\settoheight{\nIH}{\includegraphics{bathroom_0003_sem.png}}
			\includegraphics[trim=.256\nIW{} .3255\nIH{} .2745\nIW{} .078\nIH{},clip, width=\scenesTableElemWidth{}]{bathroom_0003_sem.png}&\\   

			\settowidth{\nIW}{\includegraphics{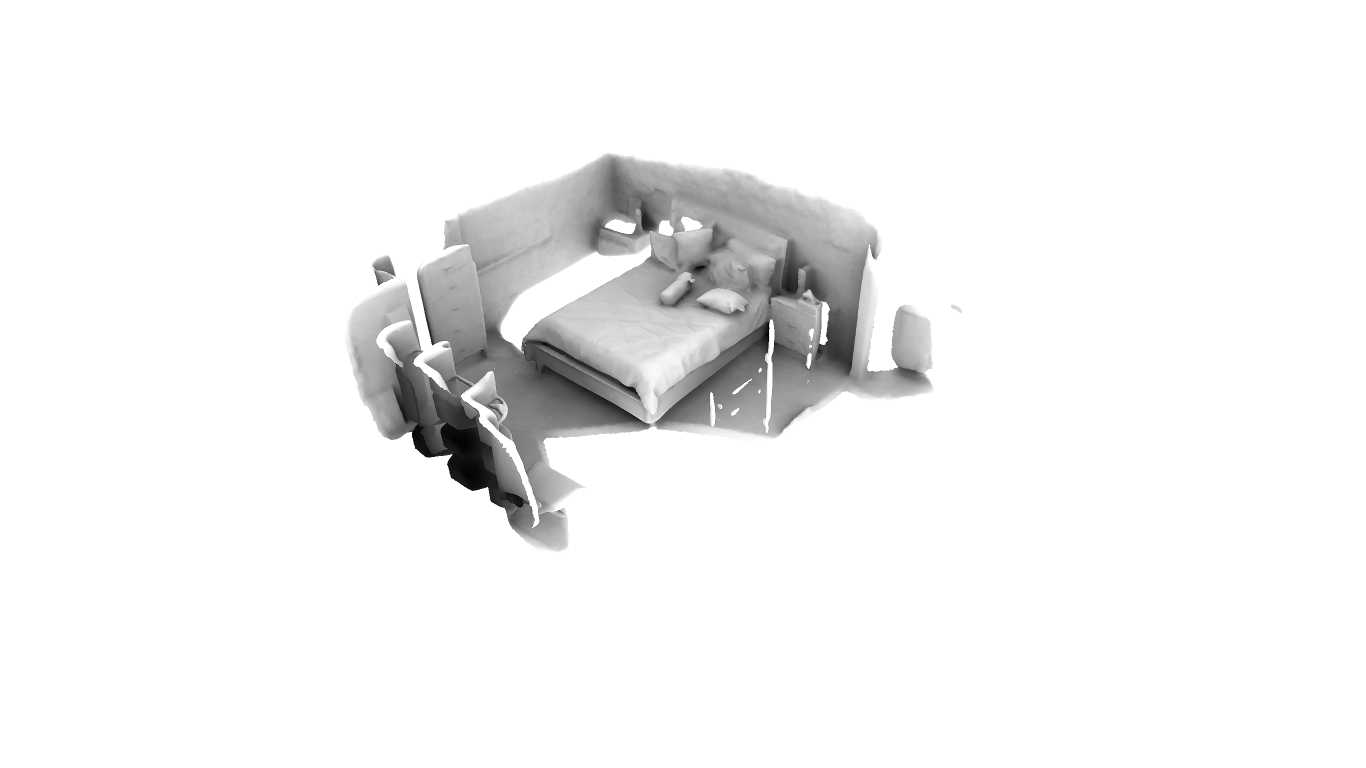}}
			\settoheight{\nIH}{\includegraphics{bedroom_0109_mesh.jpg}}
			\includegraphics[trim=.183\nIW{} .3255\nIH{} .256\nIW{} .1432\nIH{},clip, width=\scenesTableElemWidth{}]{bedroom_0109_mesh.jpg}& 
			\settowidth{\nIW}{\includegraphics{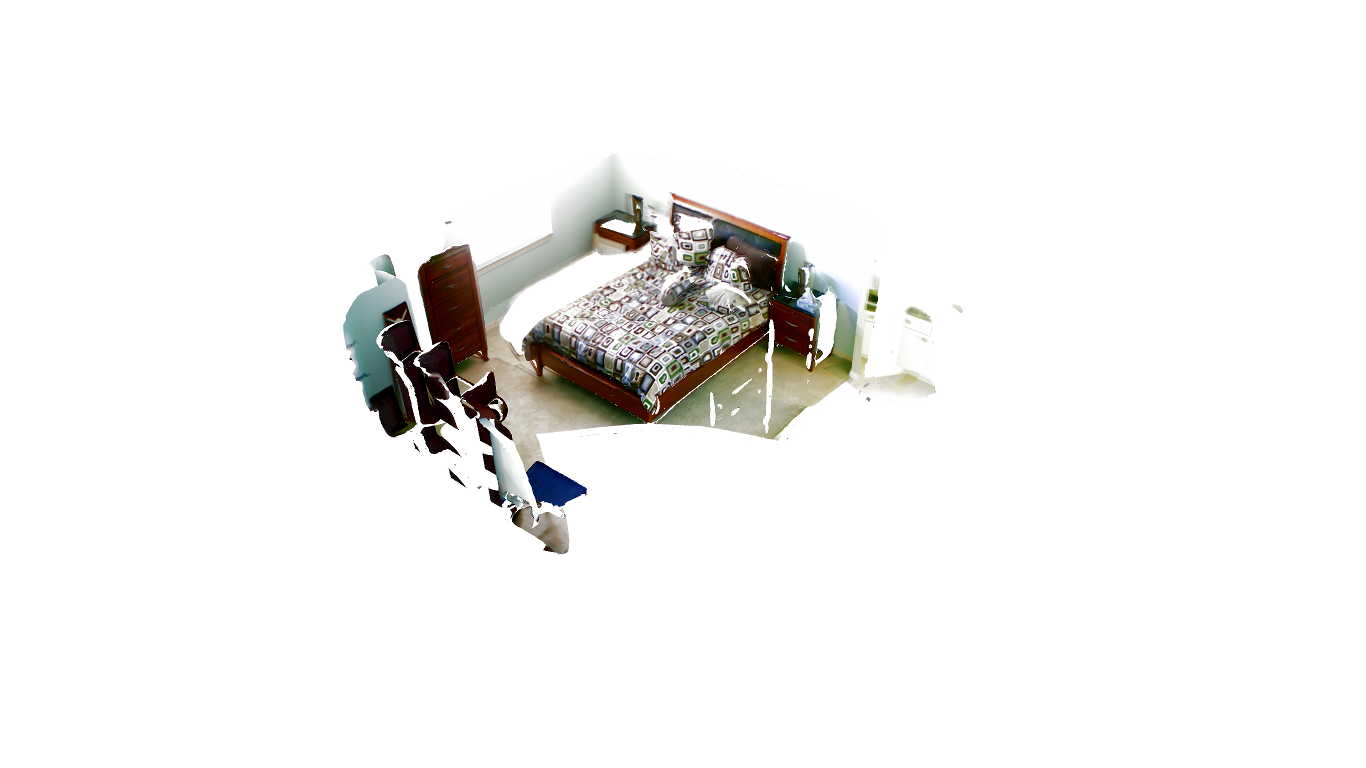}}
			\settoheight{\nIH}{\includegraphics{bedroom_0109.jpg}}
			\includegraphics[trim=.183\nIW{} .3255\nIH{} .256\nIW{} .1432\nIH{},clip, width=\scenesTableElemWidth{}]{bedroom_0109.jpg}& 
			\settowidth{\nIW}{\includegraphics{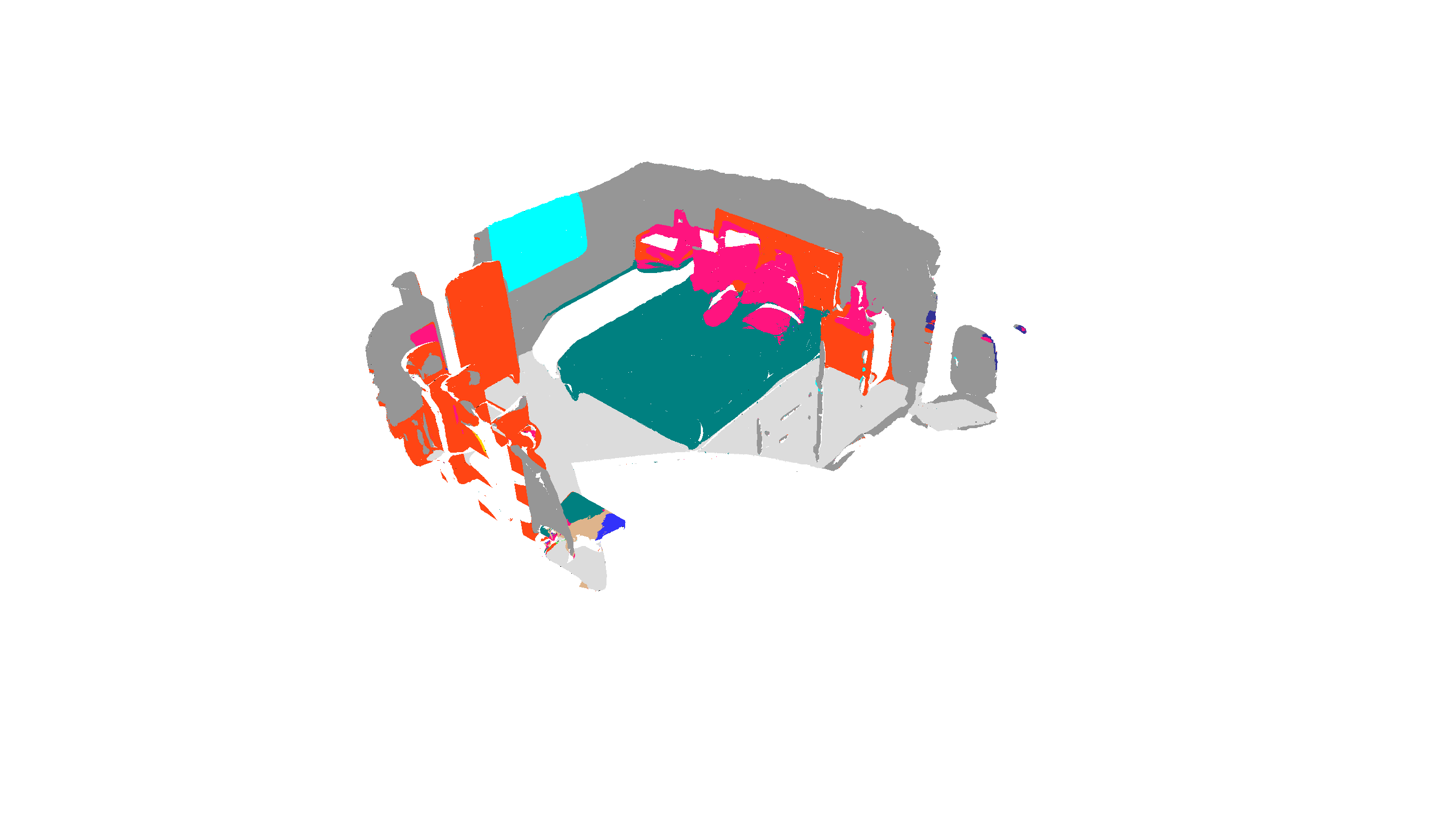}}
			\settoheight{\nIH}{\includegraphics{bedroom_0109_sem.png}}
			\includegraphics[trim=.183\nIW{} .3255\nIH{} .256\nIW{} .1432\nIH{},clip, width=\scenesTableElemWidth{}]{bedroom_0109_sem.png}&\\   

			\settowidth{\nIW}{\includegraphics{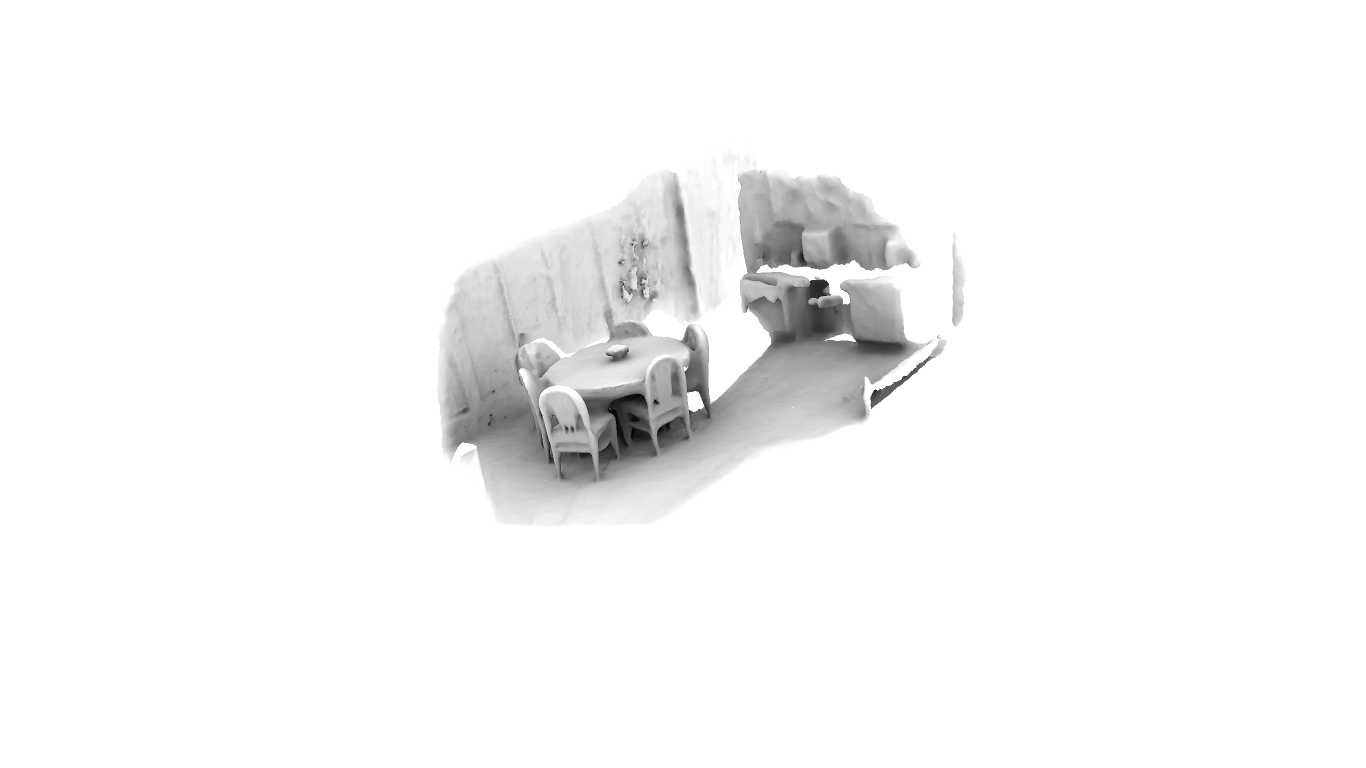}}
			\settoheight{\nIH}{\includegraphics{kitchen_0026b_mesh.jpg}}
			\includegraphics[angle=-8,trim=.311\nIW{} .3255\nIH{} .2928\nIW{} .208\nIH{},clip, width=\scenesTableElemWidth{}]{kitchen_0026b_mesh.jpg}& 
			\settowidth{\nIW}{\includegraphics{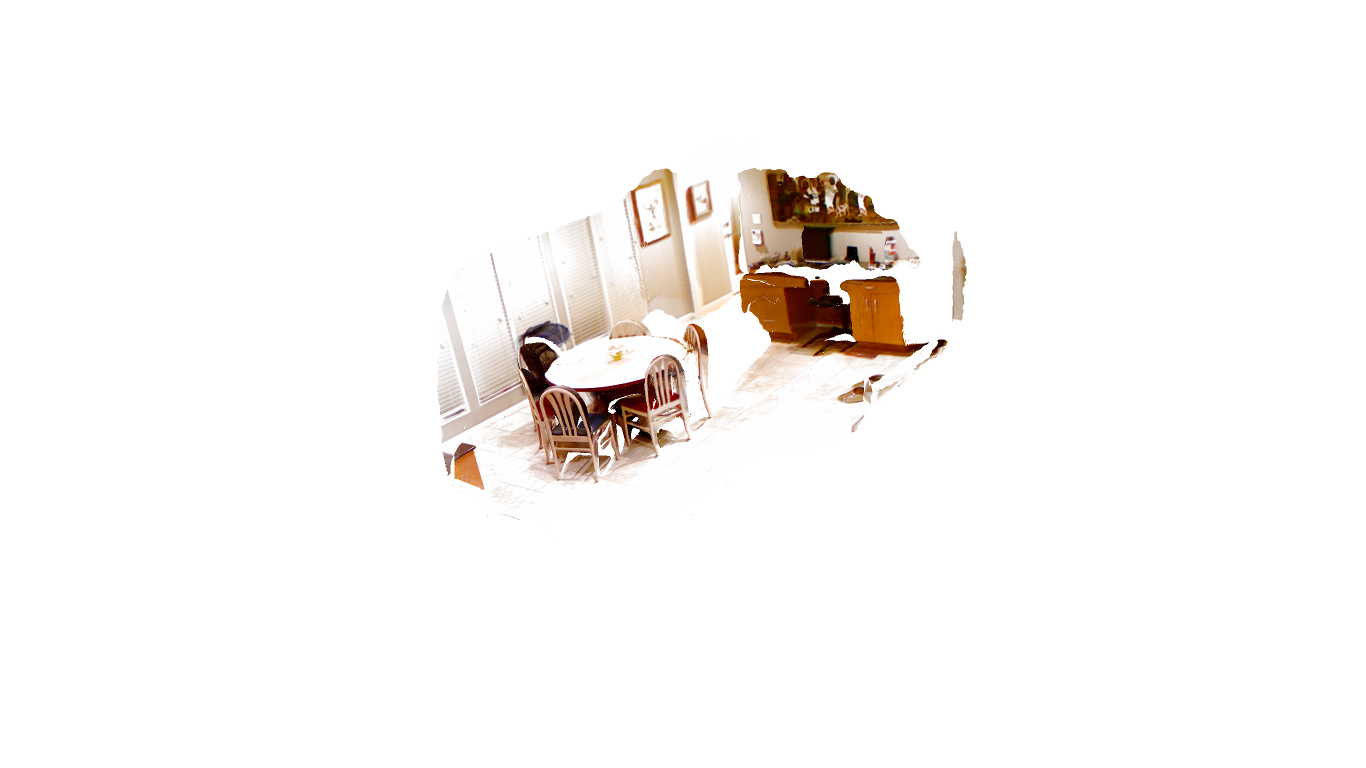}}
			\settoheight{\nIH}{\includegraphics{kitchen_0026b.jpg}}
			\includegraphics[angle=-8,trim=.311\nIW{} .3255\nIH{} .2928\nIW{} .208\nIH{},clip, width=\scenesTableElemWidth{}]{kitchen_0026b.jpg}& 
			\settowidth{\nIW}{\includegraphics{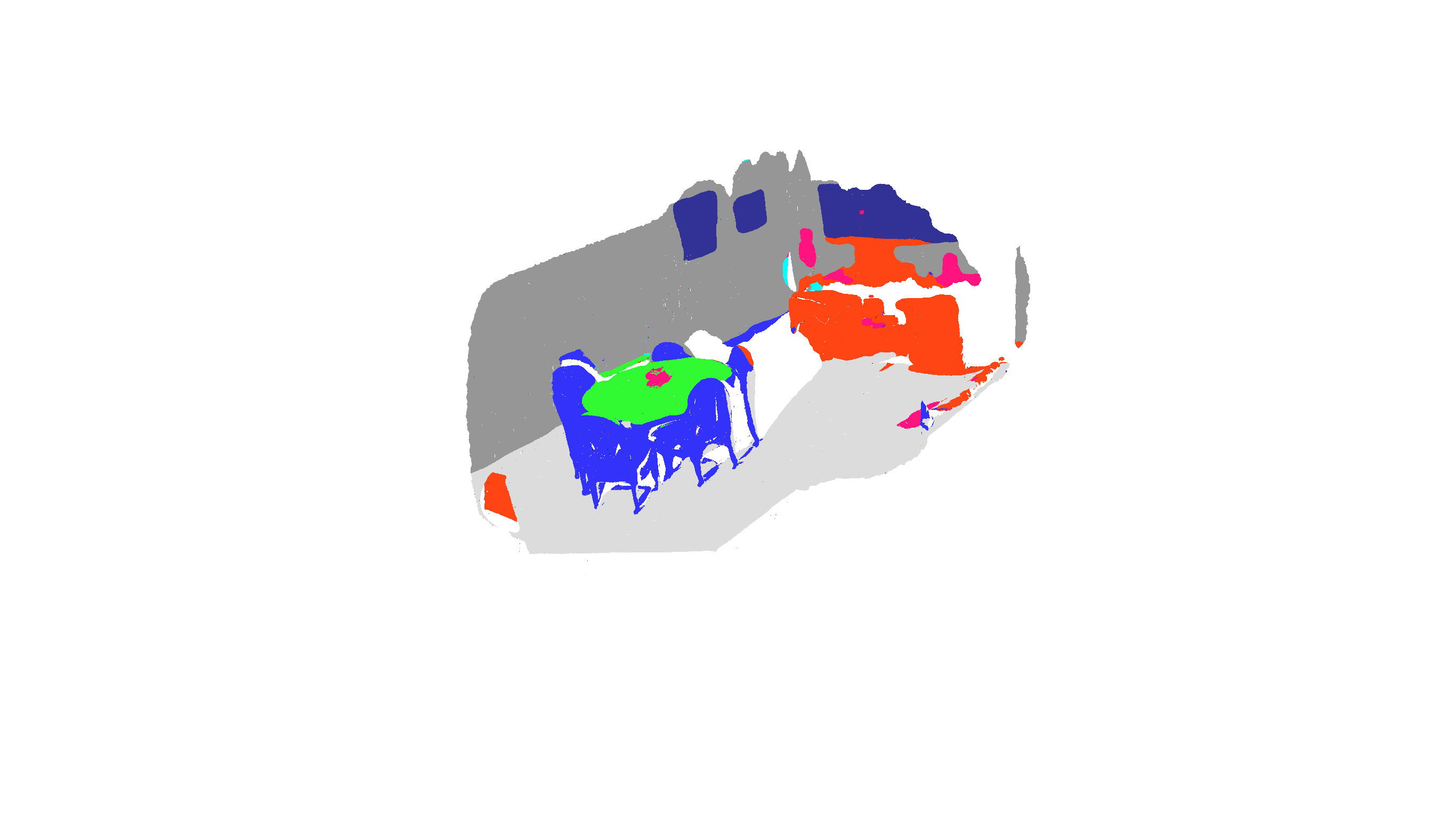}}
			\settoheight{\nIH}{\includegraphics{kitchen_0026b_sem.png}}
			\includegraphics[angle=-8,trim=.311\nIW{} .3255\nIH{} .2928\nIW{} .208\nIH{},clip, width=\scenesTableElemWidth{}]{kitchen_0026b_sem.png}&\\   

		\end{tabular}
		\caption{Scenes from NYUv2: The mesh (left) is reconstructed from the surfel map of ElasticFusion and textured with RGB appearance (middle) and semantic labels (right).}
		\label{fig:NYUreconstruction}
	\end{figure*}
	
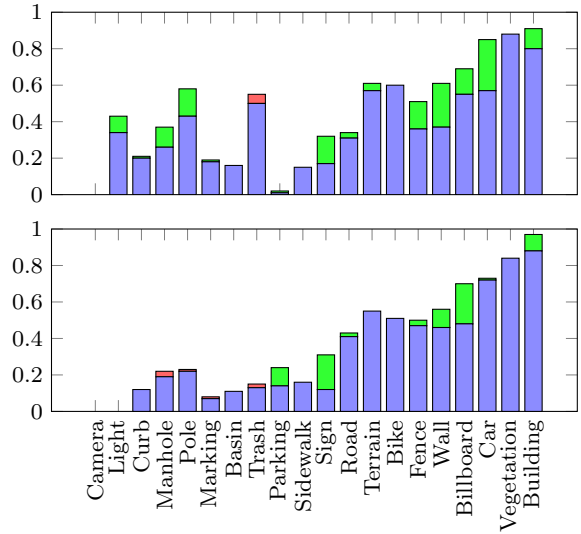
\begin{figure}
		\centering
		\begin{tikzpicture}
		\begin{axis}[
		ybar stacked,   
		ymin=0,         
		ymax=1,
		ytick={0,0.2,0.4,0.6,0.8,1},
		xtick=data,     
		legend style={at={(axis cs:65,0.2)},anchor=south west},
		xticklabels from table={lbh_single_frame_iou_increase_test.txt}{class},  
		xticklabel style = {rotate=90,anchor=east},
		xticklabels={,,}, 
		bar width=0.75,
		width=8.5cm,
		height=4cm
		]
		\addplot [fill=blue!45] table [y=min_iou, meta=class,x expr=\coordindex] {lbh_single_frame_iou_increase_test.txt};   
		\addplot [fill=green!80] table [y=iou_increase, meta=class,x expr=\coordindex] {lbh_single_frame_iou_increase_test.txt};
		\addplot [fill=red!60,
		] table [y=iou_decrease, meta=class,x expr=\coordindex] {lbh_single_frame_iou_increase_test.txt};
		\end{axis}
		\end{tikzpicture}

		\begin{tikzpicture}
		\begin{axis}[
		ybar stacked,   
		ymin=0,         
		ymax=1,
		ytick={0,0.2,0.4,0.6,0.8,1},
		xtick=data,     
		legend style={at={(axis cs:65,0.2)},anchor=south west},
		xticklabels from table={lbh_fused_iou_increase.txt}{class},  
		xticklabel style = {rotate=90,anchor=east},
		bar width=0.75,
		width=8.5cm,
		height=4cm
		]
		\addplot [fill=blue!45] table [y=min_iou, meta=class,x expr=\coordindex] {lbh_fused_iou_increase.txt};   
		\addplot [fill=green!80] table [y=iou_increase, meta=class,x expr=\coordindex] {lbh_fused_iou_increase.txt};
		\addplot [fill=red!60,
		] table [y=iou_decrease, meta=class,x expr=\coordindex] {lbh_fused_iou_increase.txt};
		\end{axis}
		\end{tikzpicture} 
	\caption{Courtyard results: We compare our method (bottom) against single-frame predictions (top). The per class IoU is denoted by blue bars. An increase in IoU due to Label Propagation is marked in green, a decrease in red. For single-frame we mask out the landing gear of the MAV and the areas which are not covered by the mesh.}
	\label{fig:LBHiou}
	\end{figure}

		We also conduct accuracy experiments on the courtyard dataset for which we densely labeled \num{48} frames around the scene. For fairness we labeled sky as background due to missing representation within the mesh.
		\reffig{fig:LBHiou} shows that retraining using label propagation greatly improves the accuracy for most classes. However, an interesting observation from this experiment is that the single-frame predictions have on average higher accuracy than the fused semantics from the mesh. This is due to the fact that both the camera poses and the mesh are imperfect, hence fusing the information from various points of view may lead to discrepancies. This limitation is further reinforced by the fact that the classes which experience a higher drop in accuracy from the fusing process are those which are spatially small (lane-markings, poles, and street lights), while broader classes like building and vegetation remain largely unaffected by errors from the scene reconstruction. 
		For this reason we conduct further experiments to evaluate the impact of misalignments in the following \refsec{sec:RegRob}.
		Nevertheless, we can conclude that label propagation grants a net improvement in the semantic accuracy, increasing the mean IoU for single-frame prediction by \SI{7}{\percent} and for the fused information by \SI{3}{\percent}.
		\reffig{fig:LBHcompare} shows a visual comparison of the semantics using various view points from the courtyard. The copters landing gear and rotor arm visible within camera images are masked out prior to evaluation. 
		A partial failure case is shown in the first row. The thin lane-markings are actually degraded through fusion and LP. We trace this back to inaccurate manual extrinsic and temporal calibration, since the corresponding single-frames continuously contained the lane-markings. Still, we observe improvements after LP on the container next to the service station.
		The second view was captured from behind the same service station. Insufficiency in the meshing process created only the top of the pole (front, left), which is correctly classified, but not connected to the ground plane reducing the overall IoU compared towards single-frame predictions. Further improvements through LP are especially visible in the last two rows. The left window is classified as a sign prior to retraining and a large portion of the sidewalk was incorrect. Also the building in the background is improved. The rooftop (third row) presents a unique novel view that is largely misclassified in the single-frame. Our mesh-based fusion improves the result as expected and allows successful retraining.

		\definecolor{M_Curb}{RGB}{196, 196, 196}
		\definecolor{M_Fence}{RGB}{190, 153, 153}
		\definecolor{M_Wall}{RGB}{102, 102, 156}
		\definecolor{M_Parking}{RGB}{250, 170, 160}
		\definecolor{M_Road}{RGB}{128, 64, 128}
		\definecolor{M_Sidewalk}{RGB}{244, 35, 232}
		\definecolor{M_Building}{RGB}{70, 70, 70}
		\definecolor{M_Marking}{RGB}{255, 255, 255}
		\definecolor{M_Terrain}{RGB}{152, 251, 152}
		\definecolor{M_Vegetation}{RGB}{107, 142, 35}
		\definecolor{M_Billboard}{RGB}{220, 220, 220}
		\definecolor{M_Basin}{RGB}{170, 170, 170}
		\definecolor{M_Camera}{RGB}{222, 40, 40}
		\definecolor{M_Manhole}{RGB}{170, 170, 170}
		\definecolor{M_Light}{RGB}{250, 170, 30}
		\definecolor{M_Pole}{RGB}{153, 153, 153}
		\definecolor{M_Sign}{RGB}{220, 220, 0}
		\definecolor{M_Trash}{RGB}{180, 165, 180}
		\definecolor{M_Bike}{RGB}{119, 11, 32}
		\definecolor{M_Car}{RGB}{0, 0, 142}
		
		\newcommand{\lbhqualitativeElemWidth}{2.5cm}
		\renewcommand{\tabcolsep}{1pt}
		\begin{figure*}
			\centering
			\begin{tabular}{cccccc}
				RGB image&Ground truth&Single frame& Single with LP &Ours&Ours with LP \\
				\includegraphics[width=\lbhqualitativeElemWidth{}]{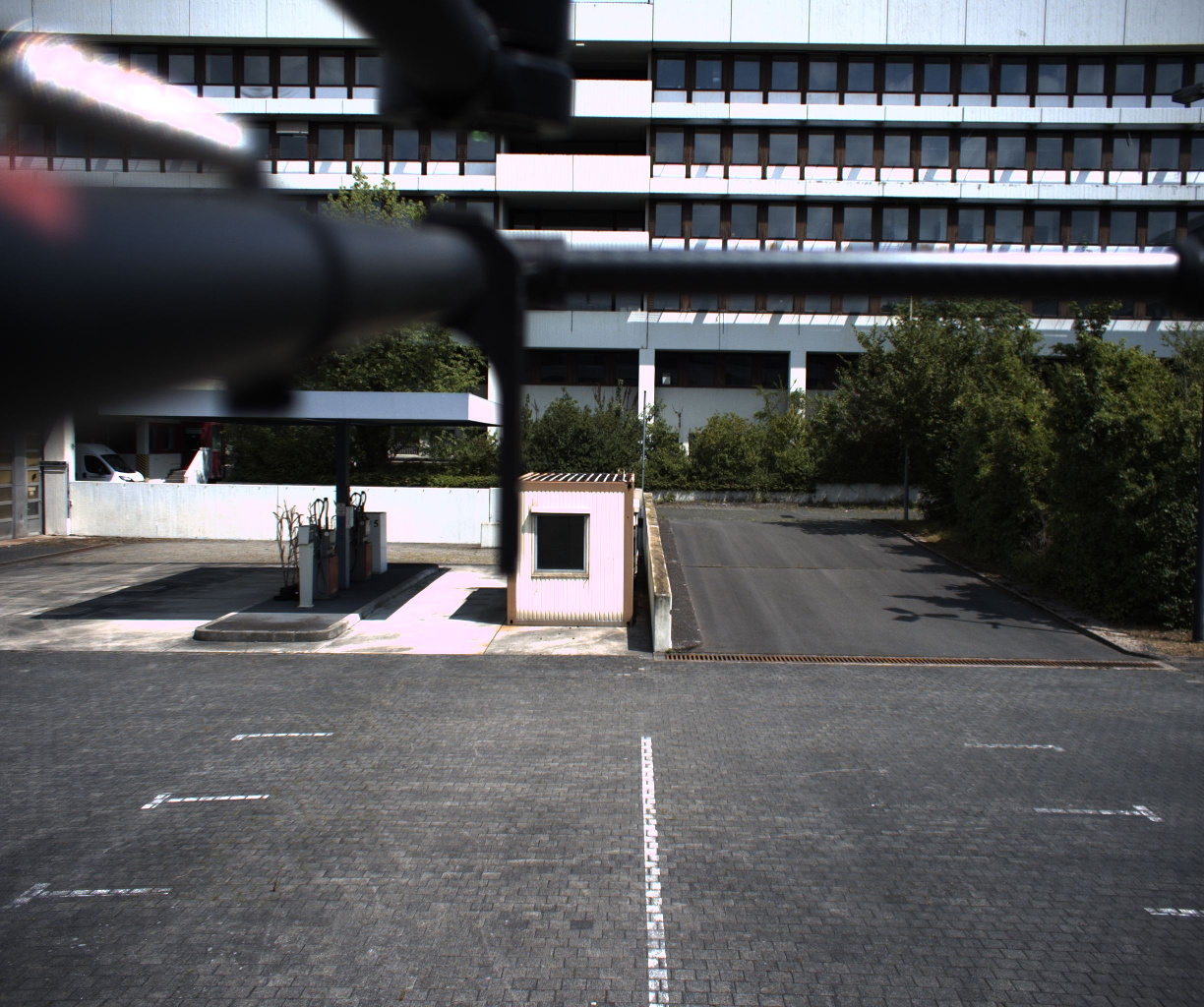}& 
				\includegraphics[width=\lbhqualitativeElemWidth{}]{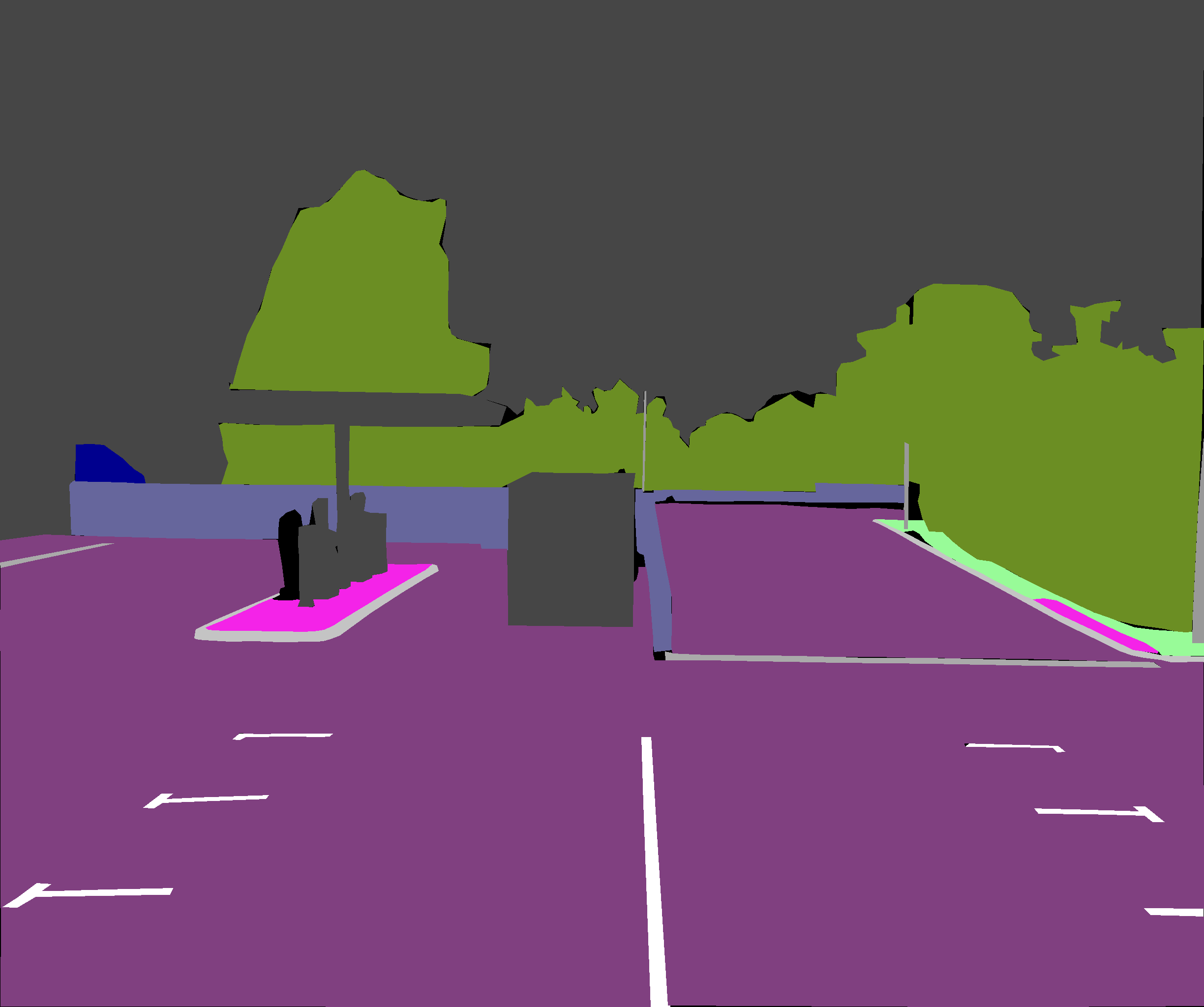}& 
				\includegraphics[width=\lbhqualitativeElemWidth{}]{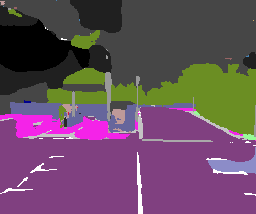}& 
				\includegraphics[width=\lbhqualitativeElemWidth{}]{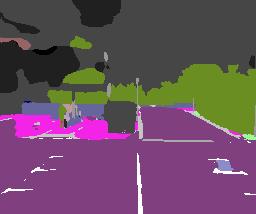}& 
				\includegraphics[width=\lbhqualitativeElemWidth{}]{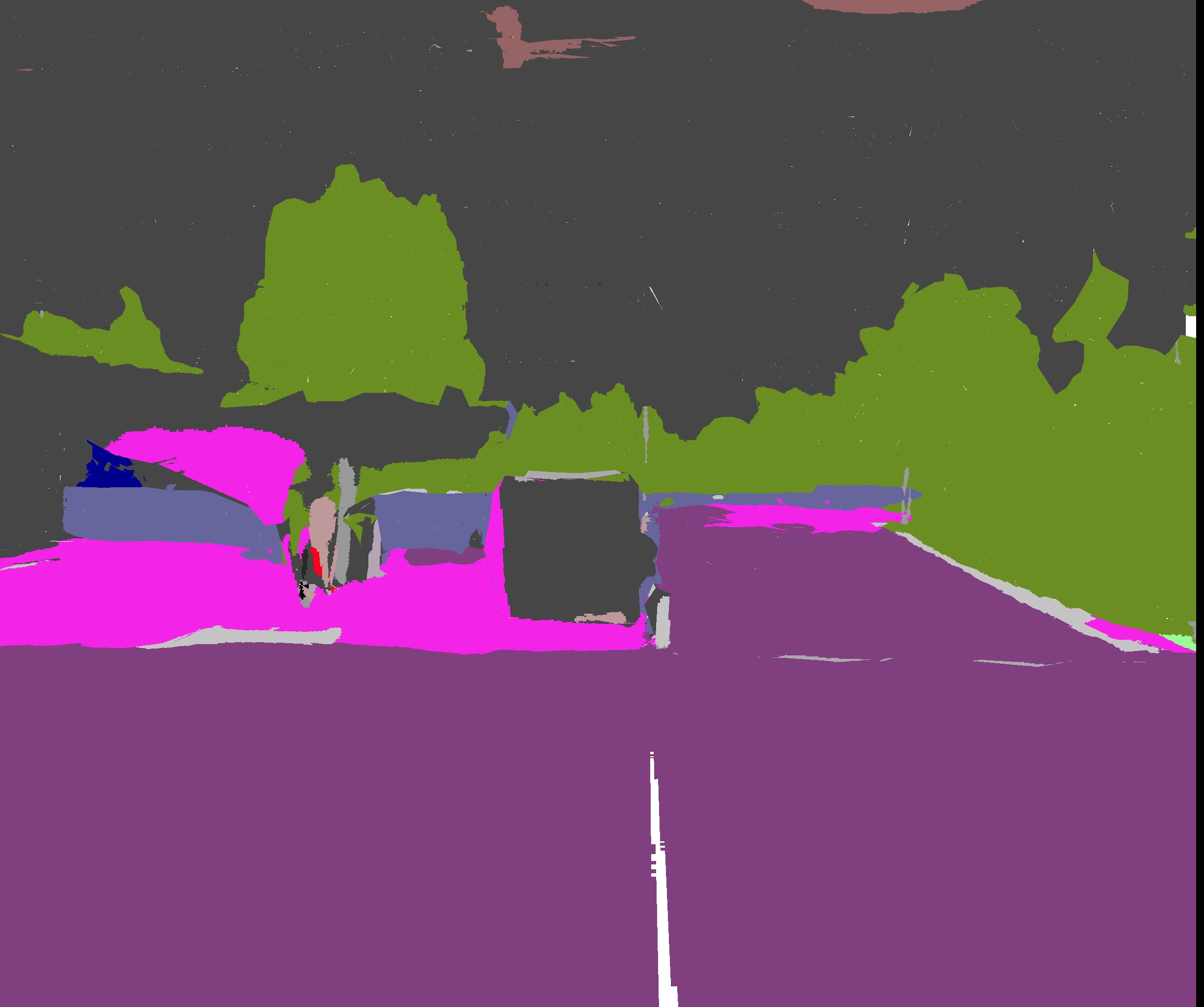}& 
				\includegraphics[width=\lbhqualitativeElemWidth{}]{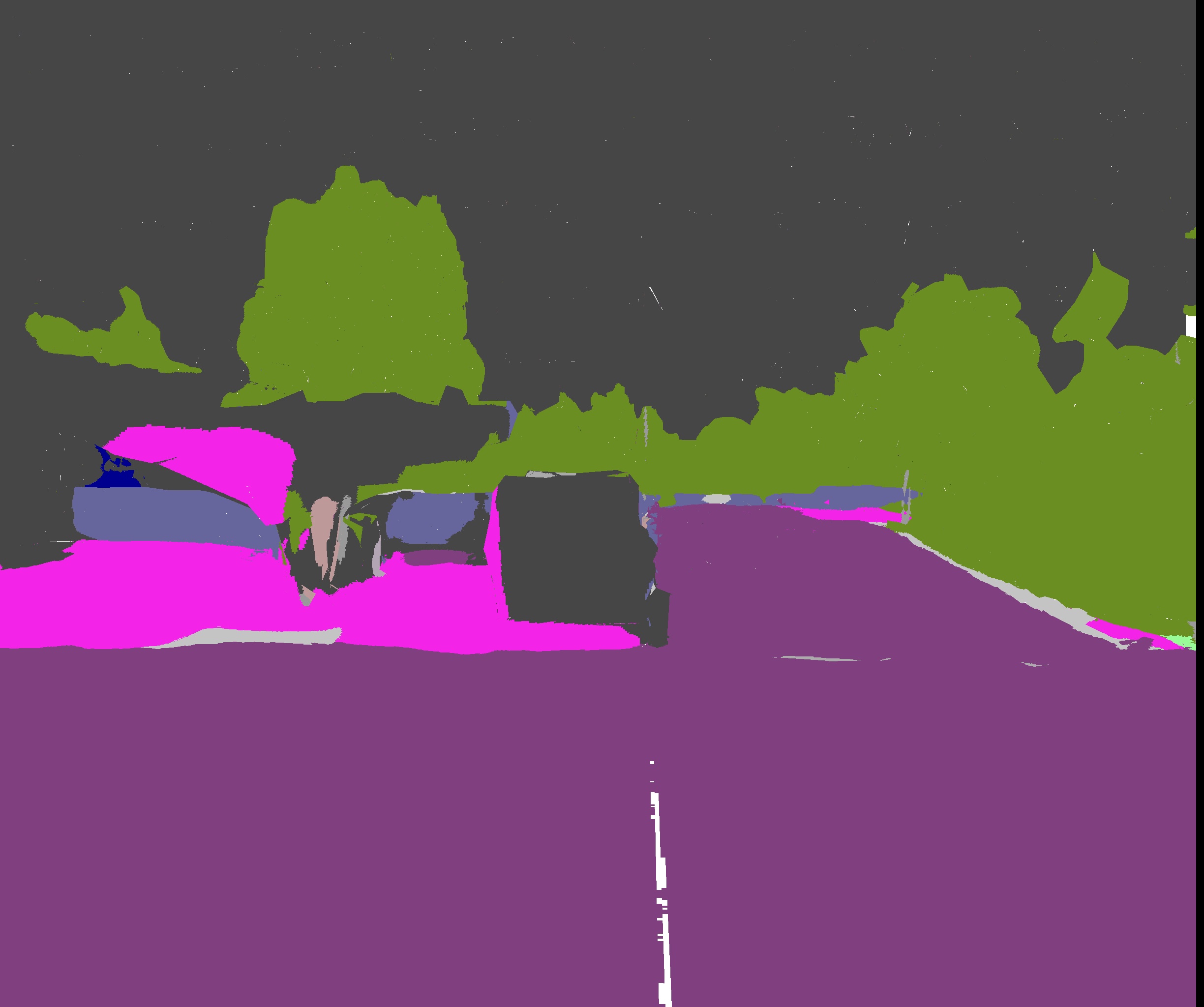}\\ 
				
				\includegraphics[width=\lbhqualitativeElemWidth{}]{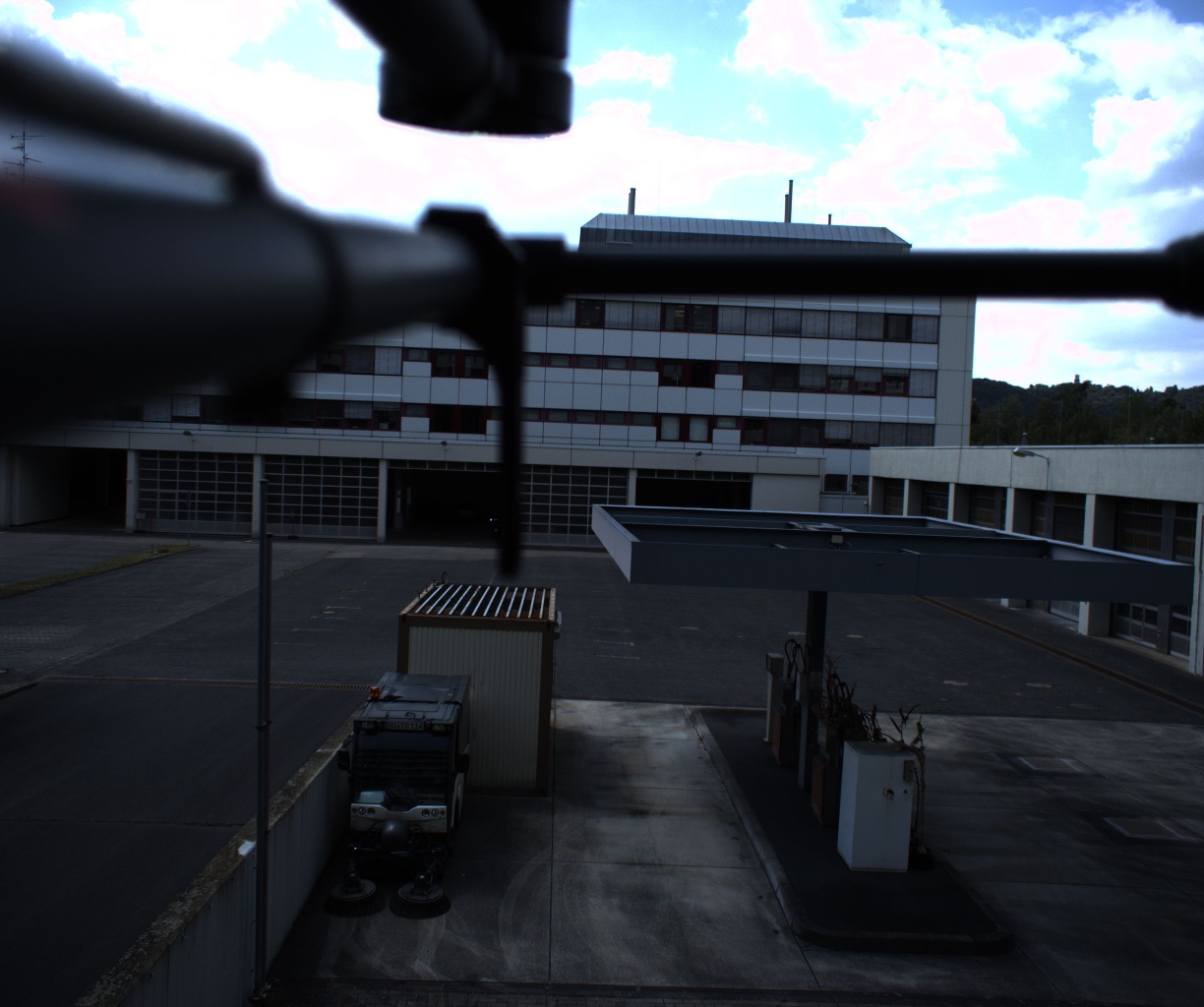}& 
				\includegraphics[width=\lbhqualitativeElemWidth{}]{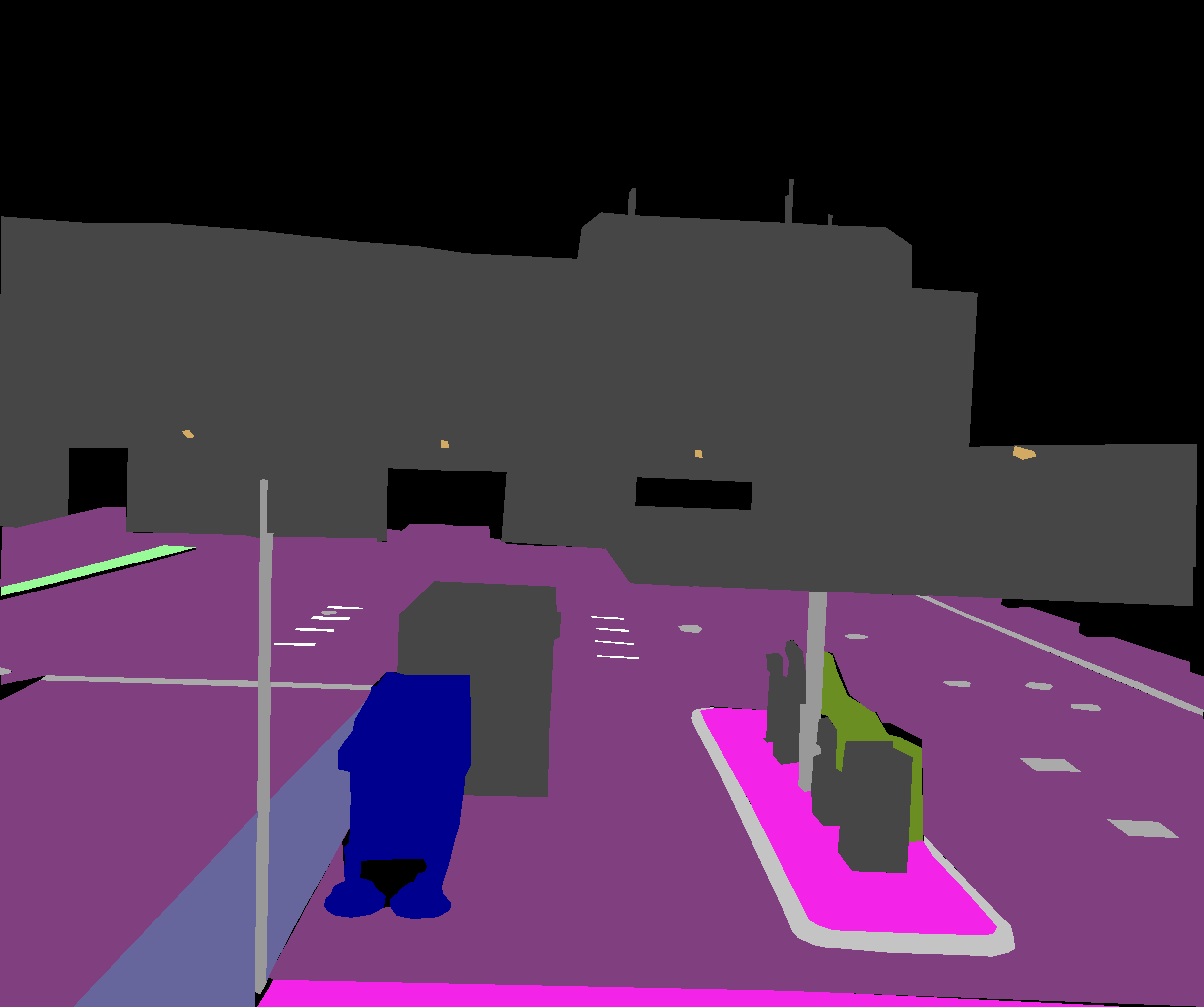}& 
				\includegraphics[width=\lbhqualitativeElemWidth{}]{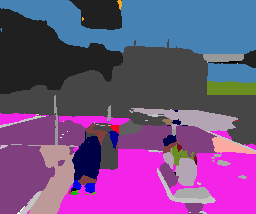}& 
				\includegraphics[width=\lbhqualitativeElemWidth{}]{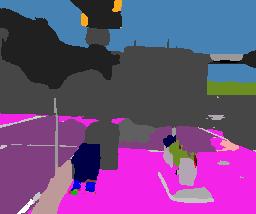}& 
				\includegraphics[width=\lbhqualitativeElemWidth{}]{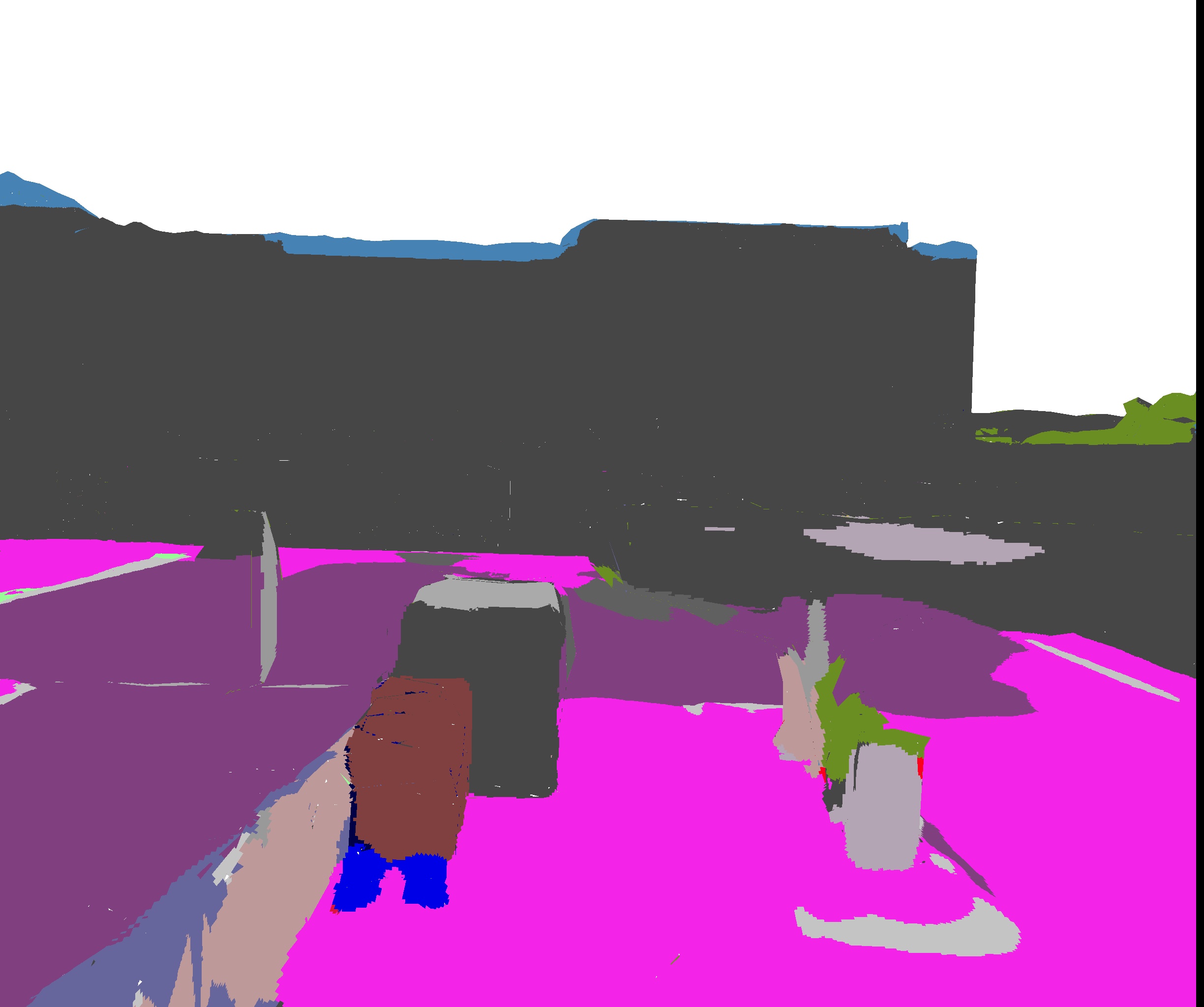}& 
				\includegraphics[width=\lbhqualitativeElemWidth{}]{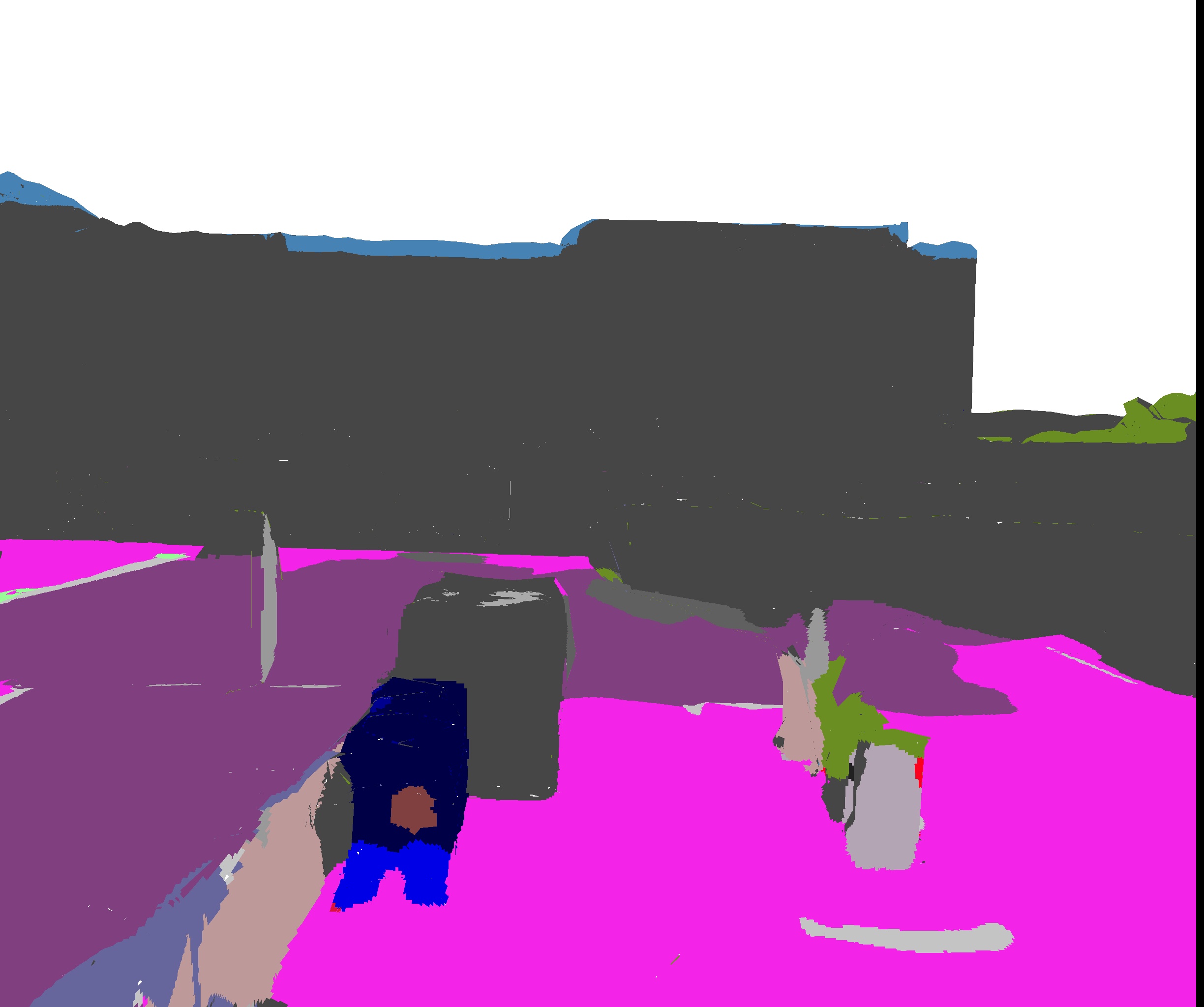}\\ 
									
				\includegraphics[width=\lbhqualitativeElemWidth{}]{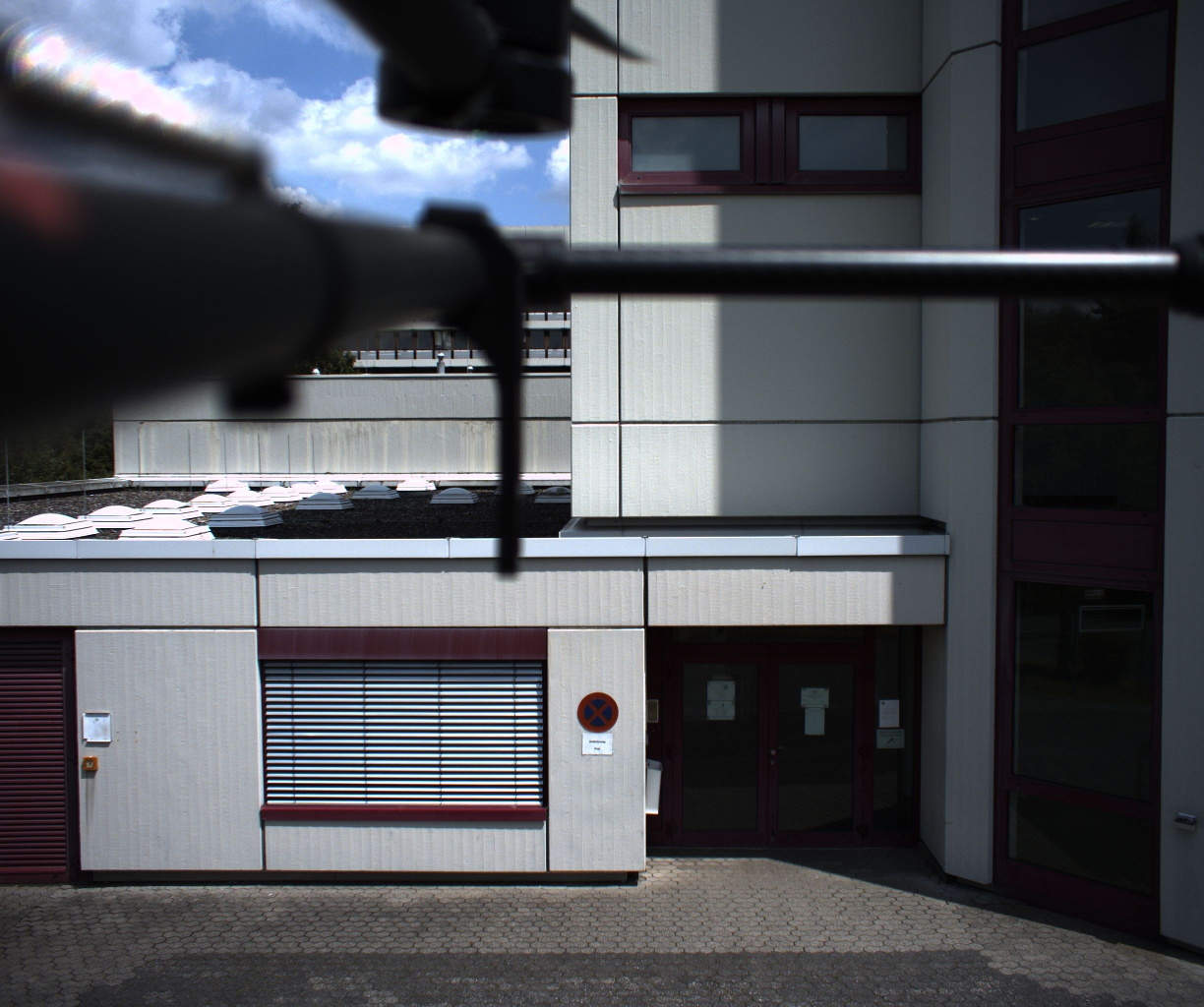}& 
				\includegraphics[width=\lbhqualitativeElemWidth{}]{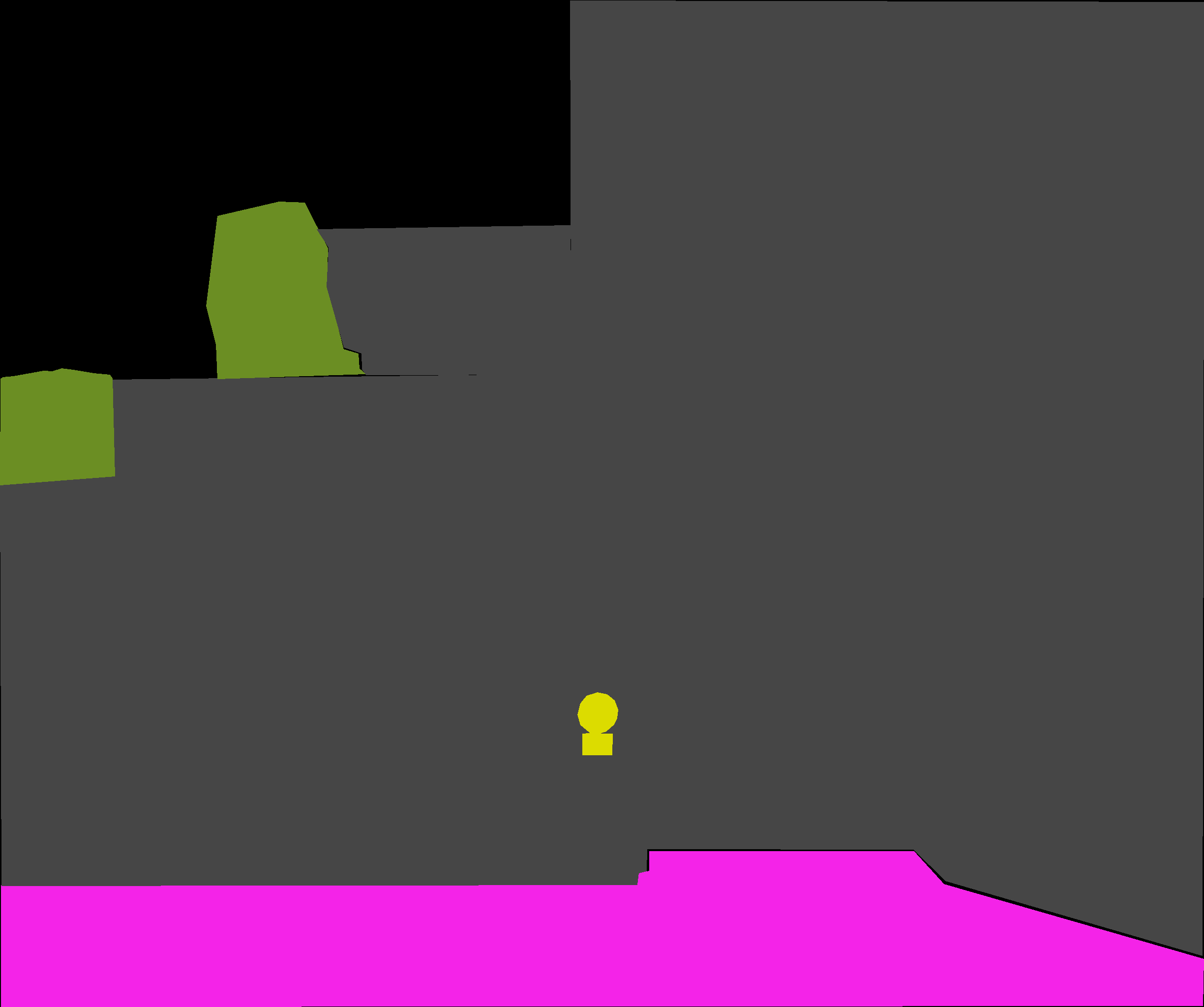}& 
				\includegraphics[width=\lbhqualitativeElemWidth{}]{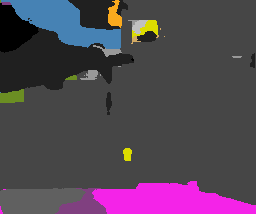}& 
				\includegraphics[width=\lbhqualitativeElemWidth{}]{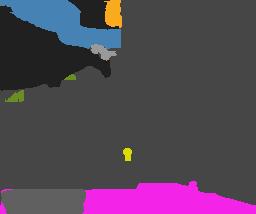}& 
				\includegraphics[width=\lbhqualitativeElemWidth{}]{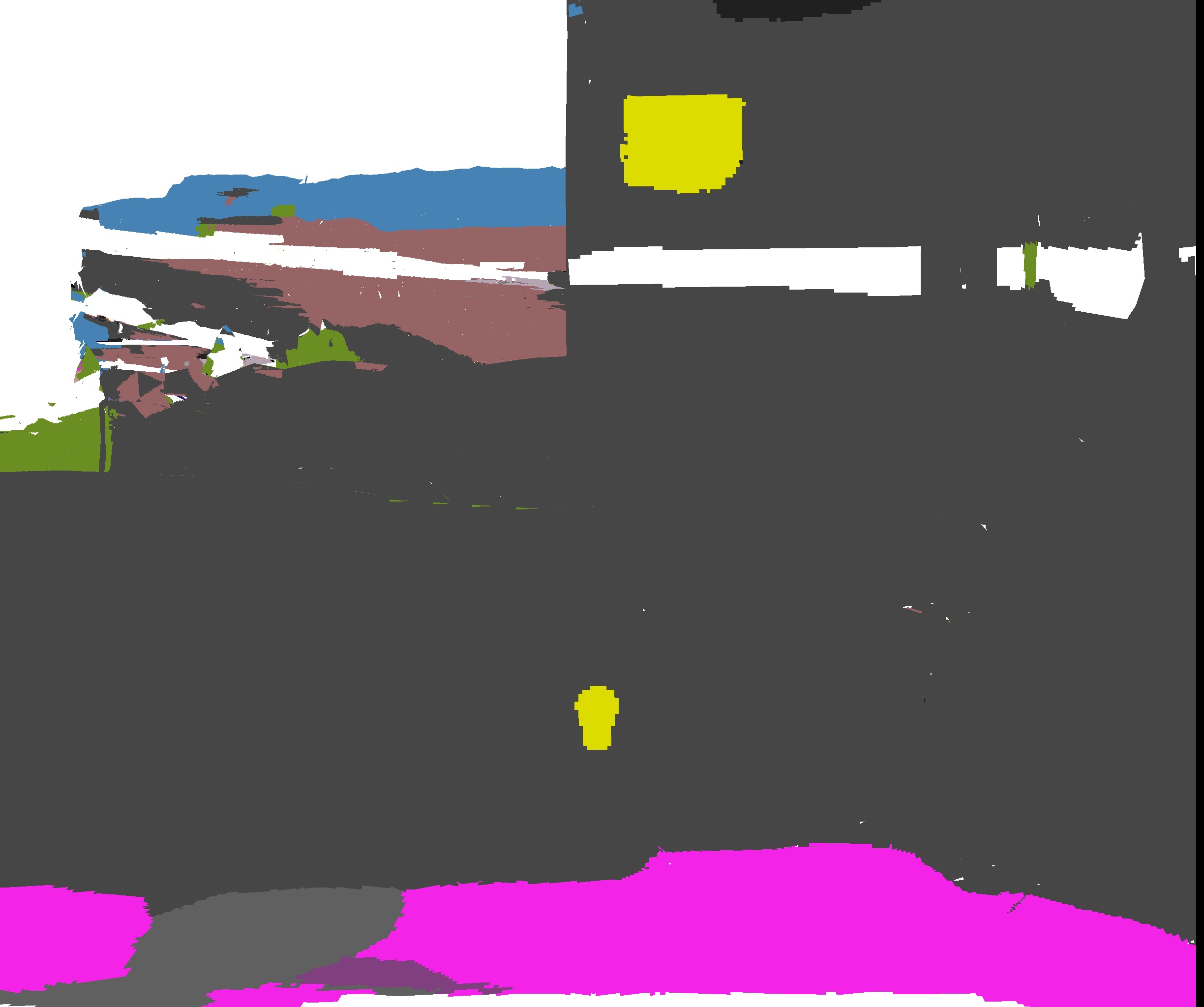}& 
				\includegraphics[width=\lbhqualitativeElemWidth{}]{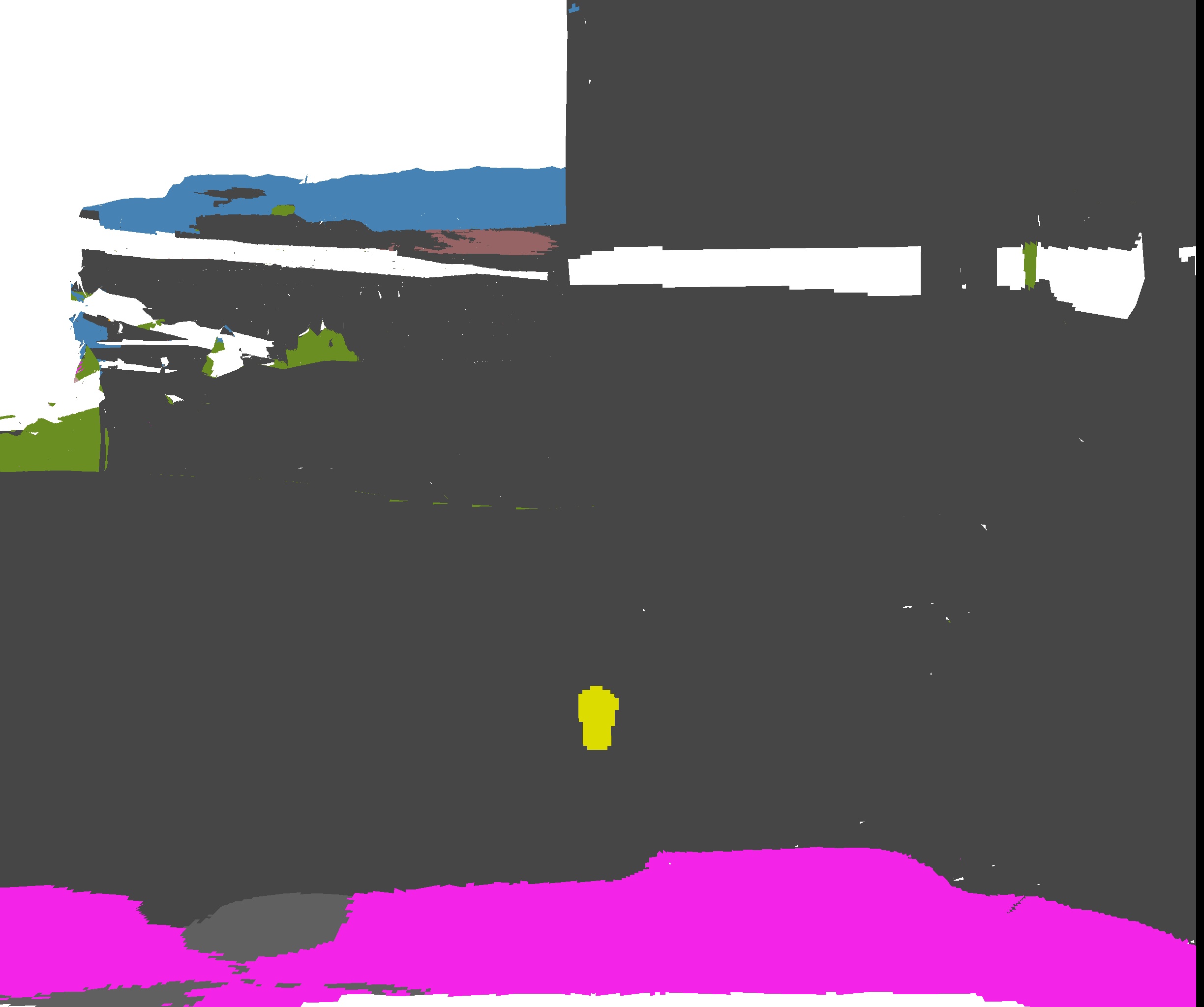}\\ 
									
				\includegraphics[width=\lbhqualitativeElemWidth{}]{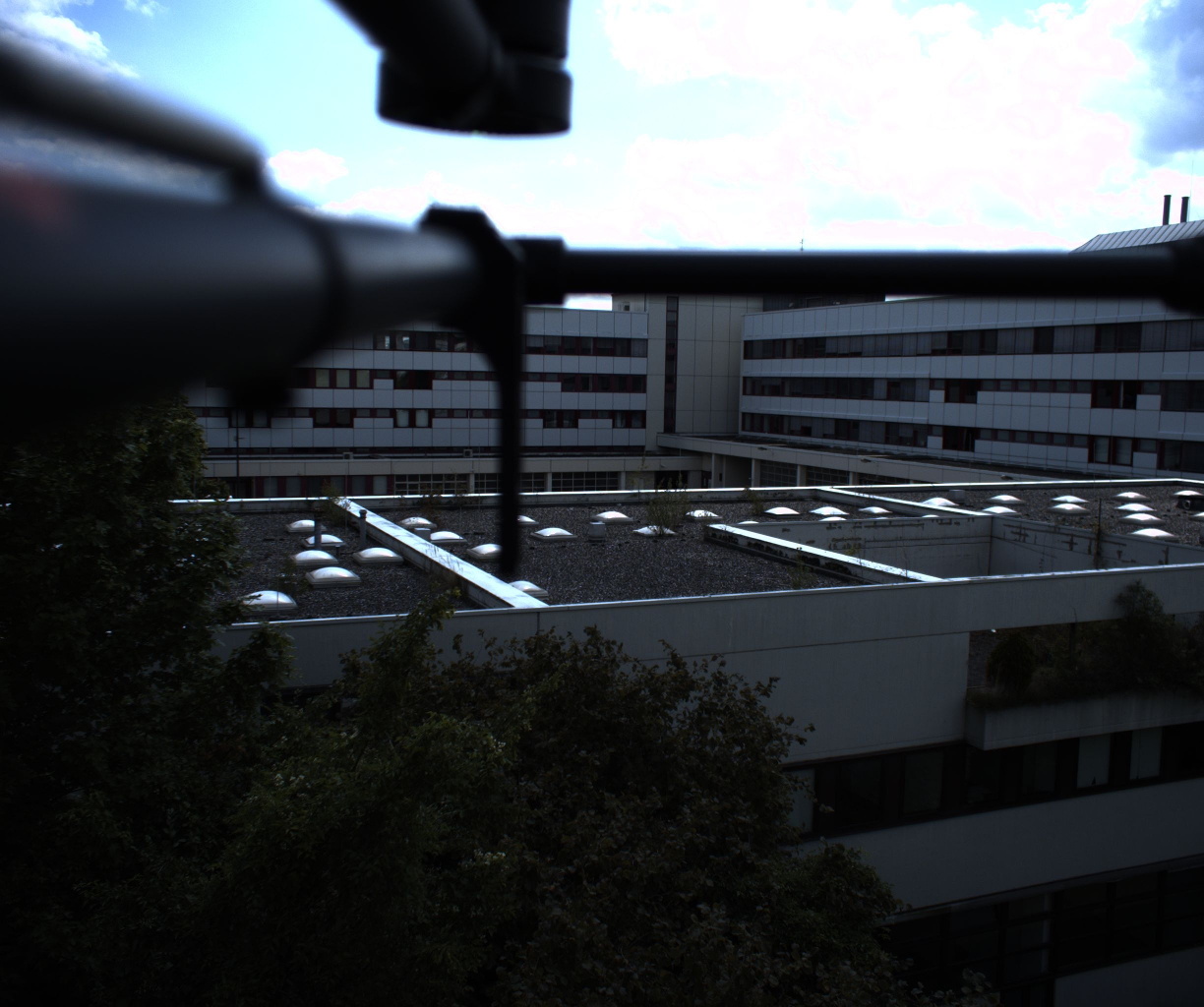}& 
				\includegraphics[width=\lbhqualitativeElemWidth{}]{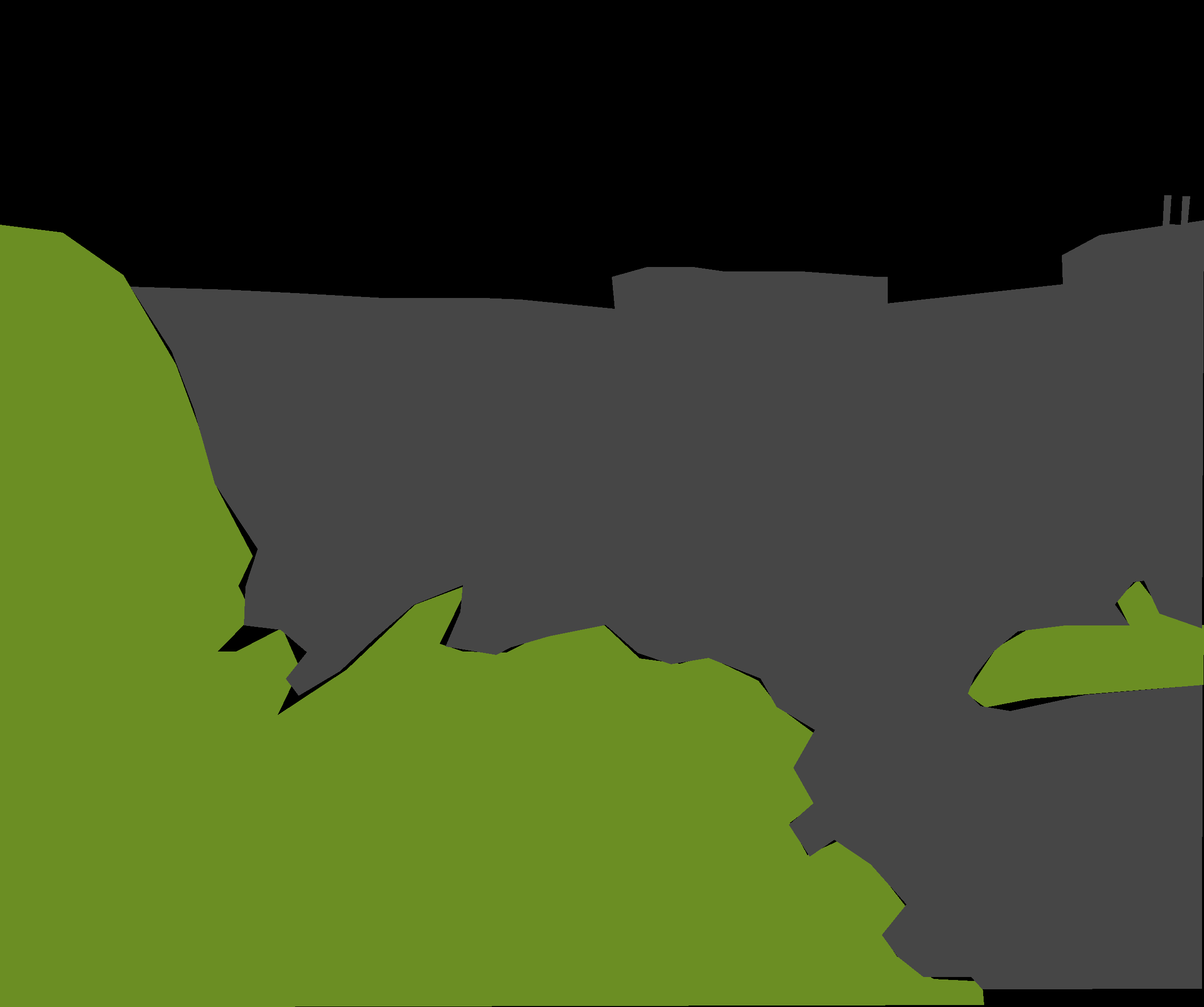}& 
				\includegraphics[width=\lbhqualitativeElemWidth{}]{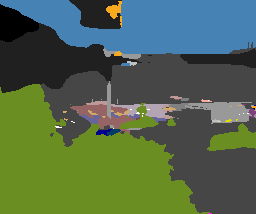}& 
				\includegraphics[width=\lbhqualitativeElemWidth{}]{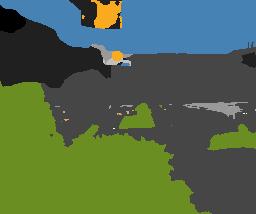}& 
				\includegraphics[width=\lbhqualitativeElemWidth{}]{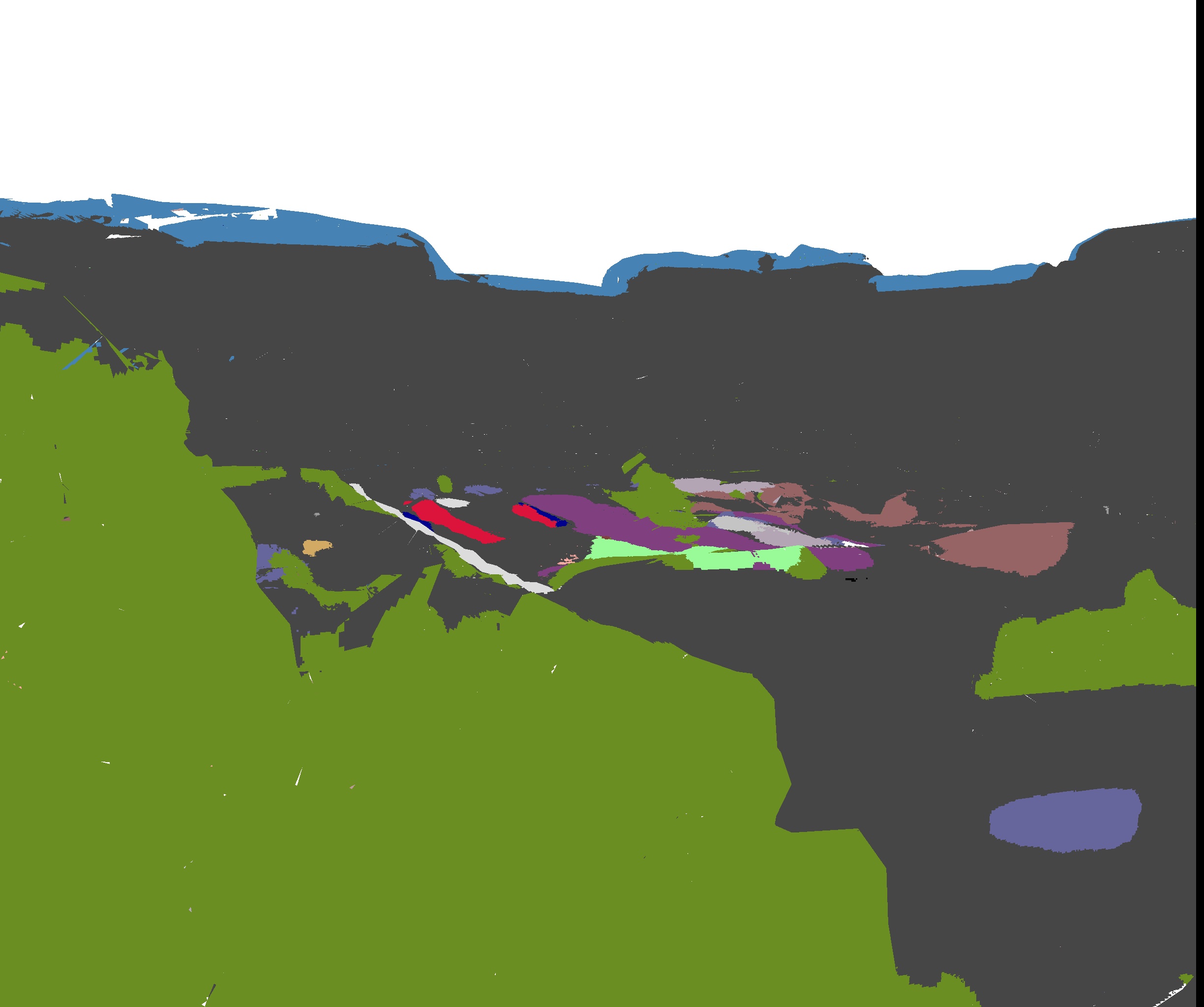}& 
				\includegraphics[width=\lbhqualitativeElemWidth{}]{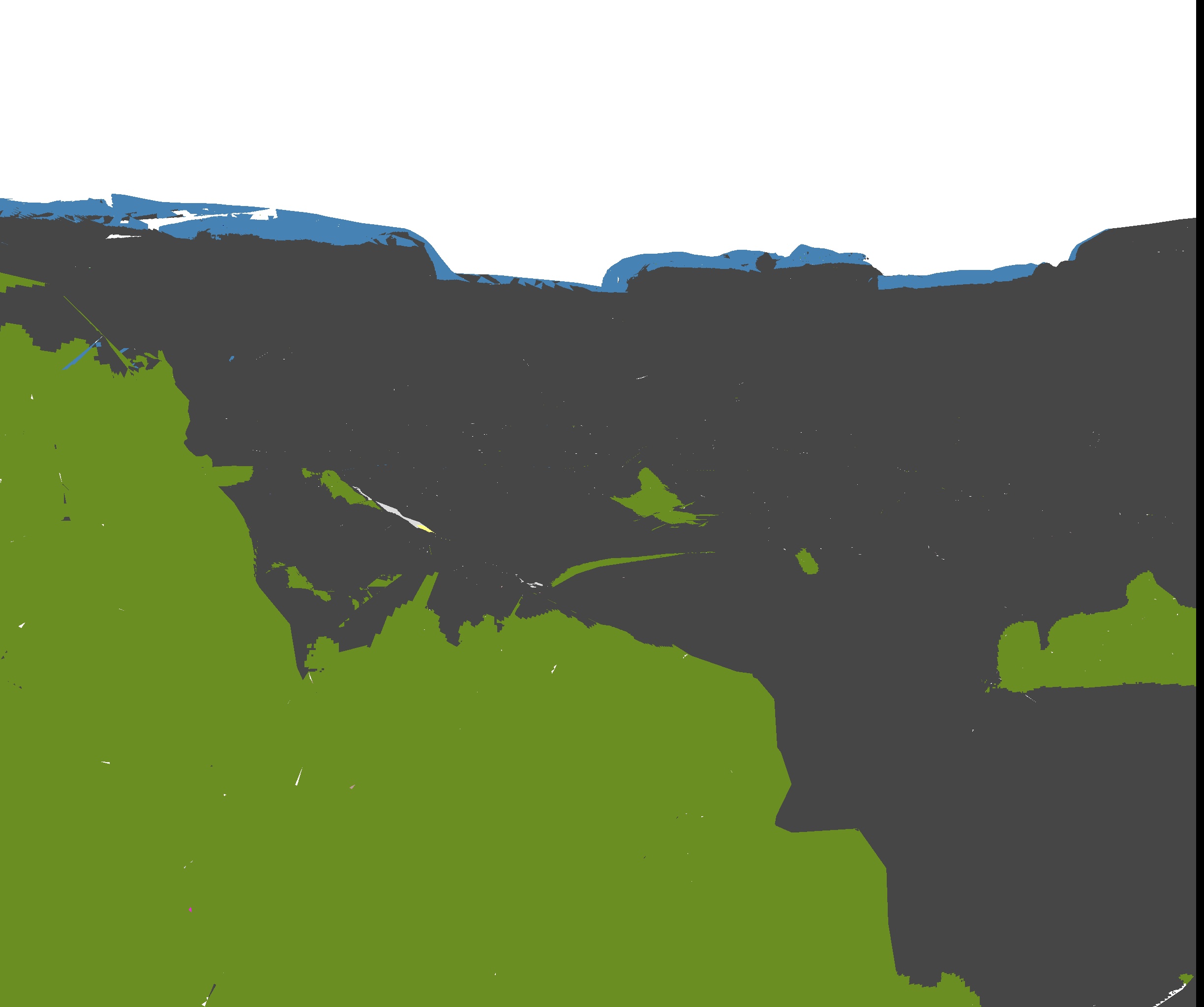}\\ 
						
			\end{tabular}
			\caption{Courtyard qualitative results: We compare our method against single-frame with and without the label propagated and retrained RefineNet.}
			\label{fig:LBHcompare}
		\end{figure*}

	\subsection{Registration Robustness} \label{sec:RegRob}
		Mapping with known poses always raises the question of how robust the system is regarding misalignment. For this we perform experiments on the synthetic Synthia dataset~\citep{ros2016synthia} which provides ground truth poses, depth and semantics. We add random noise to the poses in order to observe the effect on the accuracy of the semantic map. We chose the \emph{seq\_4\_summer} scene, due to the low number of dynamic objects. We aggregated the depth images and meshed the resulting point cloud (see \reffig{fig:SynthiaView}).
		Synthia provides images from eight cameras arranged in groups of four to create an omnidirectional view-cone. For simplicity, we choose for the reconstruction only the front-facing camera of the left group.
		In order to analyze the behavior of the semantic map under incorrect poses or incorrect calibration between depth and color camera, the IoU is calculated for increasing amounts of noise on the translational~($\leq \SI{0.5}{\metre}$) as well as rotational~($\leq \SI{5}{\degree}$) part of the camera poses. The IoU with increasing portion of noise is visualized per class in \reffig{fig:Misalign} with invisible or dynamic classes being disregarded. We choose to retain the cars as most of them were parked, and hence do not pose a problem for the reconstruction. As expected, the IoU decreases faster for smaller object classes, like poles, lights, and signs, than for buildings or the road.
		In conclusion, as the robotics community moves towards larger and bigger datasets with more semantic classes, the detail of the semantic maps will heavily depend on correct sensor poses.

		\settowidth{\nIW}{\includegraphics{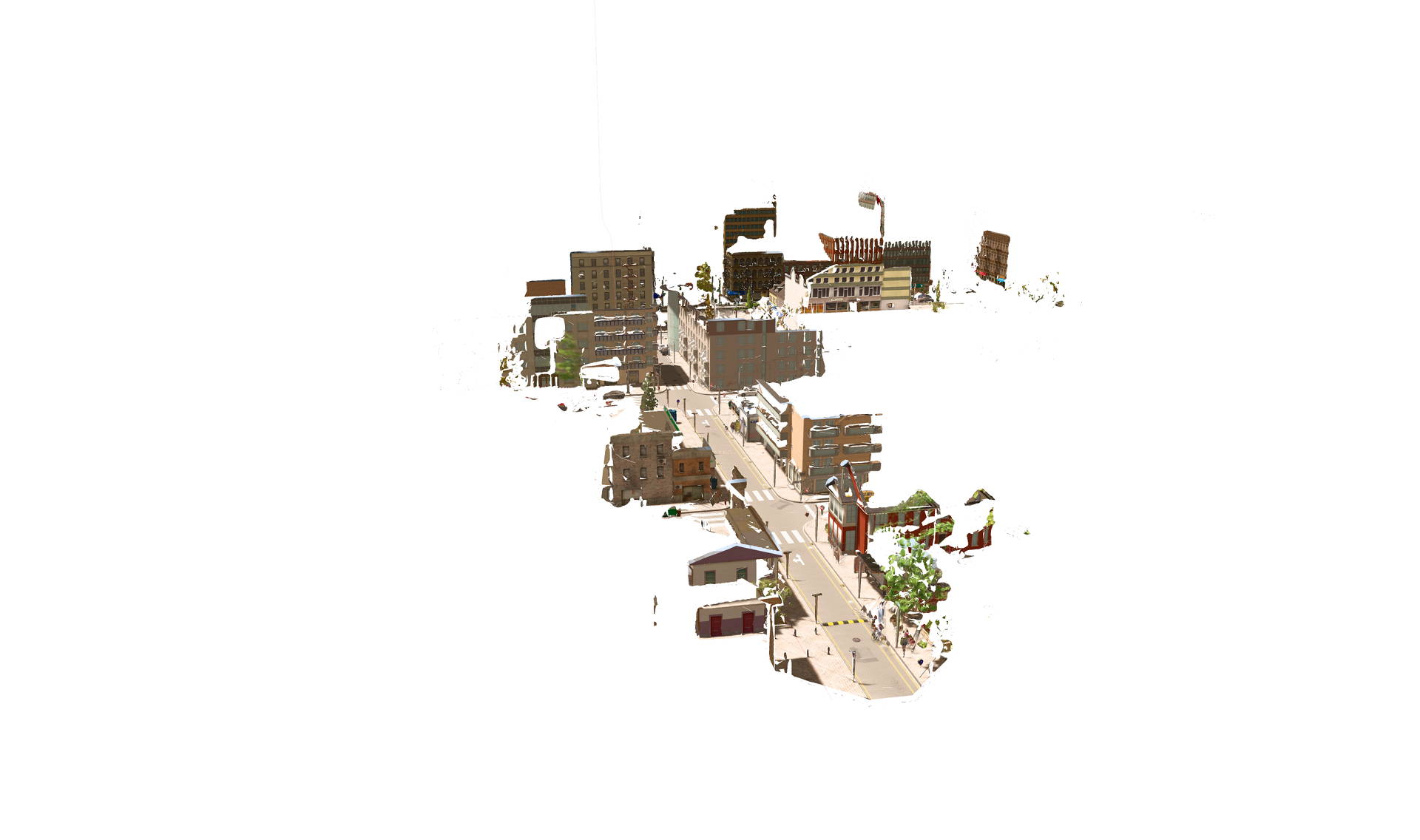}}
		\settoheight{\nIH}{\includegraphics{synthia_2_rgb.jpg}}
		\newcommand\synthiaLeftTrim{.3472\nIW{}}
		\newcommand\synthiaBottomTrim{.2641\nIH{}}
		\newcommand\synthiaRightTrim{.3559\nIW{}}
		\newcommand\synthiaTopTrim{.2641\nIH{}}
		\begin{figure}
			\captionsetup[subfloat]{labelformat=empty}
			\centering
			\subfloat[]{
				\includegraphics[trim=\synthiaLeftTrim{} \synthiaBottomTrim{} \synthiaRightTrim{} \synthiaTopTrim{},clip,width=0.47\columnwidth]{synthia_2_rgb.jpg}
			}
			\subfloat[]{
				\includegraphics[trim=\synthiaLeftTrim{} \synthiaBottomTrim{} \synthiaRightTrim{} \synthiaTopTrim{},clip,width=0.47\columnwidth]{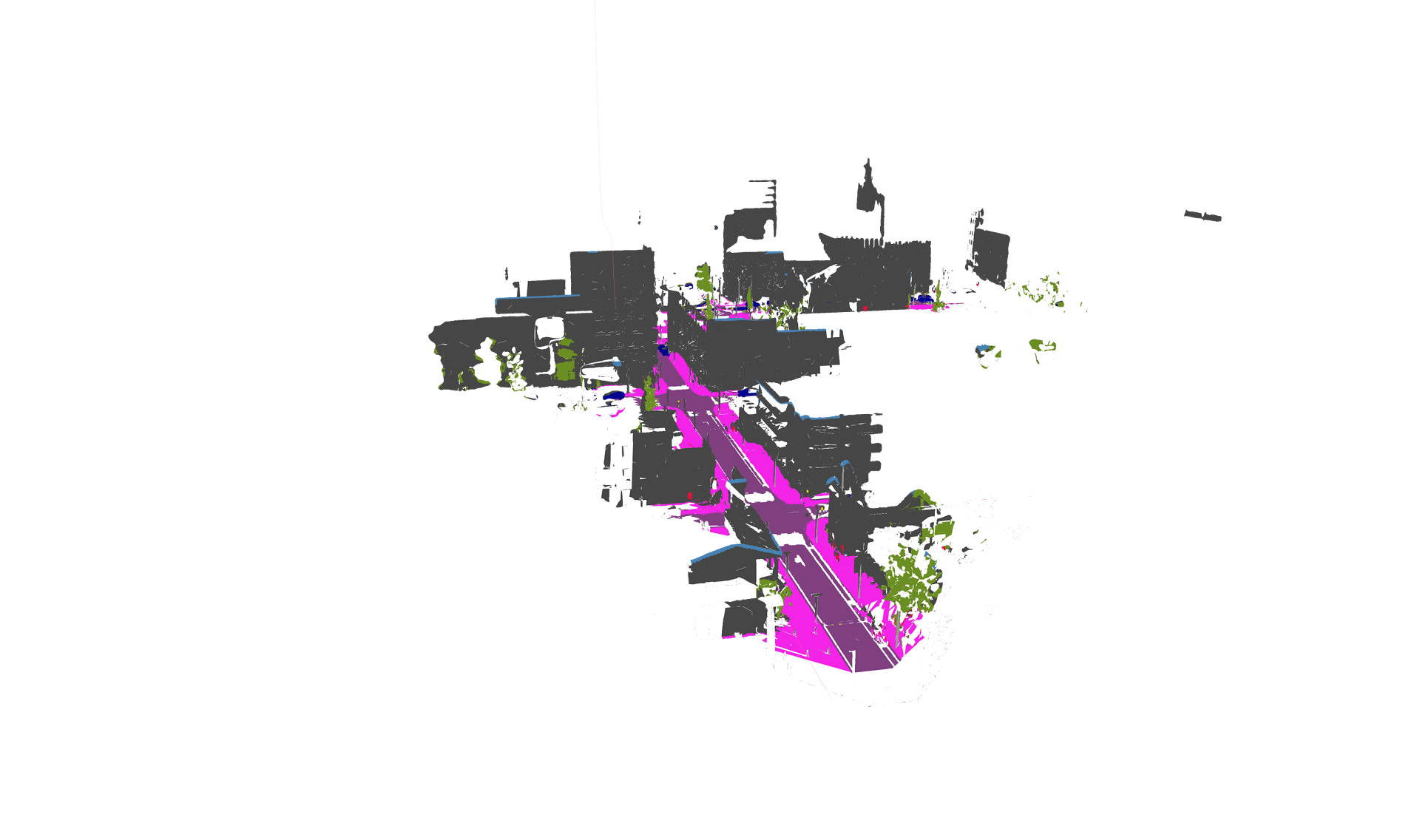}
			}
			\caption{Synthia semantic map: We reconstruct a semantic map from a subset of frames from the Synthia dataset~\citep{ros2016synthia}. We use the ground truth data and apply noise to evaluate the impact of camera misalignment on the semantic map accuracy.}
			\label{fig:SynthiaView}
		\end{figure}

		\definecolor{SynthiaSky}{RGB}{70, 130, 180}
		\definecolor{SynthiaBuilding}{RGB}{70,70,70}
		\definecolor{SynthiaRoad}{RGB}{128, 64, 128}
		\definecolor{SynthiaSidewalk}{RGB}{244, 35, 232}
		\definecolor{SynthiaVegetation}{RGB}{107, 142, 35}
		\definecolor{SynthiaPole}{RGB}{153, 153, 153}
		\definecolor{SynthiaCar}{RGB}{0, 0, 142}
		\definecolor{SynthiaSign}{RGB}{220, 220, 0}
		\definecolor{SynthiaLaneMarking}{RGB}{85, 55, 40}
		\definecolor{SynthiaTrafficLight}{RGB}{250, 170, 30}
		\begin{figure}
			\begin{minipage}[t]{0.8\columnwidth}
				\vspace{0pt}
				\resizebox {1.0\columnwidth} {!} {
				\begin{tikzpicture}
					\begin{axis}[ scale only axis, xmin=0,xmax=100, xlabel={Noise [\%]}, ylabel=Intersection-over-Union (IoU),
					legend pos = outer north east]
					\addplot [color=SynthiaSky, smooth, tension=0.1]  table [x=noise, y=sky, col sep=comma, mark=none] {noise_both_per_class.txt};
					\addplot [thick,color=SynthiaBuilding, smooth, tension=0.1]  table [x=noise, y=building, col sep=comma, mark=none] {noise_both_per_class.txt};
					\addplot [thick,color=SynthiaRoad, smooth, tension=0.1]  table [x=noise, y=road, col sep=comma, mark=none] {noise_both_per_class.txt};\\
					\addplot [thick,color=SynthiaSidewalk, smooth, tension=0.1]  table [x=noise, y=sidewalk, col sep=comma, mark=none] {noise_both_per_class.txt};\\
					\addplot [thick,color=SynthiaVegetation, smooth, tension=0.1]  table [x=noise, y=veg, col sep=comma, mark=none] {noise_both_per_class.txt};\\
					\addplot [color=SynthiaPole, smooth, tension=0.1]  table [x=noise, y=pole, col sep=comma, mark=none] {noise_both_per_class.txt};\\
					\addplot [color=SynthiaCar, smooth, tension=0.1]  table [x=noise, y=car, col sep=comma, mark=none] {noise_both_per_class.txt};\\
					\addplot [color=SynthiaSign, smooth, tension=0.1]  table [x=noise, y=sign, col sep=comma, mark=none] {noise_both_per_class.txt};\\
					\addplot [color=SynthiaTrafficLight, smooth, tension=0.1]  table [x=noise, y=light, col sep=comma, mark=none] {noise_both_per_class.txt};
					\end{axis}
				\end{tikzpicture}
				}
			\end{minipage}%
			\begin{minipage}[t]{0.2\columnwidth}
				\vspace{0pt}\raggedright
				\renewcommand{\arraystretch}{1.55}
				\begin{tabular}{>{\centering\arraybackslash}p{0.8cm}}
					\cellcolor{SynthiaSky}\textcolor{white}{Sky}\\
					\cellcolor{SynthiaBuilding}\textcolor{white}{Build.}\\
					\cellcolor{SynthiaRoad}\textcolor{white}{Road}\\
					\cellcolor{SynthiaSidewalk}Path\\
					\cellcolor{SynthiaVegetation}Veg.\\
					\cellcolor{SynthiaCar}\textcolor{white}{Car}\\
					\cellcolor{SynthiaPole}Pole\\
					\cellcolor{SynthiaSign}Sign\\
					\cellcolor{SynthiaTrafficLight}Light\\
				\end{tabular}
				\renewcommand{\arraystretch}{1.0}
			\end{minipage}
			\caption{Registration robustness: Incorrect sensor poses for semantic map creation affects the accuracy as measured by IoU of larger object classes like buildings less than light posts or signs. Increasing amounts of noise are applied on the translation~($\leq \SI{0.5}{\metre}$) and rotation~($\leq \SI{5}{\degree}$) of the sensor poses.}
			\label{fig:Misalign}
		\end{figure}
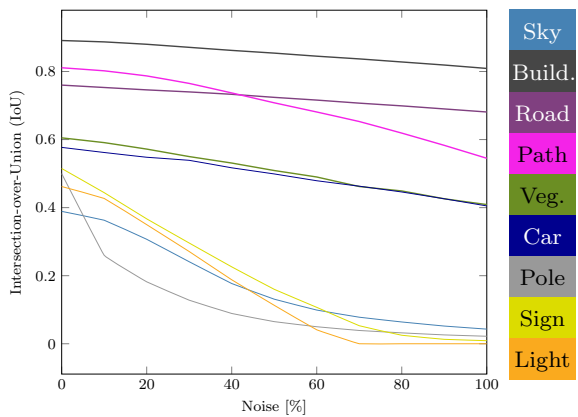

	\subsection{Runtime Performance}
		We evaluate the runtime performance of our meshing and texturing modules separately, as they are performed sequentially with no overlap.
		\reffig{fig:poisson_perf_compare} shows the resulting meshes after Poisson reconstruction for our simplified cloud, and the na\"ively aggregated full point set recorded by the Velodyne scanner. It can be observed that the reconstruction quality does not suffer while the runtime and memory consumption is significantly decreased (see \reftab{tab:PoissonRecon}).

	\bgroup
	\arrayrulecolor{gray} 
		\begin{table}[]
		\centering
		\begin{tabular}{|c|c|c|c|c|}
			\hline
			\textbf{Cloud} &
			\textbf{\#Points} &
			\textbf{\#Verts}&
			\textbf{Time(\SI{}{\second})} &
			\textbf{Mem(\SI{}{\giga\byte})}\\
			\hline

			full&
			\num{103}M & 
			\num{4.5}M & 
			\num{521.9} & 
			\num{2.82} \\ 
			\hline

			simple& 
			\num{21}M & 
			\num{3.3}M & 
			\num{193.8} & 
			\num{1.99} \\ 
			\hline
		\end{tabular}
		\caption{Poisson reconstruction using the na\"ively aggregated cloud and our edge-aware simplified cloud. We report the number of points of the input cloud, the number of vertices of the reconstructed mesh, and the time and peak memory used by the reconstruction process.}
		\label{tab:PoissonRecon}
	\end{table}
	\egroup

		\newcommand\poissonComparisonLeftTrim{50bp}
		\newcommand\poissonComparisonBottomTrim{100bp}
		\newcommand\poissonComparisonRightTrim{250bp}
		\newcommand\poissonComparisonTopTrim{100bp}
		\begin{figure}
			\captionsetup[subfloat]{labelformat=empty}
			\centering
			\subfloat[]{
			\includegraphics[trim=\poissonComparisonLeftTrim{} \poissonComparisonBottomTrim{} \poissonComparisonRightTrim{} \poissonComparisonTopTrim{},clip,width=0.47\columnwidth]{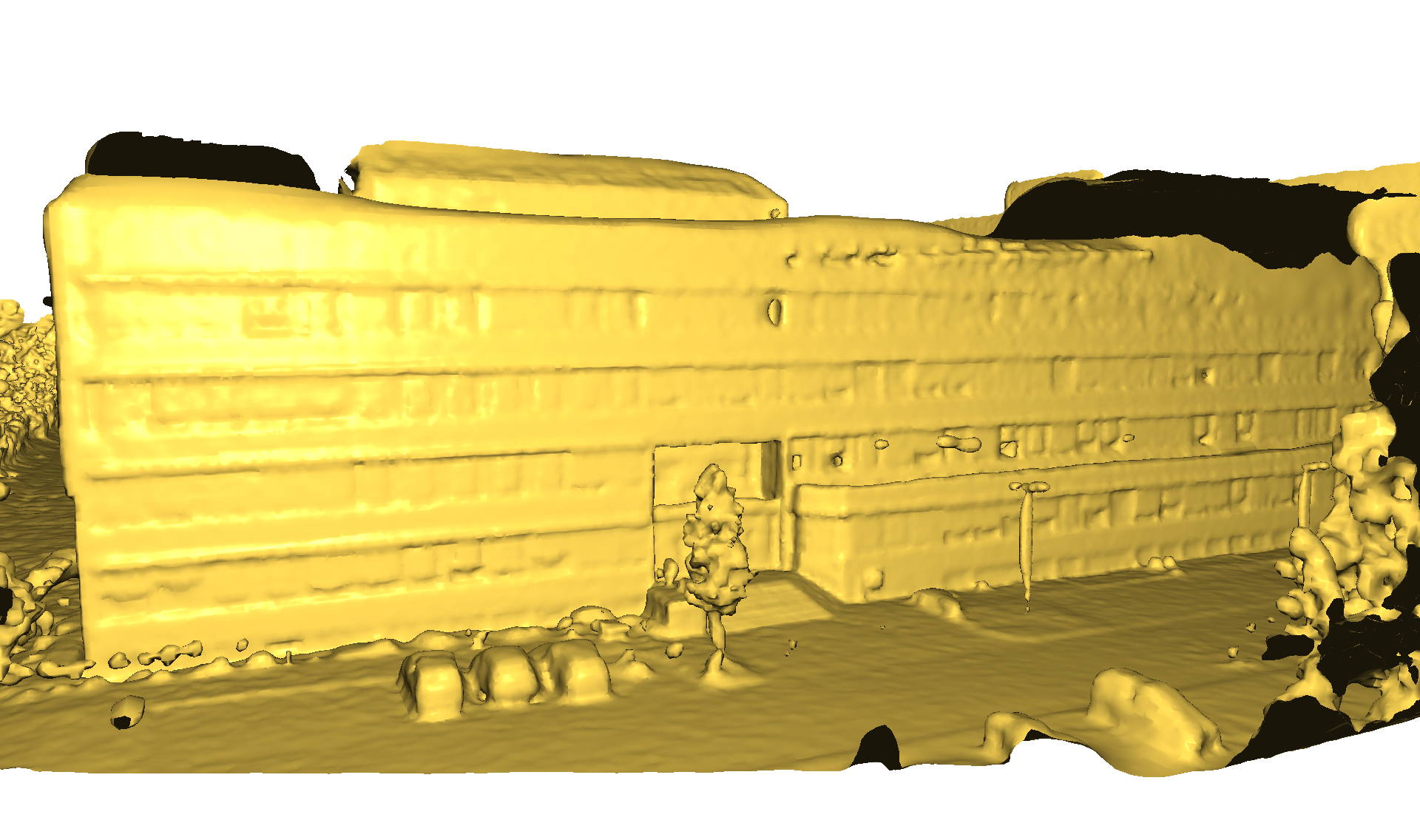}
			}
			\subfloat[]{
	\includegraphics[trim=\poissonComparisonLeftTrim{} \poissonComparisonBottomTrim{} \poissonComparisonRightTrim{} \poissonComparisonTopTrim{},clip,width=0.47\columnwidth]{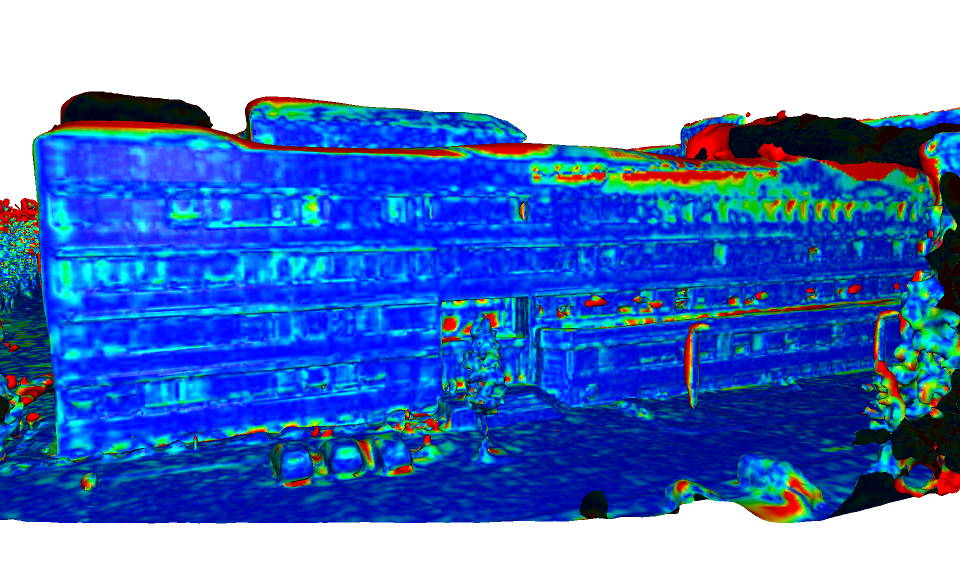}
			}
			\caption{Poisson reconstruction comparison: Reconstruction from the edge-aware simplification (left), and its difference (right) toward the full reconstruction.
			The colormap denotes the deviation between the two meshes where red equals a difference of \SI{15}{\centi\meter} and blue shows no difference. 
			The deviation is minimal in areas of interest while reconstruction after simplification is faster (see \reftab{tab:PoissonRecon}).}
			\label{fig:poisson_perf_compare}
		\end{figure}

		\begin{figure}
		\includegraphics{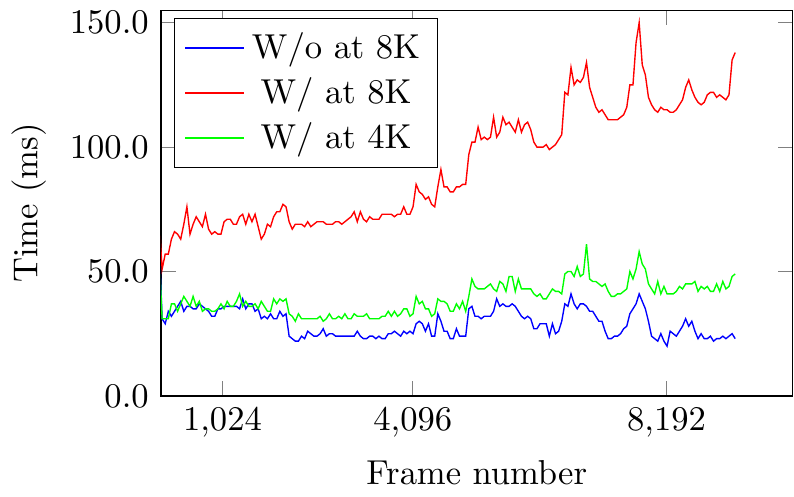}
			\caption{Timing results for the courtyard dataset: Semantic integration using a texture resolution of 8K without decommitting (blue line) can be performed in real-time. Enabling the decommiting at the same texture resolution (red line) proves to be too slow for real-time usage.
			Lowering the texture resolution to 4K allows the semantic integration with decommittment (green line) to be performed at real-time speeds.}
			\label{plt:Timings}
		\end{figure}

		The runtime of the semantic integration on the courtyard dataset is summarized in \reffig{plt:Timings}. We achieve real-time performance with an average texturing time of \SI{27.1}{\milli\second} per frame using a \num{8}K texture. Decommitting the sparse volume is, however, a demanding functionality, and causes the average time per frame to increase to \SI{90.1}{\milli\second}. Nevertheless, the semantic integration achieves \SI{38.6}{\milli\second} per frame for a smaller texture resolution of \num{4}K.

	\subsection{Memory Consumption}
\begin{figure}
			\includegraphics{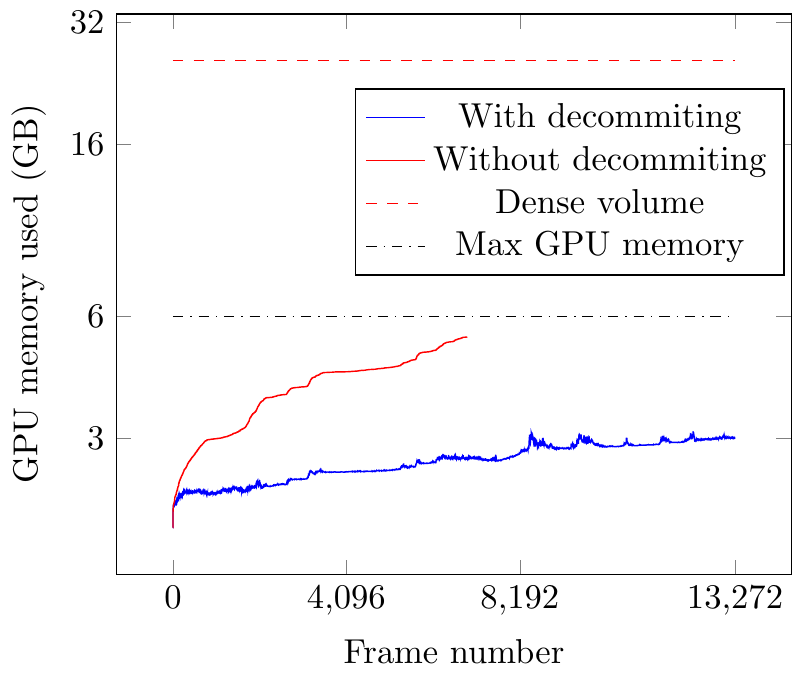}
			\caption{GPU memory usage for \num{10}K texture on the courtyard dataset: Dense allocation (red dashed line) would occupy more than \SI{25.7}{\giga\byte}. Sparse allocation (red line) without decommitting quickly overburdens the available \SI{6}{\giga\byte} causing a system failure. The memory usage drops with decommitting (blue line) below \SI{3.1}{\giga\byte} at all times enabling the full reconstruction of the semantic map.}
			\label{plt:GPUmem}
		\end{figure}

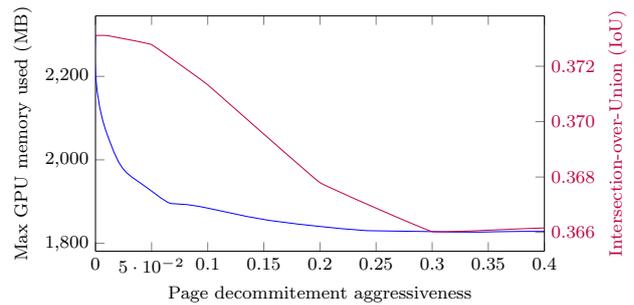
\begin{figure}
			\resizebox {\columnwidth} {!} {
				\begin{tikzpicture}
				\begin{axis}[width=9cm, height=5.5cm, xmin=0,xmax=0.4, axis y line*=left, xlabel=Page decommitement aggressiveness, ylabel=Max GPU memory used ($\SI{}{\mega\byte}$)]
				\addplot [color=blue, smooth, tension=1]  table [x=thresh, y=mem, col sep=comma, mark=none] {mem_decomm_iou_3.txt};
				\end{axis}
				\begin{axis}[width=9cm, height=5.5cm, xmin=0,xmax=0.4, axis y line*=right, axis x line=none, ylabel=\textcolor{purple}{Intersection-over-Union (IoU)},
				yticklabel style=purple,
				y tick label style={
					/pgf/number format/.cd,
					fixed,
					fixed zerofill,
					precision=3
				},]%
				\addplot [color=purple, smooth, tension=0.1]  table [x=thresh, y=iou, col sep=comma, mark=none] {mem_decomm_iou_3.txt};
				\end{axis}
				\end{tikzpicture}
			}
			\caption{Average GPU memory usage for different decommitting thresholds on NYUv2:
			Increasing the decommitting threshold quickly reduces the memory consumption (blue line), while the IoU decreases slowly. As a consequence, we typically fix the threshold to \num{0.1}.}
			\label{plt:GPUDecommitting}
		\end{figure}

		Memory usage of the texturing system is also evaluated with and without decommitting of pages of the sparse texture. The results for the courtyard dataset are summarized in \reffig{plt:GPUmem}.
		We also analyze the relation between the decommitting threshold (the probability below which the pages in the sparse texture are deallocated) and the Intersection-over-Union~(IoU) in \reffig{plt:GPUDecommitting}. We evaluate this measure on the NYU dataset due to the presence of more labeled images than in the courtyard dataset which allows for a more accurate evaluation. We observe that while low values of the threshold greatly reduces memory usage, higher values cause the IoU to degrade as more valuable information from the semantic texture is disregarded. However the decrease is still minor ($\le \SI{0.6}{\percent}$), as we restrain from decommitting pages that contain the argmax class. For further experiments, we chose a threshold value of \num{0.1}.

	\subsection{Texture Resolution and Semantic Accuracy}
	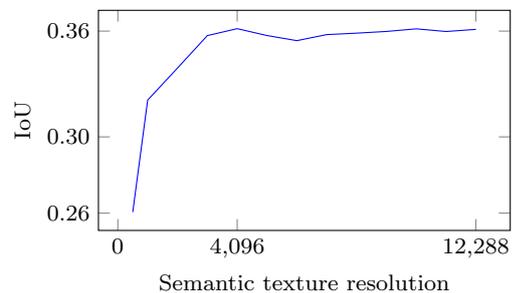
\begin{figure}
		\begin{tikzpicture}
		\begin{axis}[
		width=7cm,
		height=4.5cm,
		legend style={at={(0.02,0.98)},anchor=north west },
		xlabel={Semantic texture resolution},
		ylabel={IoU},
		y tick label style={/pgf/number format/.cd, fixed, fixed zerofill,precision=2},
		xtick={0,4096,12288},
		xticklabel style={
			/pgf/number format/fixed,
			/pgf/number format/precision=5
		},
		scaled x ticks=false,
		ytick={0.256, 0.30, 0.361}
		]
		\addplot [color=blue]  table [x=tex_size, y=mean_iou, col sep=comma, mark=none, patch,mesh,patch type=quadratic spline] {texsize_iou.txt};
		\end{axis}
		\end{tikzpicture}
		\caption[Texture resolution and IoU]{Texture resolution and IoU: We evaluate the mean IoU on the courtyard dataset as we increase the resolution of the semantic texture. The accuracy quickly rises and converges at a resolution of around \num{4}K.}
		\label{plt:TexSize}
	\end{figure}

	We evaluate the impact of the semantic texture resolution and the accuracy of the semantic map. We perform the evaluation on the courtyard dataset as it spans a larger area than the NYU dataset and therefore the impact of the texture size becomes more noticeable. Semantic texture integration is performed for a series of texture resolutions ranging from \num{512} to \num{12288} pixels and IoU is evaluated for each one. We observe that the IoU steadily increased ans becomes stable at around a resolution of $4$k.
	
	\subsection{Distance Weighting and Semantic Accuracy}
	\renewcommand{\tabcolsep}{1pt}
	\renewcommand{\arraystretch}{1.2} 
	\begin{table}
		\centering
		\begin{tabular}{|>{\centering\arraybackslash}m{1.1cm}
				|>{\centering\arraybackslash}m{1.75cm}
				|>{\centering\arraybackslash}m{1.75cm}
				|>{\centering\arraybackslash}m{1.75cm}
				|>{\centering\arraybackslash}m{1.75cm}|}
			
			\hline
			& No weight & \SI{30}{\metre}-\SI{100}{\metre} & \SI{0}{\metre}-\SI{100}{\metre} & \SI{0}{\metre}-\SI{30}{\metre}\\ 
			\hline
			IoU & 0.294 & 0.3259 & \textbf{0.3289} & 0.309\\ 
			\hline			
		\end{tabular}		
		\caption{Distance fall-off: We experiment with various thresholds for $d_{min}$ and $d_{max}$ for the linear fall-off during semantic integration on the courtyard dataset. Having no distance weighting (full confidence in the semantic information regardless of distance), achieves the lowest IoU. Modifying $d_{min}$ has a marginal effect while restricting $d_{max}$ is detrimental as too much distant information is discarded.}
		\label{fig:DistanceFalloff}
	\end{table}
	During semantic integration the only weighting applied while projecting from the 2D segmentation into the 3D scene is based on a linear fall-off between $d_{min}$ and $d_{max}$. The impact of the weighting is evaluated on the courtyard dataset as it presents larger variability in terms of distance. 
	Three different values for the distance thresholds are evaluated and compared to no weighting (full confidence in the semantic information regardless of distance). The results are gathered in \reftab{fig:DistanceFalloff}. We observe that having no distance weighting results in the lowest IoU while the highest one was achieved by a linear fall-off between \SI{0}{\metre} and \SI{100}{\metre}. Small variations in the thresholds have little impact unless the $d_{max}$ is severely restricted (for example to only \SI{30}{\metre}) at which point distant parts of the scene are ignored and thus severely affecting the IoU.
	
	\section{Limitations}
		In the previous sections we have seen the impact of incorrect poses for semantic mapping as well as inadequate geometric meshes reduce the accuracy especially for small and thin classes.
		Another limitation arises from the Expectation Maximization strategy itself. The current belief is reinforced and thus not able to correct the estimate in all cases.
		Currently, dynamic objects are completely disregarded in our approach, but often present in collected data. This introduces artifacts in the reconstruction as well as the semantic map and reduces the accuracy for moving classes like pedestrians or cars. Furthermore, retraining from such sequences will bias the semantic segmentation towards static classes. Hence, incorporating object tracking could alleviate this problem.
		In terms of implementation the sparsity of the probability volume is greatly influenced by the quality of the UV parametrization. If areas of the mesh that are spatially close also lie nearby in the parametric space and are from the same class then they cover a page in GPU memory more efficiently. However, in the extreme case where each triangle is mapped to independent texture coordinates, the chance for decommitting reduces significantly. Luckily, most man-made environments are generally planar and their parametrization can be easily performed. Despite this, the limitation is an important one and it rises the question of a possibly semantic-aware parametrization in which the parametric area is influenced by the class, e.g. assigning lower resolution for vegetation and buildings and higher for cars and pedestrians.		
		
\section{Conclusion}
	We presented a novel semantic mapping system that uses textured meshes to store the semantic information and allows to couple semantic and geometric representation at independent resolution improving scalability and accuracy even for large datasets.
	The mesh-based representation allows to enforce spatial and temporal consistency over multiple observations as well as enabling semi-supervised retraining from novel view points by propagation of labels in an iterative manner.
	 Quantitative and qualitative results on the NYUv2 dataset demonstrate the benefits of our approach compared to the state of the art method SemanticFusion.

\begin{acknowledgements}
We would like to thank David Droeschel for his effort in providing accurate poses for the courtyard dataset.
This work was supported by grant BE 2556/7 of the German Research Foundation (DFG).
\end{acknowledgements}

\bibliographystyle{spbasic}      
\bibliography{bibliography.bib}   
\end{document}